\documentclass[journal]{IEEEtran}

\usepackage{booktabs}
\usepackage{graphicx}
\usepackage{subcaption}
\usepackage{multirow}
\usepackage{array}
\usepackage{amsmath}
\usepackage{hyperref}
\usepackage{epstopdf}
\usepackage[linesnumbered,ruled]{algorithm2e}
\usepackage{multicol}
\usepackage{marvosym}
\usepackage{enumitem}
\usepackage{amsthm}

\usepackage{float}
\usepackage{graphics}
\usepackage{amssymb}
\usepackage{pifont}

\usepackage[justification=centering]{caption}
\usepackage{indentfirst}
\usepackage{makecell}

\usepackage{cite}

\newtheorem{mydef}{Definition}

\hypersetup{hidelinks}

\title{A robust three-way classifier with shadowed granular-balls based on justifiable granularity}

\author{
    Jie Yang\textsuperscript{\href{https://orcid.org/0000-0002-6580-9287}{\includegraphics[scale=0.05]{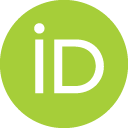}}},
    Lingyun Xiaodiao,
    Guoyin Wang\textsuperscript{\href{https://orcid.org/0000-0002-8521-5232}{\includegraphics[scale=0.05]{ORCIDiD.png}}},~\IEEEmembership{Senior Member,~IEEE},
    Witold Pedrycz,~\IEEEmembership{Life Fellow,~IEEE},\\
    Shuyin Xia\textsuperscript{\href{https://orcid.org/0000-0001-5993-9563}{\includegraphics[scale=0.05]{ORCIDiD.png}}},
    Qinghua Zhang\textsuperscript{\href{https://orcid.org/0000-0002-6154-4656}{\includegraphics[scale=0.05]{ORCIDiD.png}}},~\IEEEmembership{Senior Member,~IEEE},
    Di Wu\textsuperscript{\href{https://orcid.org/0000-0002-7788-9202}{\includegraphics[scale=0.05]{ORCIDiD.png}}}

    \thanks{\Letter \, Corresponding author: Qinghua Zhang (e-mail: zhangqh@cqupt.edu.cn), Di Wu(e-mail: wudi1986@swu.edu.cn).}
    \thanks{Jie Yang is with the Key Laboratory of Cyberspace Big Data Intelligent Security, Ministry of Education, Chongqing University of Posts and Telecommunications, Chongqing 400065, China, and with the School of Physics and Electronic Science, Zunyi Normal University, Zunyi 563002, China(e-mail: yj530966074@foxmail.com).
    
    Lingyun Xiaodiao, Guoyin Wang, Shuyin Xia, Qinghuang Zhang are with the Key Laboratory of Cyberspace Big Data Intelligent Security, Chongqing University of Posts and Telecommunications, Chongqing 400065, China(e-mail: S230201126@stu.cqupt.edu.cn, wanggy@cqupt.edu.cn, xiasy@cqupt.edu.cn, zhangqh@cqupt.edu.cn).

    Witold Pedrycz is with the computational intelligence with the Department of Electrical and Computer Engineering, University of Alberta, Edmonton, AB, Canada(e-mail: wpedrycz@ualberta.ca).
    
    Di Wu is with College of Computer and Information Science, Southwest University, Chongqing 400715, China, and with the Key Laboratory of Cyberspace Big Data Intelligent Security, Ministry of Education, Chongqing University
    of Posts and Telecommunications, Chongqing 400065, China(e-mail: wudi1986@swu.edu.cn).
    }
}
    
\begin{document}

\maketitle

\begin{abstract}
    The granular-ball (GB)-based classifier introduced by Xia, exhibits adaptability in creating coarse-grained information granules for input, thereby enhancing its generality and flexibility. Nevertheless, the current GB-based classifiers rigidly assign a specific class label to each data instance and lacks of the necessary strategies to address uncertain instances. These far-fetched certain classification approachs toward uncertain instances may suffer considerable risks. To solve this problem, we construct a robust three-way classifier with shadowed GBs for uncertain data. Firstly, combine with information entropy, we propose an enhanced GB generation method with the principle of justifiable granularity. Subsequently, based on minimum uncertainty, a shadowed mapping is utilized to partition a GB into Core region, Important region and Unessential region. Based on the constructed shadowed GBs, we establish a three-way classifier to categorize data instances into certain classes and uncertain case. Finally, extensive comparative experiments are conducted with 2 three-way classifiers, 3 state-of-the-art GB-based classifiers, and 3 classical machine learning classifiers on 12 public benchmark datasets. The results show that our model demonstrates robustness in managing uncertain data and effectively mitigates classification risks. Furthermore, our model almost outperforms the other comparison methods in both effectiveness and efficiency.
\end{abstract}

\begin{IEEEkeywords}
  Shadowed granular-balls; justifiable granularity; granular-ball generation; three-way classifier.
\end{IEEEkeywords}

\section{Introduction}
\label{sec:1}
Granular computing (GrC) \cite{1,2,3,4} focuses on the formation, processing, and communication of information granules, aiming to mimic human cognitive thinking in addressing complex problems. Rough sets \cite{5}, fuzzy sets \cite{6}, and quotient spaces \cite{7} are three primary models of GrC. Rough sets are widely used to measure the uncertainty and incompleteness inherent in information systems. To handle continuous data, neighborhood rough sets (NRS) \cite{8} introduce the concept of neighborhood granulation to convert the equivalence relation into the covering relation in neighborhood space. Furthermore, numerous extended GrC-based classifiers \cite{9,10,11,12,yue2020shadowed} were developed by utilizing information granules as the fundamental computational unit, enhancing the efficiency of knowledge discovery. However, a notable limitation of these classifiers is that they primarily treat granules as a preliminary feature processing method, without modifying the underlying mathematical model or elevating the core performance of the classifiers themselves.

Based on the idea of GrC, granular-ball computing (GBC) \cite{13,14,xia2021granular}, introduced by Xia, represents a groundbreaking approach to data processing and knowledge representation. This method replaces traditional information granule inputs with granular-balls (GBs), adhering to the principle of global topology precedence \cite{chen1982topological}. Over the years, GBC has undergone continuous advancements in terms of its methodologies and applications. Xia \cite{15} proposed the GB-based $k$NN, which significantly outperforms the existing $k$NN in terms of efficiency, especially when dealing with large-scale datasets, achieving an improvement of hundreds of times. Furthermore, Xia \cite{16} introduced a novel rough sets model, namely, granular-ball rough sets (GBRS). Compared to traditional NRS methods, GBRS stands out as a multi-granularity learning tool, offering greater robustness and efficiency by substituting neighborhood granules with GBs. In addition, Chen \cite{17} proposed a GB-based attribute selector to achieve the superior classification performance through effective feature reduction. Xie \cite{18} introduced a fast and stable GB generation method based on the attention mechanism. Xie \cite{19} constructed a multi-granularity representation of the data using the GBC model, thereby boosting the algorithm's time efficiency. Chen \cite{20} innovated by introducing a novel GB-based density peaks clustering, resulting in significantly reduced runtime without the need for parameterization. However, current researches on GB generation in GBC only consider the purity thresholds to control the granularity of GB space, which is not in accord with various practical applications. 

From the perspective of the justifiable granularity principle\cite{22,23,24}, both coverage and specificity should be taken into account during the GB generation process. Coverage refers to a GB that encompass as much experimental evidence (data) as possible. Specificity refers to all GBs generated on the universe having distinct and well-defined semantics or all GBs being well distinguished from each other. Typically, the fewer the data instances in a GB, the higher the specificity. Therefore, it is significant to generate justifiable GB space to solve problems by combining with the level of coverage and specificity. Moreover, the above GB-based classification methods rigidly ascribe a fixed class label to each data instance, lacking strategies to handle instances with uncertainty. The methodology of uncertain data classification is invaluable in mitigating decision risk and enhancing decision efficiency through human-machine collaboration, thus playing a pivotal role in decision support systems. For instance, when applying classification methods to develop a computer-aided diagnosis (CAD) system for liver cancer \cite{25}, it is crucial to accurately classify uncertain tumors for further cautious diagnosis and certain far-fetched classifications produced by the system may cause serious costs.

As is well known, three-way decision (3WD) \cite{26,27,28} theory proposed by Yao has emerged as a promising approach to tackle complex problems involving uncertainty. The fundamental principle of 3WD lies in dividing a universe into three distinct regions, with each region corresponding to a specific decision action. As a generalization of the traditional two-way decision model, 3WD further incorporates a third option, enabling a trisecting-and-acting approach to decision-making. Currently, 3WD has found widespread application across various fields \cite{29,30,31,32}. Xu \cite{33} enhanced the flexibility and evolution capability of concept learning, addressing the limitations of existing two-way learning approaches by incorporating a novel cognitive mechanism and movement 3WD method. Zhang \cite{34} introduced an efficient multi-scale decision system method that integrates a sequential 3WD model with the Hasse diagram to optimize scale combination selection and attribute reduction. Du \cite{35} presented a novel multistep three-way clustering algorithm that enhances the accuracy and adaptability of cluster representations. In term of the advantage of 3WD, to address the limitations of GB-based classification models in classifying uncertain data, we introduce 3WD to construct the three-way approximations of GBs, which is called shadowed GBs. More detailed, the traditional GBs are extended to shadowed GBs for both certain regions and uncertain boundaries from the perspective of uncertainty, and shadowed GBs constructs a tripartitioned approximation of data distribution for three-way classification. The data instances will be categorized into a certain class or an uncertain case based on their locations relative to the shadowed GBs. Specifically, the Core region of GBs corresponding to the same class is confidently determined the class of instances, whereas the Important region has uncertainty for classification, and the Unessential region has almost no contribution to classification. This ensures more effective to classify uncertain data. The major contributions of this paper are summed up as follows:
\begin{enumerate}
    \item[(1)] Combine with information entropy, we propose an improved GB generation method based on the principle of justifiable granularity.
    \item[(2)] We introduce 3WD to construct and optimize shadowed GBs for modeling uncertain data.
    \item[(3)] We further establish a three-way classifier with shadowed GBs to categorize data instances into certain classes and uncertain case. 
\end{enumerate}

The following sections of this paper are structured with the subsequent manner. In section \ref{sec:2} is a review of related preliminary definitions. In Section \ref{sec:3}, an improved GB generation method is constructed based on the principle of justifiable granularity. In Section \ref{sec:4} introduces the construction of shadowed GBs. Then, a three-way classifier with shadowed GBs is presented. The relevant experiments for the verification of viability and rationality of our models are shown in Section \ref{sec:5}. Finally, Section \ref{sec:6} summarizes the conclusions.

\section{Preliminaries}
\label{sec:2}
In this section, to set up the framework of this paper, we recall some necessary definitions related to GBC, shadowed sets and 3WD. 

GBC is first proposed by Xia\cite{15}, which is to use GBs to cover the sample space and replace traditional information granules with GBs for data processing. We first define the GBC as follow:
\begin{mydef}(GBC)\cite{15}\label{def:1}
    Let $U=\{x_1,x_2,...,x_n\}$ be a non-empty finite set. Generating $\mathbb{G}$ to replace the inputs for computation can be called GBC. Where $\mathbb{G}=\{gb_1,gb_2,...,gb_m\}$ is a granular-ball space of $U$, and a granular-ball $gb_i$ covers a subset of $U$ such that $gb_i=\{x_{i1},x_{i2},...,x_{i|gb_i|}\} (i=1,2,...,m)$.
\end{mydef}
Although a single GB contains a complete subset of $D$, we need to use some attributes of GB to replace the inputs for accelerated computation. Common attributes are defined as follow:
\begin{mydef}(GB Attributes)\cite{15}\label{def:gba}
    Let $U=\{x_1,x_2,...,x_n\}$ be a non-empty finite set, and $\mathbb{G}=\{gb_1,gb_2,...,gb_m\}$ is a granular-ball space of $U$. For each $gb_i\in \mathbb{G}$, $gb_i=\{x_{i1},x_{i2},...,x_{i|gb_i|}\} (i=1,2,...,m)$, its attributes can be defined as follow:
    \begin{align}
        &\quad c_{gb_i} = \frac{1}{|gb_i|} \sum_{j = 1}^{|gb_i|} x_{ij},
        \quad\quad
        r_{gb_i} = \frac{1}{|gb_i|} \sum_{j = 1}^{|gb_i|} \left \| x_{ij}-c_{gb_i} \right \|,\notag \\
        & l_{gb_i} = \max(l_{ij}(j=1,2,...,n)),  \quad
        p_{gb_i} = \frac{\left | \left \{ x_{ij} \mid l_{ij} = l_{gb_i} \right \}  \right | }{|gb_i|} 
    \end{align}
    Where $c_{gb_i}$ is the center, $r_{gb_i}$ is the radius, $l_{gb_i}$ is the label, and $p$ is the purity of $gb_i$. $l_{ij}$ is the label of $x_{ij}(j=1,2,...,|gb_i|)$.
\end{mydef}
The above attributes are commonly used in GBC, where the center and label of GB are generally used to replace all the samples in the GB as a input, and the radius and purity of GB are used as a weight of this input, which can distinguish the contributions of different GB.

Shadowed sets \cite{37} was proposed by Pedrycz and constructed through the fuzzy-rough transformation of fuzzy sets, which provided an effective tool to model and analyze the concepts with uncertainty. We can describe fuzzy-rough transformation as follows:
\begin{mydef}(Fuzzy Sets)\cite{6}\label{def:fu}
    Let $U=\{x_1, x_2, . . . , x_n\}$ be a non-empty finite set. A fuzzy set $\tilde{A}$ in $U$ is characterized by a membership function $\mu_{\tilde{A}}: U\to [0,1]$. For each element $x\in U$, $\mu_{\tilde{A}}(x)$ represents the fuzzy membership of $x$ in the fuzzy set $\tilde{A}$.
\end{mydef}
\begin{mydef}(Fuzzy-rough Transformation)\cite{37}\label{def:2}
    Let $U=\{x_1, x_2, . . . , x_n\}$ be a non-empty finite set. $\mu_{\tilde{A}}(x)$ is the fuzzy memberships of samples $x \in U$ in fuzzy set $\tilde{A}$ in $U$, mapping its into a triplet set $\{0, [0, 1], 1\}$ can be called fuzzy-rough transformation, and the mapping is formulated as follows:
    \begin{align}
        S^{\alpha}_{\mu_{\tilde{A}}}(x)=\left\{
        \begin{array}{@{}ll}
        1, & \mu_{\tilde{A}}(x)\ge1-\alpha\\
        {[}0,1{]}, & \alpha<\mu_{\tilde{A}}(x)<1-\alpha\\
        0, &  \mu_{\tilde{A}}(x)\le\alpha
        \end{array}
        \right.
    \end{align}
    Where $\alpha \in [0,0.5]$ is the threshold parameter, and after that mapping, we get a shadowed set by threshold pair $(\alpha,1-\alpha)$.
\end{mydef}
In this transformation, memberships lower than $\alpha$ will be reduced to 0, while memberships higher than $1-\alpha$ will be elevated to 1. So there will be a loss of membership and this loss can be called uncertainty variance\cite{37}\cite{38}. The uncertainty variance of transforming fuzzy memberships into a shadowed set can describe as follow:
\begin{mydef}(Uncertainty Variance)\cite{38}\label{def:3}
    Let $U=\{x_1, x_2, . . . , x_n\}$ be a non-empty finite set. $\mu_{\tilde{A}}(x)$ is the fuzzy memberships of samples $x \in U$ in fuzzy set $\tilde{A}$ in $U$, and $S^{\alpha}_{\mu_{\tilde{A}}}$ is a fuzzy-rough transformation on $\tilde{A}$. The uncertainty variance $V(\alpha)$ of this fuzzy-rough transformation is formulated as:
    \begin{align}
        V(\alpha) & = \sum_{x\in X} \left | \mu_{\tilde{A}}(x) -S^{\alpha}_{\mu_{\tilde{A}}}(x) \right | \notag\\
        & = \sum_{\mu_{\tilde{A}}(x)\le \alpha}S^{\alpha}_{\mu_{\tilde{A}}}(x) + \sum_{\mu_{\tilde{A}}(x)\ge 1-\alpha}(1-S^{\alpha}_{\mu_{\tilde{A}}}(x))
    \end{align}
\end{mydef}
To measure the uncertainty entropy in fuzzy set, Zhang \cite{39} proposed the average fuzziness of a fuzzy set can describe as follow:
\begin{mydef}(Average Fuzziness)\cite{39}\label{def:4}
    Let $U$ = $\{x_1, x_2, . . . , x_n\}$ be a non-empty finite set. $\mu_{\tilde{A}}(x)$ is the fuzzy memberships of samples $x \in U$ in fuzzy set $\tilde{A}$ in $U$, the fuzziness of $x$ is defined as:
    \begin{align}
        \hbar (x)=4\mu_{\tilde{A}}(x)(1-\mu_{\tilde{A}}(x))
    \end{align}
    The average fuzziness of $U$ is represented by:
    \begin{align}
        \Gamma (U)=\frac{1}{|U|} \sum_{x \in U}\hbar (x)
    \end{align}
\end{mydef}
The 3WD\cite{40} is proposed as an extension of the commonly used binary decision model through adding a third option. In general, the approach of 3WD divides the universe into the positive, negative, and boundary regions which denote the regions of acceptance, rejection, and noncommitment. The 3WD on a fuzzy set can describe as follow:
\begin{mydef}(3WD)\cite{40}\label{def:5}
    Let $U$ = $\{x_1, x_2, . . . , x_n\}$ be a non-empty finite set. $\mu_{\tilde{A}}(x)$ is the fuzzy memberships of samples $x \in U$ in fuzzy set $\tilde{A}$ in $U$, to divide it to three parts by 3WD under two thresholds $\alpha < \beta$ are defined as:
    \begin{align}
        \text{POS}_{\alpha,\beta} & = \{ x\in U|\mu_{\tilde{A}}(x) \ge \alpha\} \notag \\
        \text{NEG}_{\alpha,\beta} & = \{ x\in U|\mu_{\tilde{A}}(x) \le \beta\} \\
        \text{BND}_{\alpha,\beta} & = \{ x\in U|\alpha < \mu_{\tilde{A}}(x)< \beta\} \notag
    \end{align}
\end{mydef}

\section{Granluar-ball Generation Based on Justifiable Granularity}
\label{sec:3}

\begin{figure}[htbp]
    \centering
    \includegraphics[width=\linewidth]{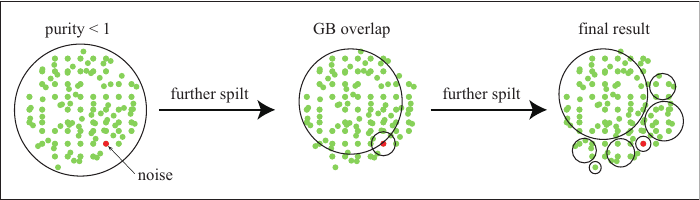}
    \caption{The GB generation with purity when facing noise}
    \label{fig:1}
\end{figure}

For GB generation, there are different approaches, e.g., using $k$-means\cite{15} or using $k$-division\cite{36} to split the GB. However, all generation methods can be summarized as starting with a GB that initially covers all samples, and then iteratively split it until a control condition is satisfied. The detailed process of GB generation method is outlined in Algorithm S1 of supplementary file. A key point in GB generation is setting the control condition, which directly affect the results of GB generation. Currently, most GB generation methods used purity as the control condition\cite{15,18,20,36,41}. Although this approach offers the advantages such as fast computation and strong interpretability, it overlooks the uncertainty existed in original dataset. As shown in Fig. \ref{fig:1}, GB generation with purity tend to generate more GBs when facing noise. This brings about two deficiencies as follows:
\begin{enumerate}
    \item[(1)]Overly split the final GBs, leading to a lack of justifiable granularity for problem solving and increase the number of GBs in subsequent calculations.
    \item[(2)]Generated GBs with noise label that impacts further compution and reduces the robustness of the GBC.
\end{enumerate}

To solve the above problems, searching for a reasonable stop condition is critical. Pedrycz\cite{24} proposed two criteria behind the principle of justifiable granularity, namely coverage and specificity, and they are conflict with each other\cite{42}. Based on the coverage entropy and specificity entropy on GBs are defined as follows:
\begin{mydef}(Coverage Entropy)\label{def:6}
    Let $U=\{x_1,x_2,...,x_n\}$ be a non-empty finite set, and $\mathbb{G}=\{gb_1,gb_2,...,gb_m\}$ is a granular-ball space of $U$. For each $gb_i \in \mathbb{G}(i=1,2,...,m)$, the coverage entropy of $gb_i$ is formulated as follow:
    \begin{align}
        H^c_{gb_i} = -\frac{\left | gb_i \right | }{\left | D \right | } \log \frac{\left | gb_i \right | }{\left | D \right | }
    \end{align}
    The coverage entropy of $\mathbb{G}$ is formulated as follow:
    \begin{align}
        H^c_{\mathbb{G}} = \sum_{gb_i\in \mathbb{G}} H^c_{gb_i}
    \end{align}
\end{mydef}
\begin{mydef}(Specificity Entropy)\label{def:7}
    Let $U=\{x_1,x_2,...,x_n\}$ be a non-empty finite set, and $\mathbb{G}=\{gb_1,gb_2,...,gb_m\}$ is a granular-ball space of $U$. For each $gb_i \in \mathbb{G}(i=1,2,...,m)$, $gb_i=\{x_{i1},x_{i2},...,x_{i|gb_i|}\}$, and the $lab=\{l_{i1},l_{i2},...,l_{i|gb_i|}\}$, where $l_{ij}$ is the label of $x_{ij}(j=1,2,...,|gb_i|)$. A label set $lab_{unique} = \{l_k|l_k\in lab\}$ only contains distinct labels in $lab$. The specificity entropy of $gb_i$ is formulated as follow:
    \begin{align}
        H^s_{gb_i} = -\sum_{l_k \in lab_{unique}} \frac{ | gb_i^{l_k}  | }{\left | gb_i \right | } \log \frac{ | gb_i^{l_k}  | }{\left | gb_i \right | }
    \end{align}
    Where $|gb^{l_k}_i|=|\{x_{ij}| l_{ij}=l_k\}|$. The specificity entropy of $\mathbb{G}$ is formulated as follow:
    \begin{align}
        H^s_{\mathbb{G}} = \sum_{gb_i \in \mathbb{G}} \frac{\left | gb_i \right | }{\left | D \right | } H^s_{gb_i}
    \end{align}
\end{mydef}
Coverage entropy refers to a GB that covers as much experimental data as possible. From the properties of information entropy, it is easy to deduce that coverage entropy strictly increases with the split of GBs. That is, when there is only an initial GB, coverage entropy takes the minimum value of 0, and when all GBs are split into single sample, coverage entropy takes the maximum value. Specificity entropy refers to all GBs having distinct semantics or being well distinguished from each other. Similar to coverage entropy, specificity entropy takes the maximum value when there is only an initial GB, and takes the minimum value of 0 when all GBs are split into single sample. Specificity entropy decreases with the split of GB. The detailed proof for the  monotonicity of coverage entropy and specificity entropy with the granularity being finer can be found in Section S2 of supplementary file. Therefore, coverage entropy and specificity entropy are contradictory. To balance these two entropies and achieve the justifiable granularity in the GB generation, a control condition is defined as follow:
\begin{mydef}(Measure of Justifiable Granularity)\label{def:8}
    Let $U=\{x_1,x_2,...,x_n\}$ be a non-empty finite set, and $\mathbb{G}$ is a granular-ball space of $U$. The measure of justifiable granularity on $\mathbb{G}$ is defined as follow:
    \begin{align}
        L_{\mathbb{G}} = \theta H^c_{\mathbb{G}} + \varepsilon (1-\theta) H^s_{\mathbb{G}}
    \end{align}
    Where $\theta$ is a parameter that controls the weight of two entropies, $\varepsilon$ is an adaptive parameter that balances the slope of two entropies, in this paper, we use $ \varepsilon = \max H^c_{\mathbb{G}}/\max H^s_{\mathbb{G}}$. 
\end{mydef}
The control condition of GB generation can be defined as follow:
\begin{align}
    \arg\min_{\mathbb{G}} L_{\mathbb{G}}
\end{align}

To conveniently compute the minimum $L_{\mathbb{G}}$, we use $\Delta$ to denote the difference in $L_{\mathbb{G}}$ after and before a GB split. When $\Delta$ is negative, it indicates that this split will decrease $L_{\mathbb{G}}$; when it's positive, it indicates that this split will increase $L_{\mathbb{G}}$. When we consider the scenario where GB are split until $purity=1$, the specificity entropy is already 0. Continuing to split will keep the specificity entropy at 0, while the coverage entropy will continue to increase. Therefore, when GBs are split until $purity=1$, the split must have gone through the GBs with the minimum $L_{\mathbb{G}}$, and further split will not result in the minimum $L_{\mathbb{G}}$. So, $purity=1$ will be a break condition in the process of finding the minimum $L_{\mathbb{G}}$. The GB generation method based on justifiable granularity is shown in Algorithm \ref{alg:1}.

\begin{algorithm} 
    \caption{The GB Generation Method Based on Justifiable Granularity}\label{alg:1}
    \KwIn{Datesets $D$, parameter $\theta$;}
    \KwOut{The GBs $\mathbb{G}$ with minimum $L_{\mathbb{G}}$;}
  
    initialize a $gb$ on $D$, then $current\_\mathbb{G} = \{gb\}$\;
    calculate $\varepsilon$, $L_{current\_\mathbb{G}}$ by Definition \ref{def:8}\;
    $min\_L,current\_L=L_{current\_\mathbb{G}}$\;

    \SetKwFunction{PreSplitGB}{Pre\_Split\_GB}
    \SetKwFunction{MainGB}{Main}

    \SetAlgoNlRelativeSize{0}
    \SetKwProg{Fn}{Function}{:}{\KwRet}

    \Fn{\PreSplitGB{$gb$}}{
        Split $gb$ to $gb_1$ and $gb_2$\;
        Calculate the $\Delta = L_{\{gb_1\cup gb_2\}} - L_{\{gb\}}$\;
        $gb.\Delta \gets  \Delta, gb.split \gets  \{gb_1\cup gb_2\}$;
    }

    \Fn{\MainGB}{
        \PreSplitGB{$current\_\mathbb{G}.top$}\;
        \While{\textbf{exist any} $gb$ \textbf{in} $current\_\mathbb{G}: gb.p \neq 1$}{
            $top\_gb = current\_\mathbb{G}.top$\;
            $new\_\mathbb{G}\gets top\_gb.split$\;
            \For{$gb$ \textbf{in} $new\_\mathbb{G}$}{
                \PreSplitGB{$gb$};
            }
            $current\_\mathbb{G}.delete(top\_gb)$\;
            \For{$gb$ \textbf{in} $new\_\mathbb{G}$}{
                $current\_\mathbb{G}.add(gb)$\;
                $current\_L\gets L+\textit{GB}.\Delta$;
            }
            \If{$current\_L<min\_L$}{
                $best\_\mathbb{G} \gets \mathbb{G}$\; 
                $min\_L \gets current\_L$;
            }
            Sort $current\_\mathbb{G}$ by $gb.\Delta$\ in ascending order;
        }
        \textbf{return} $best\_\mathbb{G}$\;  
    }
\end{algorithm}

In simple terms, Algorithm \ref{alg:1} splits the $\mathbb{G}$ by selecting a GB with the smallest $\Delta$ each time. Repeated this operation, until the break condition $purity=1$ is reached, then get the $\mathbb{G}$ with the smallest $L_{\mathbb{G}}$. To calculate the $\Delta$ for each GB, we modify the GB split operation to a pre-processing split. When a GB is added to the $\mathbb{G}$, it attempts to split the GB and records the $\Delta$ and split result. When this GB needs to be split, the pre-processing split result can be directly used. This pre-processing split does not increase the time complexity of GB generation, and each GB still requires only one split operation. The main difference between Algorithm \ref{alg:1} and traditional GB generation in Section S1 of supplementary file is that it only split a GB with the smallest $\Delta$ each time. This ensures that $L_{\mathbb{G}}$ always decreases in the fastest direction after each split. To implement this operation, we need to maintain a min-heap to record the $\mathbb{G}$. The push and pop operations on the min-heap are both less than $O(\log n)$, and in each round of the loop, whether it is based on $k$-means or $k$-division, the time complexity of split is $O(n)$. Overall, Algorithm \ref{alg:1} does not increase the time complexity compared with traditional GB generation algorithms.

\section{Three-way Classification with Shadowed Granluar-balls}
\label{sec:4}
GBC uses GB attributes in Definition \ref{def:gba} to replace the inputs, which results in each sample contributing equally to the GB it belongs to. For example, when faces with classification problems in GBC, GB attributes only offer the center and radius of each GB, and predicted sample only have two states, in or not in the GB. Although this method is simple enough, it ignores the high uncertainty of sample points near the radius, in other words, it lacks robustness. To address this issue, we construct shadowed GBs based on 3WD theory. First, the GB membership for samples is defined as follow:
\begin{mydef}(GB Membership)\label{def:9}
    Let $U=\{x_1,x_2,...,x_n\}$ be a non-empty finite set, and $\mathbb{G}=\{gb_1,gb_2,...,gb_m\}$ is a granular-ball space of $U$. Given a membership function $\mu_{gb_i}: gb \to [0,1]$ to each $gb_i \in \mathbb{G}(i=1,2,...,m)$. The GB membership for each sample $x \in gb_i$ is formulated as follow:
    \begin{align}\label{eq:12}
        \mu _{gb_i}(x) = \exp\left(-\frac{d(x,c_{gb_i})^2}{2 \sigma^2 r^2_{gb_i}}\right)
    \end{align}
    Where $\sigma$ is the function order, $c_{gb_i}$ is the center of $gb_i$, $r_{gb_i}$ is the radius of $gb_i$, and $d(x,c_{gb_i})$ is the distance between $x$ and $c_{gb_i}$.
\end{mydef}
When a sample is at the center of GB, its GB membership takes the maximum value of 1, and it decreases as the sample moves away from the center. In this paper, we set $\sigma = 1$, so the membership of sample near the GB radius is close to 0.5. This is related to the calculation of the GB radius using the mean value. If the GB radius calculation uses the maximum value, $\sigma$ needs to be adjusted to ensure that the membership is reasonable. Then, based on the fuzzy-rough transformation theory, we defined the shadowed GB as follow:
\begin{mydef}(Shadowed GB)\label{def:sgb}
    Let $U=\{x_1,x_2,...,x_n\}$ be a non-empty finite set, and $\mathbb{G}=\{gb_1,gb_2,...,gb_m\}$ is a granular-ball space of $U$. For each $gb_i \in \mathbb{G}(i=1,2,...,m)$, a mapping that maps GB memberships into a triplet set $\{0, [0, 1], 1\}$ can be called shadowed GB, and the mapping is formulated as follows:
    \begin{align}
        \mathbb{S}^{\alpha}_{\mu_{gb_i}}(x)=\left\{
        \begin{array}{@{}ll}
        1, & \mu_{gb_i}(x)\ge1-\alpha\\
        \mu_{gb_i}(x), & \alpha<\mu_{gb_i}(x)<1-\alpha\\
        0, &  \mu_{gb_i}(x)\le\alpha
        \end{array}
        \right.
    \end{align}
    Where $x \in gb_i$, and $(\alpha,1-\alpha)$ is the threshold pair of shadowed GB.
\end{mydef}

\begin{figure}[htbp]
    \centering
    \includegraphics[width=0.8\linewidth]{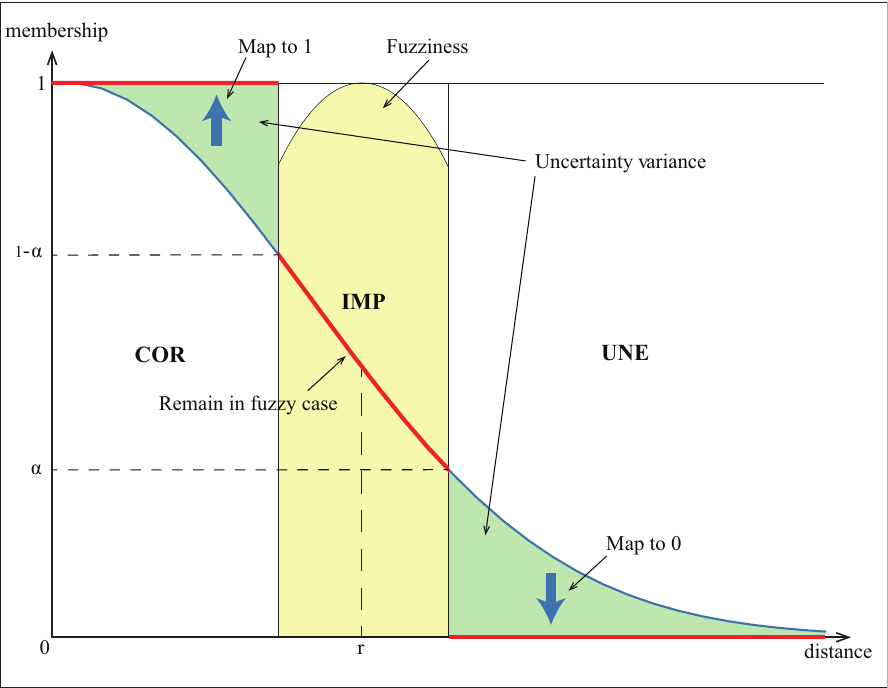}
    \caption{Transformation from GB membership to \\shadowed GB}
    \label{fig:2}
\end{figure}

The shadowed GB is shown in Fig. \ref{fig:2}. Membership lower than $\alpha$ are mapped to 0, while those higher than $1-\alpha$ are mapped to 1; the rest remain in fuzzy case. According to Definition \ref{def:3}, actions that map will cause a loss of uncertainty, which called uncertainty variance, and it can be understood as the green area in Fig. \ref{fig:2}. According to Definition \ref{def:4}, remain the GB membership in fuzzy case will result in fuzziness, and it can be understood as the yellow area in Fig. \ref{fig:2}. 

\begin{figure}[htbp]
    \centering
    \includegraphics[width=0.8\linewidth]{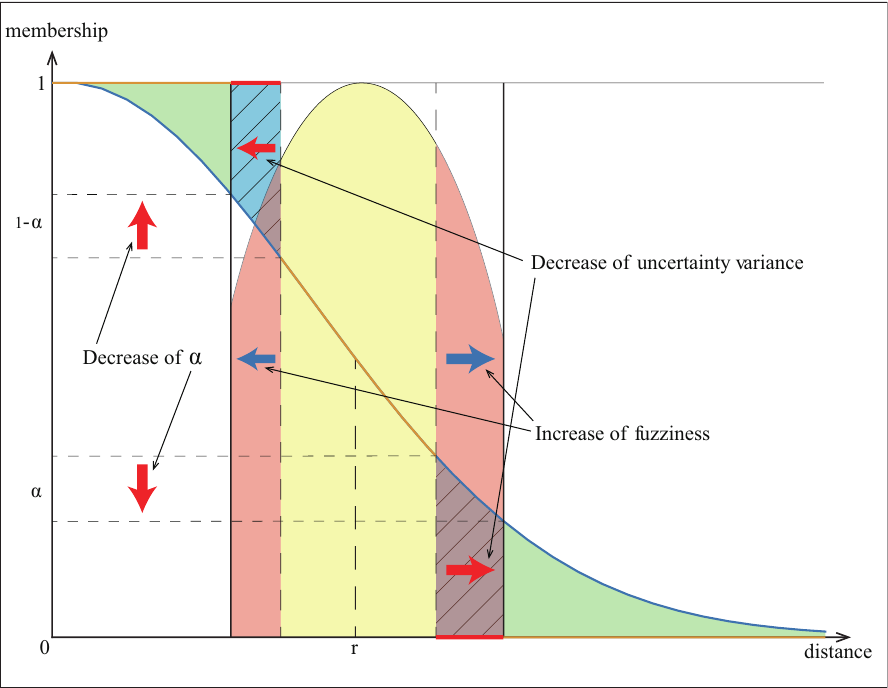}
    \caption{The changes of uncertainty variance and fuzziness when $\alpha$ changes}
    \label{fig:3}
\end{figure}

We analyzed the changes of uncertainty variance and fuzziness with the changing of $\alpha$, as shown in Fig. \ref{fig:3}. When $\alpha$ decreases, fewer GB memberships are mapped to certain negative and certain positive regions, meaning uncertainty variance decreases. Conversely, more GB memberships remain in fuzzy case, resulting in an increase in fuzziness. This indicates that uncertainty variance and fuzziness are conflicting criteria. We utilize GB membership to distinguish the contribution of samples to their GBs. However, if the uncertainty variance is excessively high, it will cause many samples to collapse into binary states, rendering the GB membership ineffective. Thus, it is crucial to keep the uncertainty variance as low as possible. Additionally, we employ shadowed GB to expedite further computations. However, if the fuzziness is too high, it will result in an excessive number of fuzzy memberships that must be processed, defeating the purpose of using shadowed GB. Therefore, it is essential to minimize fuzziness to maintain the efficiency of the shadowed GB. We need to search for a balance point between uncertainty variance and fuzziness. To achieve this, we defined the optimal $\alpha$ as follow:
\begin{mydef}(The Optimal $\alpha$)
    Let $U=\{x_1,x_2,...,x_n\}$ be a non-empty finite set, and $\mathbb{G}=\{gb_1,gb_2,...,gb_m\}$ is a granular-ball space of $U$. For each $gb_i \in \mathbb{G}(i=1,2,...,m)$, $gb_i=\{x_{i1},x_{i2},...,x_{i|gb_i|}\}$, uses $(\alpha,1-\alpha)$ as threshold pair to build shadowed GB. The measure of uncertainty variance and fuzziness in that building is defined as follow: 
    \begin{align}\label{eq:13}
        f(\alpha)=&\left(\sum_{\mu_{gb_i}(x_{ij})\le\alpha}\mu_{gb_i}(x_{ij})+\sum_{\mu_{gb_i}(x_{ij})\ge1-\alpha}(1-\mu_{gb_i}(x_{ij}))\right)\notag\\
        &\oplus\sum_{\alpha<\mu_{gb_i}(x_{ij})<1-\alpha}4\mu_{gb_i}(x_{ij})(1-\mu_{gb_i}(x_{ij})) 
    \end{align}
    Where $\oplus$ denotes the normalized and weighted combination of uncertainty variance and fuzziness. So the optimal $\alpha$ can be defined as: 
    \begin{align}
        \arg \min_{\alpha} f(\alpha)
    \end{align}
\end{mydef}

In formula (\ref{eq:13}), the left side of $\oplus$ represents to uncertainty variance, and the right side represents to fuzziness. We use the normalized and weighted combination to ensure that both criteria contribute appropriately to the overall function. The details on combining the two criteria and calculate the optimal $\alpha$ can be found in Section S3 of supplementary file.

Based on the threshold pair ($\alpha,1-\alpha$), the shadowed GB can be divided into three regions by Definition \ref{def:5}. To make it easier to understand, we renamed the three regions as: Core region (COR), Important region (IMP), and Unessential region (UNE). For $gb_i$, its regions are defined as follow:
\begin{align}\label{eq:14}
    \text{COR}_{gb_i} &= \{x\mid \mu_{gb_i}(x) \ge 1-\alpha\} \notag \\
    \text{IMP}_{gb_i} &= \{x\mid \alpha<\mu_{gb_i}(x) <1-\alpha\}\\
    \text{UNE}_{gb_i} &= \{x\mid \mu_{gb_i}(x) \le \alpha\}\notag
\end{align}

According to formula (\ref{eq:14}), samples located in COR and UNE are certain samples, while samples in IMP are uncertain samples. Based on this, we attempt to establish three-way classification rules based on the Shadowed GBs. We adopt different classification strategies for different type of regions. Predicted samples often lie under overlaps of various shadowed GBs. Therefore, we need a set of priority rules to omit irrelevant regions. The relevant regions that play an important role in influencing the sample prediction are called controlled regions in this paper. We set a predicted sample $\xi$, a GB space $\mathbb{G}=\{gb_1,gb_2,...,gb_m\}$, the GB memberships of $\xi$ to each GB can be calculated as $\mathbb{M}=\{\mu_{gb_1}(\xi),\mu_{gb_2}(\xi),...,\mu_{gb_m}(\xi)\}$. For $\mu_{gb_i}(\xi)\in\mathbb{M}$, the controlled regions of $\xi$ are calculated as follows:
\begin{itemize}
    \item[(1)] If $\exists \mu_{gb_i}(\xi)\ge 1-\alpha$, the controlled region of $\xi$ is $\text{COR}_{gb_i}$, and the label set $\mathbb{COR}=\{l_{gb_i}|\mu_{gb_i}(\xi)\ge 1-\alpha\}$ only contains distinct labels.
    \item[(2)] If $\forall \mu_{gb_i}(\xi)< 1-\alpha$ and $\exists \mu_{gb_i}(\xi)> \alpha$, the controlled region of $\xi$ is $\text{IMP}_{gb_i}$, and the label set $\mathbb{IMP}=\{(l_{gb_i},\sum_{\mu_{gb_j}(\xi)>\alpha,l_{gb_j}=l_{gb_i}}\mu_{gb_j}(\xi))|\mu_{gb_i}(\xi) >\alpha\}$ only contains distinct labels.
    \item[(3)] If $\forall \mu_{gb_i}(\xi)\le\alpha$, where $\mu_{gb_i}(\xi)\in\mathbb{M}$, the controlled region is $\text{UNE}_{gb_i}$.
\end{itemize}

After finding the labels of controlled regions, if they are completely consistent, e.g., $|\mathbb{COR}|=1$, there is no reason to proceed with further computation. The sample will be classified with the same label of controlled regions. However, when facing inconsistent labels from controlled regions, we need different principles based on the type of controlled region as follow:
\begin{itemize}
    \item[(1)] If $|\mathbb{COR}|>1$, it means the sample belongs to a high-risk classification category, and we should delay the classification of this sample. In this paper, delay the classification equal to classify it into uncertain case.
    \item[(2)] If $|\mathbb{IMP}|>1$, we need calculate the $proportion(l_{gb_i})=\frac{\mu_i}{\sum_{(l,\mu)\in \mathbb{IMP}}\mu}, (l_{gb_i},\mu_i)\in \mathbb{IMP}$. If $proportion(l_{gb_i})> 50\%$, it means that the labels of controlled regions are relatively consistent, and the sample can be classified with $l_{gb_i}$. Otherwise, the sample has a high classification risk, and we should classify it into uncertain case.
    \item[(3)] If the type is UNE, indicating that the sample does not belong to any shadowed GB. In other words, there are no references in the training data to classify this sample. This suggests a high classification risk, and we should classify it into uncertain case.
\end{itemize}

\begin{figure}[htbp]
    \centering
    \includegraphics[width=0.9\linewidth]{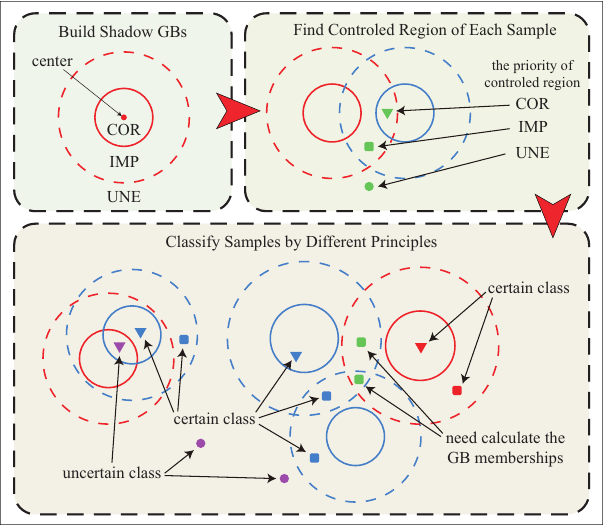}
    \caption{Three-way classification based on shadowed GBs}
    \label{fig:4}
\end{figure}

The overall process is illustrated in Fig. \ref{fig:4}. When faced with samples that need classification, we follow these three steps:
\begin{itemize}[left=27pt]
    \item[\textbf{Step.1}:]Build the shadowed GBs by fuzzy-rough transformation according to $(\alpha,1-\alpha)$.
    \item[\textbf{Step.2}:]Find the controlled regions that impact each sample.
    \item[\textbf{Step.3}:]Classify the samples to certain class or uncertain case by our proposed principles.
\end{itemize}

Using 3WD with shadowed GBs for classification can reduce classification risk by classify the sample to certain class and uncertainty case. The algorithm of 3WC-SGB can be found in Algorithm S2 of supplementary file, and time complexity analysis can be found in Section S3 of supplementary file.

\section{Experimental Studies}
\label{sec:5}
In this section, we comprehensively compared our proposed 3WC-SGB with other three-way classifiers (3WC-FNC \cite{44}, 3WC-SVM \cite{45}), state-of-the-art GB-based classifiers (GBKNN \cite{15}, GBG++ \cite{18}, ACCGBKNN \cite{36}) and traditional classifiers (Decision Tree, Naive Bayes, SVM), respectively. The detailed introduction of these compared classifiers are provided in Section S4 of supplementary file. The experiment consists of the following four parts:

\begin{itemize}
    \item Comparison with three-way classifiers to verify the advantages of applying shadowed GBs to 3WC-SGB.
    \item Comparison with state-of-the-art GB-based classifiers to demonstrate the advantages of applying three-way classification to 3WC-SGB.
    \item Comparison with traditional classifiers and conducted a comprehensive statistical analysis on all compared classifiers to demonstrate the significant advantages of 3WC-SGB.
    \item Investigated the hyperparameter configuration of 3WCSGB
    and proposed a computational method for recommending
    optimal hyperparameter.
\end{itemize}

All experiments were conducted on an Intel i5-13600kf CPU, with 16GB of DDR4 running memory, running Windows 10 64-bit OS, and utilizing Python 3.11 as the programming software. For all the tests in the experiment, tenfold cross validation was performed on each dataset. Experiments were performed on twelve UCI datasets (\href{https://archive.ics.uci.edu/datasets}{\textit{https://archive.ics.uci.edu/datasets}}), as listed in Table \ref{tab:1}.

\begin{table}[htbp]
    \caption{Experimental datasets}
    \label{tab:1}
    \centering
    \resizebox{\linewidth}{!}{
        \begin{tabular}{cccccc}
            \toprule
            No. & Datesets & Feature & Instances & Class Ratio & Type \\
            \midrule
            1 & SPECTF & 44 & 269 & 20\% vs. 80\% & Integer \\
            2 & Fourclass & 2 & 862 & 64\% vs. 36\% & Integer \\
            3 & Endgame & 9 & 958 & 65\% vs. 35\% & Integer \\
            4 & Cloud & 9 & 1024 & 2\% vs. 98\% & Integer,Real \\
            5 & Banknote Authentication & 4 & 1372 & 44\% vs. 56\% & Real \\
            6 & NHANES Age Prediction & 8 & 2278 & 16\% vs. 84\% & Integer,Real \\
            7 & Segment & 19 & 2310 & 14\% vs. 86\% & Integer,Real \\
            8 & Shill Bidding & 9 & 6321 & 11\% vs. 89\% & Integer,Real \\
            9 & Satimage & 36 & 6435 & 24\% vs. 76\% & Integer \\
            10 & Mushroom & 22 & 8124 & 52\% vs. 48\% & Integer \\
            11 & Elect & 13 & 10000 & 36\% vs. 64\% & Real \\
            12 & Online Shoppers Intention & 16 & 12330 & 15\% vs. 85\% & Integer,Real \\
            \bottomrule
        \end{tabular}
    }
\end{table}

Unlike conventional classification methods, 3WC-SGB categorizes data instances into certain classes (e.g., positive and negative in binary classification problems) and uncertain case, which helps mitigate classification risks. Based on this, we adopt the traditional 3WD to evaluate the performance of different classifiers and used six statistics namely $Accuracy$, $Precision$, $Recall$, $\mathit{F1}\ Score$, the ratio of uncertain instances ($UR$), and average cost ($Cost$). Additionally, we conducted statistical analysis including $Win/Loss$, $Rank$, and $P\text{-}value$. The detailed introduction of these statistics are provided in Section S4 of supplementary file.

To validate the robustness of 3WC, noise will be added to the training data by flipping the labels, where the ratio of flipping is the noise rate.

\begin{figure*}[htbp]
    \centering
    \resizebox{\textwidth}{!}{
        \begin{subfigure}[b]{0.25\textwidth}
            \centering
            \includegraphics[width=\textwidth]{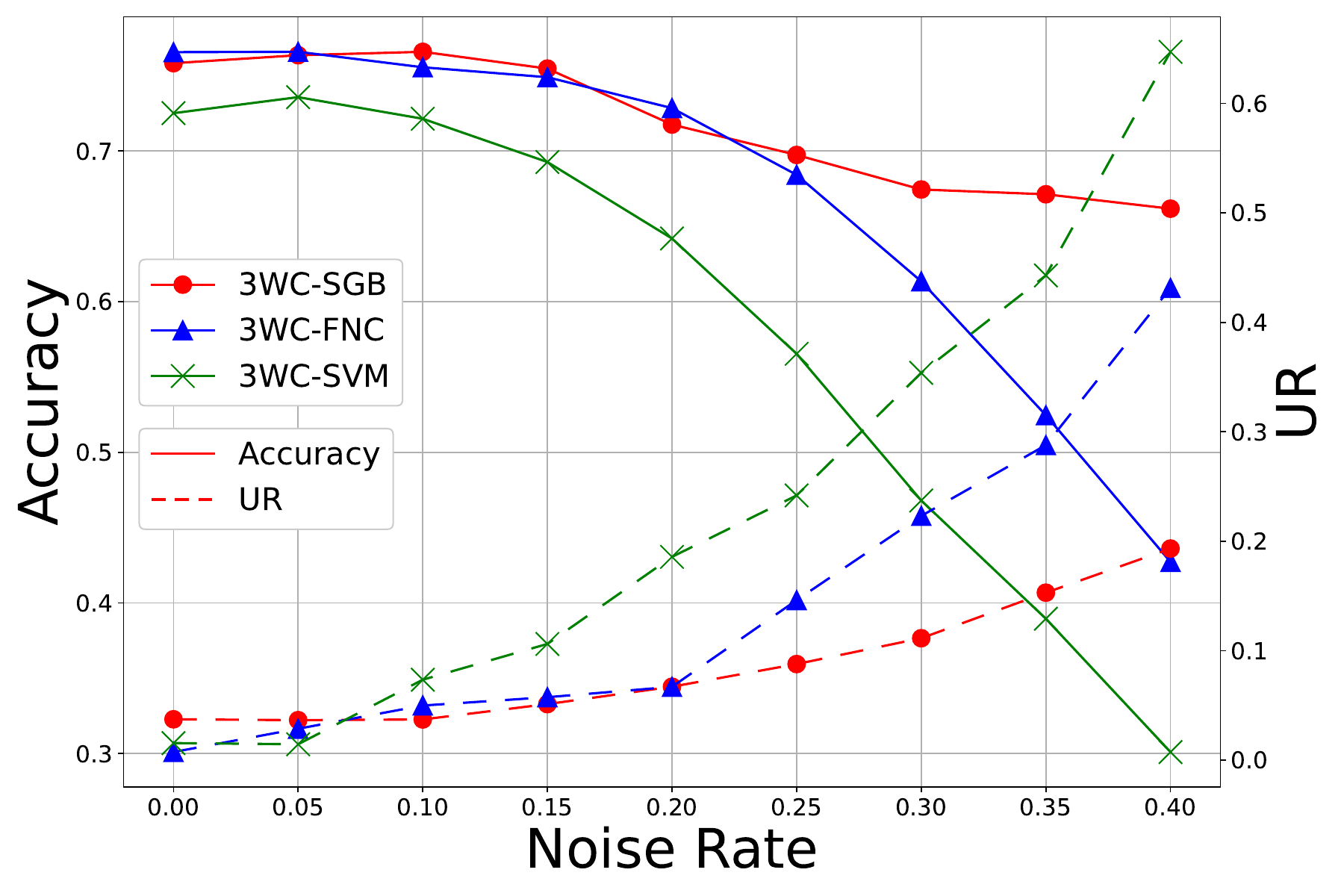}
            \caption{SPECTF}
            \label{fig:5:1}
        \end{subfigure}
        \hfill
        \begin{subfigure}[b]{0.25\textwidth}
            \centering
            \includegraphics[width=\textwidth]{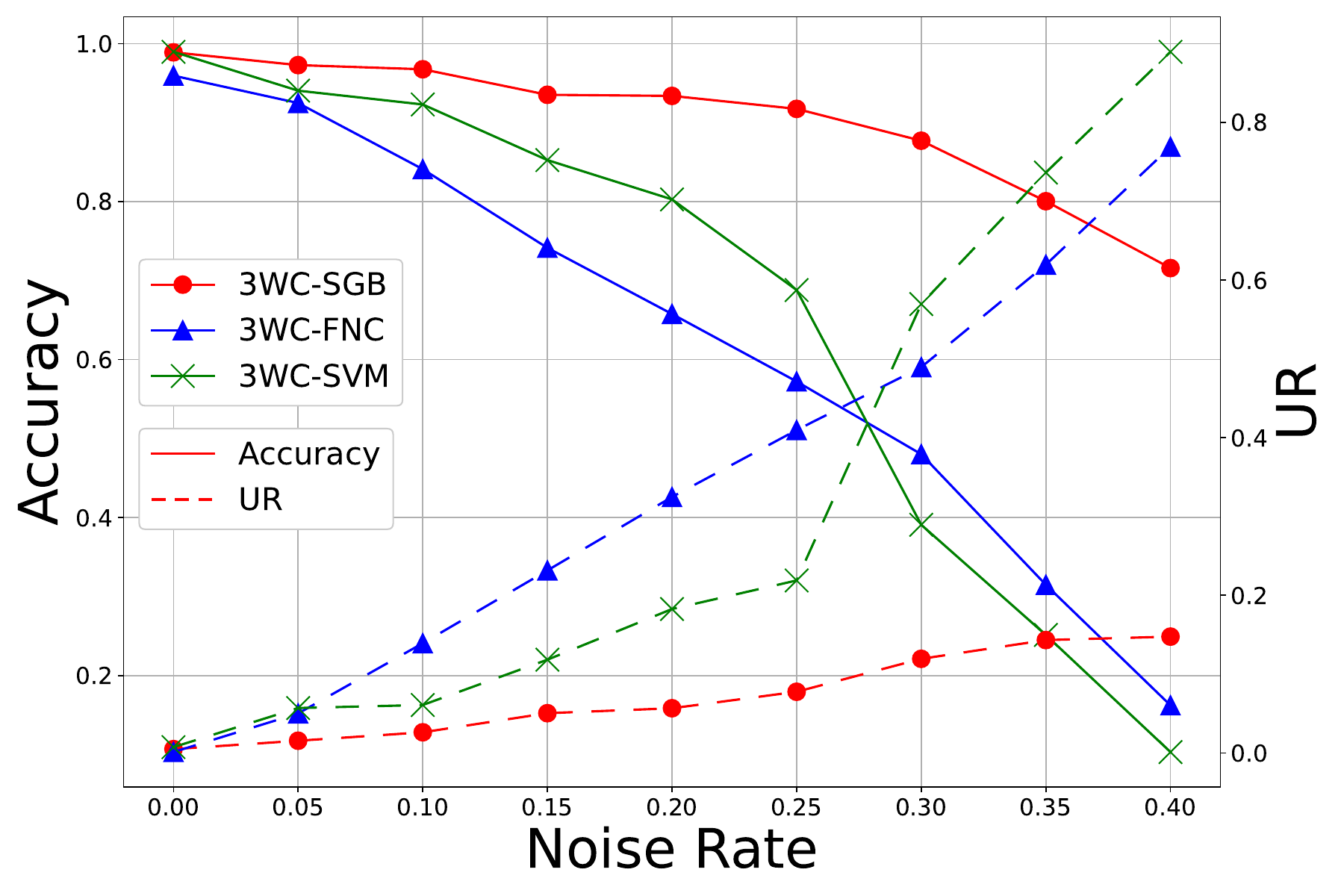}
            \caption{Fourclass}
            \label{fig:5:2}
        \end{subfigure}
        \hfill
        \begin{subfigure}[b]{0.25\textwidth}
            \centering
            \includegraphics[width=\textwidth]{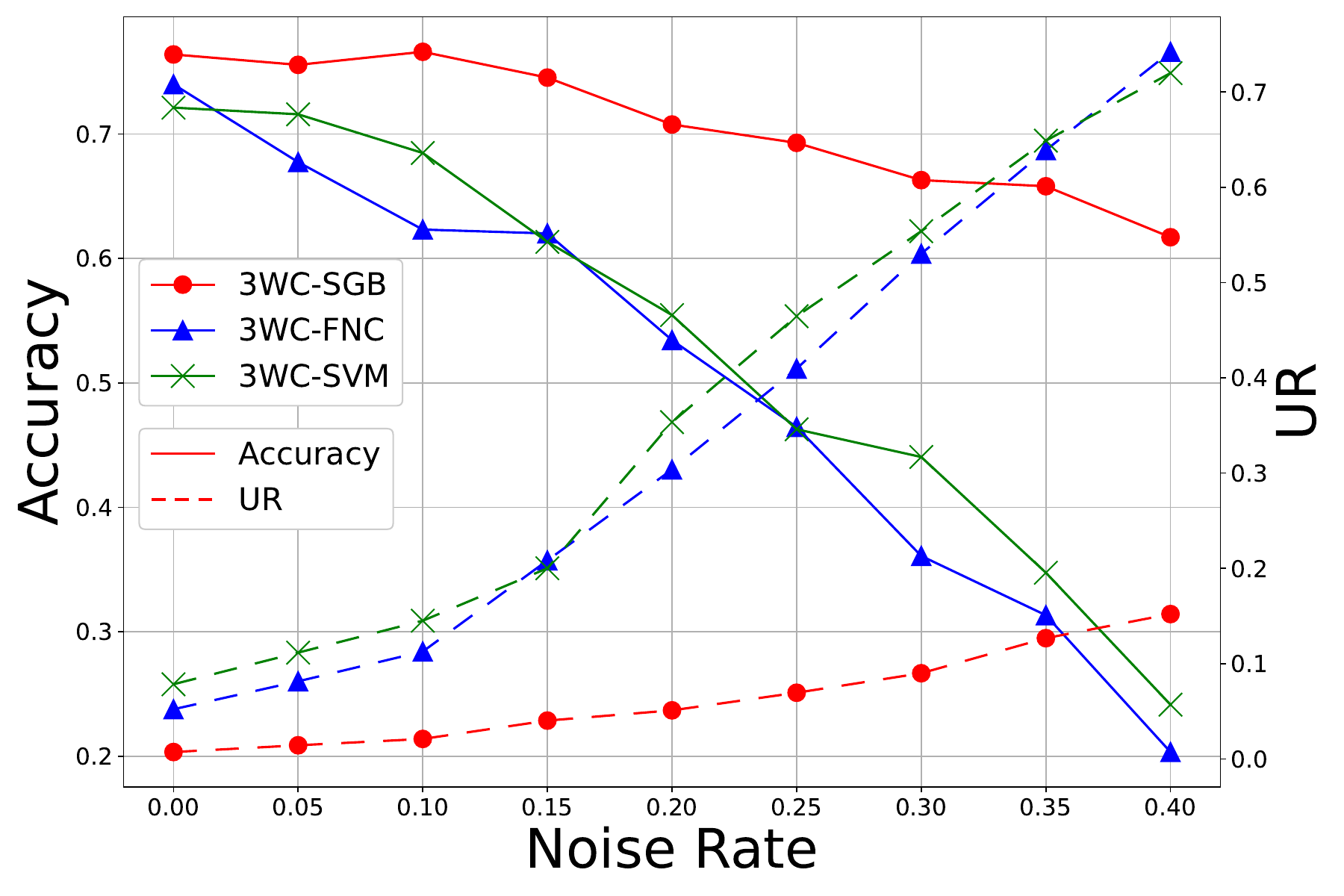}
            \caption{Endgame}
            \label{fig:5:3}
        \end{subfigure}
        \hfill
        \begin{subfigure}[b]{0.25\textwidth}
            \centering
            \includegraphics[width=\textwidth]{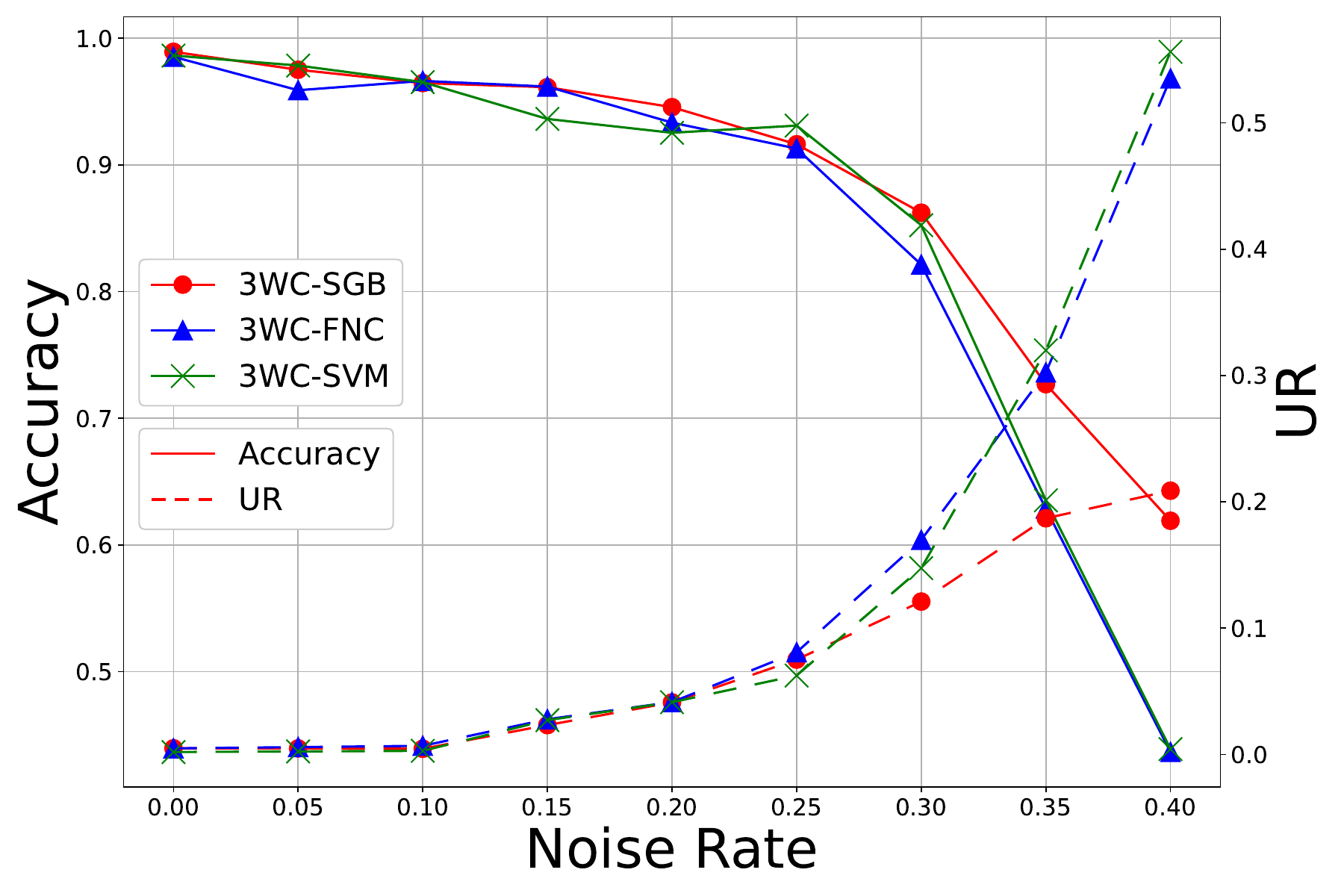}
            \caption{Cloud}
            \label{fig:5:4}
        \end{subfigure}
    }

    \medskip

    \resizebox{\textwidth}{!}{
        \begin{subfigure}[b]{0.25\textwidth}
            \centering
            \includegraphics[width=\textwidth]{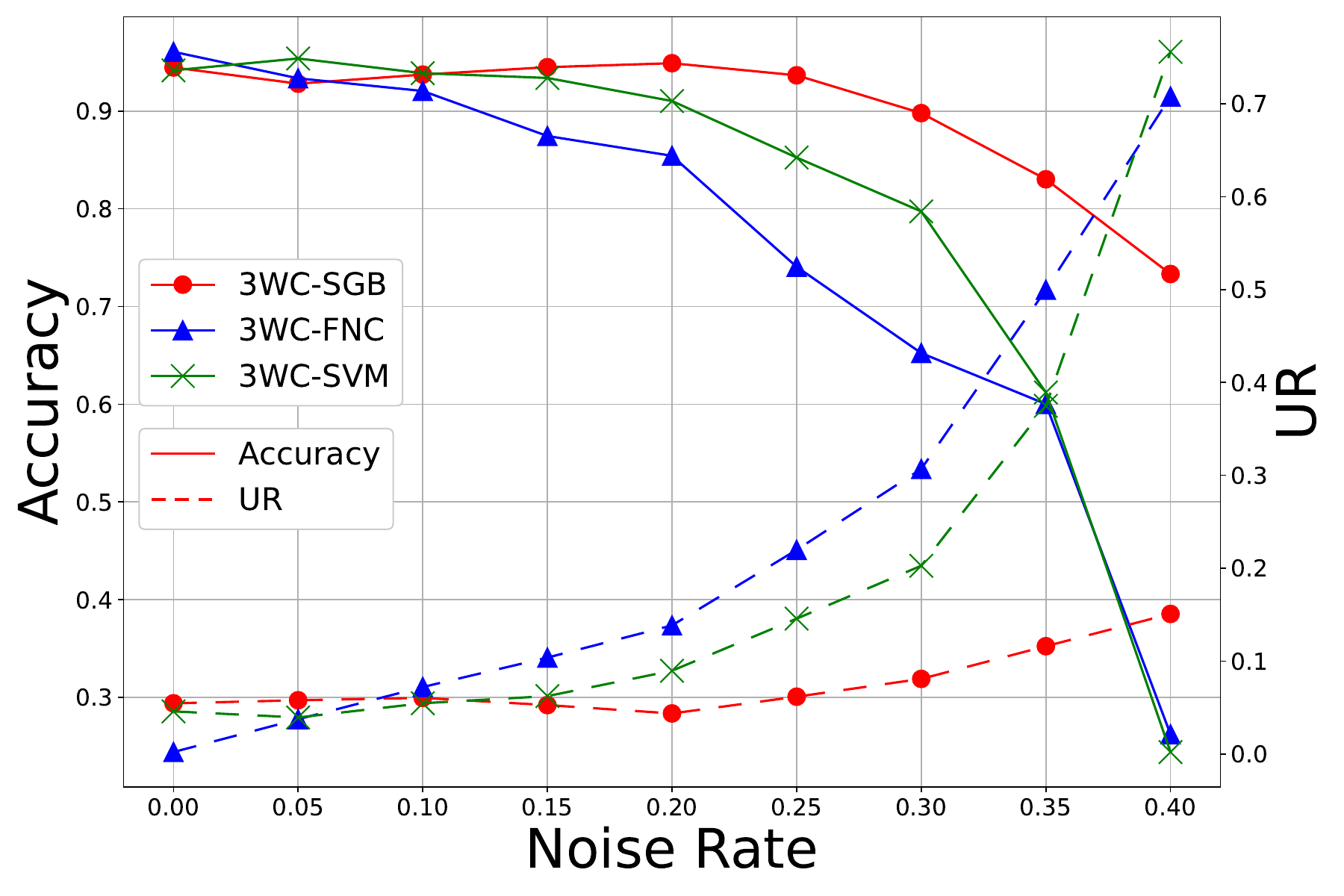}
            \caption{Banknote Authentication}
            \label{fig:5:5}
        \end{subfigure}
        \hfill
        \begin{subfigure}[b]{0.25\textwidth}
            \centering
            \includegraphics[width=\textwidth]{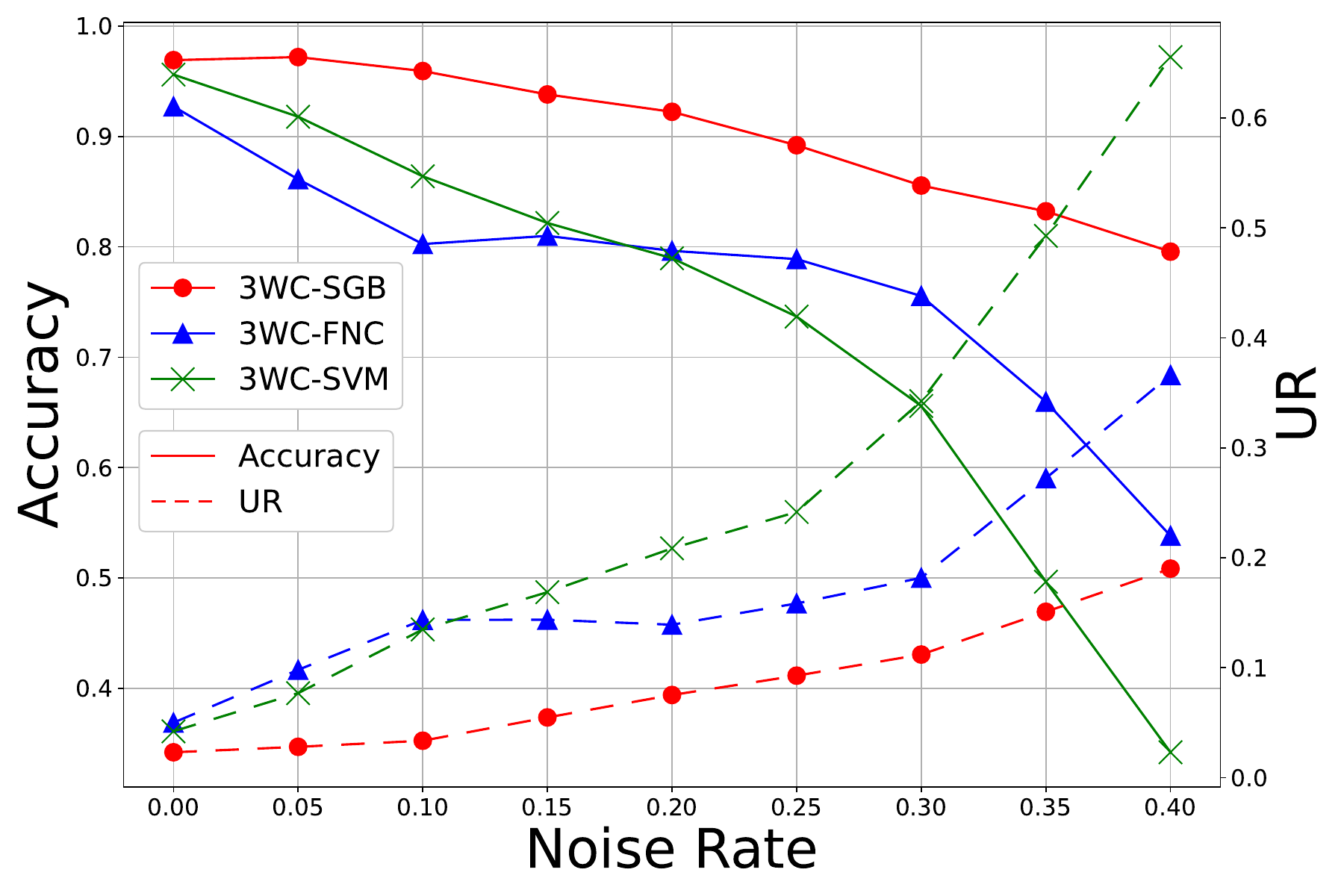}
            \caption{NHANES Age Prediction}
            \label{fig:5:6}
        \end{subfigure}
        \hfill
        \begin{subfigure}[b]{0.25\textwidth}
            \centering
            \includegraphics[width=\textwidth]{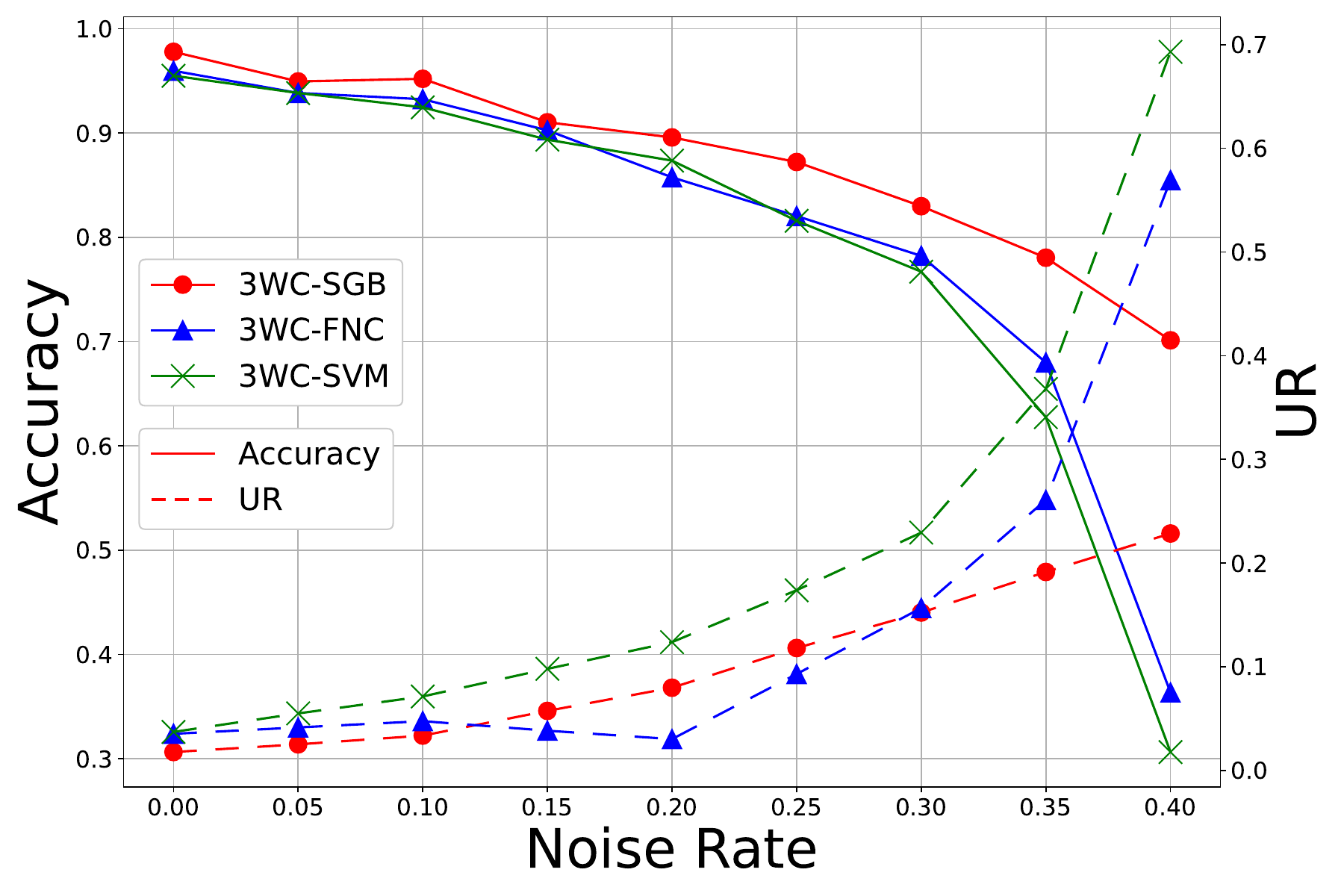}
            \caption{Segment}
            \label{fig:5:7}
        \end{subfigure}
        \hfill
        \begin{subfigure}[b]{0.25\textwidth}
            \centering
            \includegraphics[width=\textwidth]{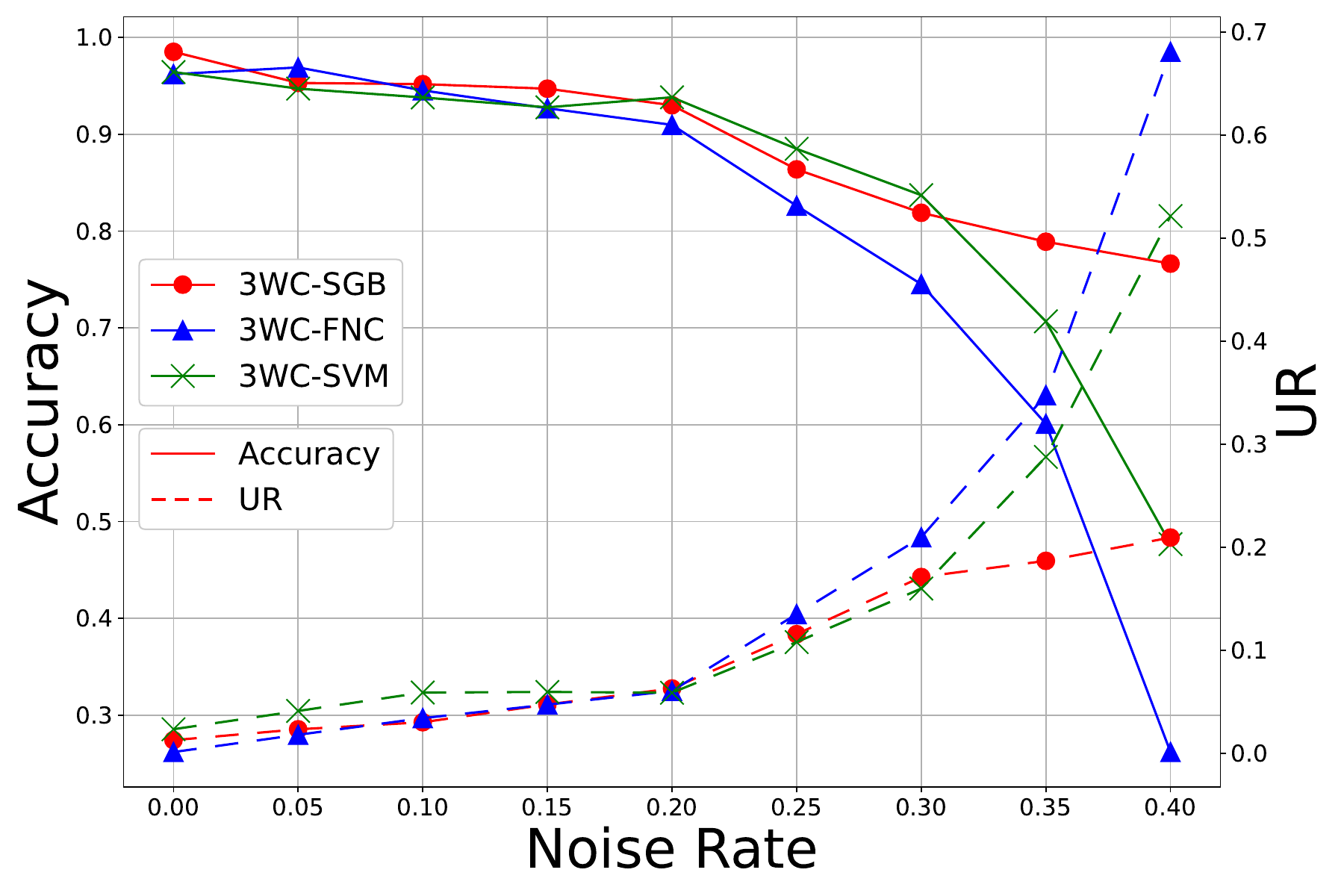}
            \caption{Shill Bidding}
            \label{fig:5:8}
        \end{subfigure}
    }

    \medskip

    \resizebox{\textwidth}{!}{
        \begin{subfigure}[b]{0.25\textwidth}
            \centering
            \includegraphics[width=\textwidth]{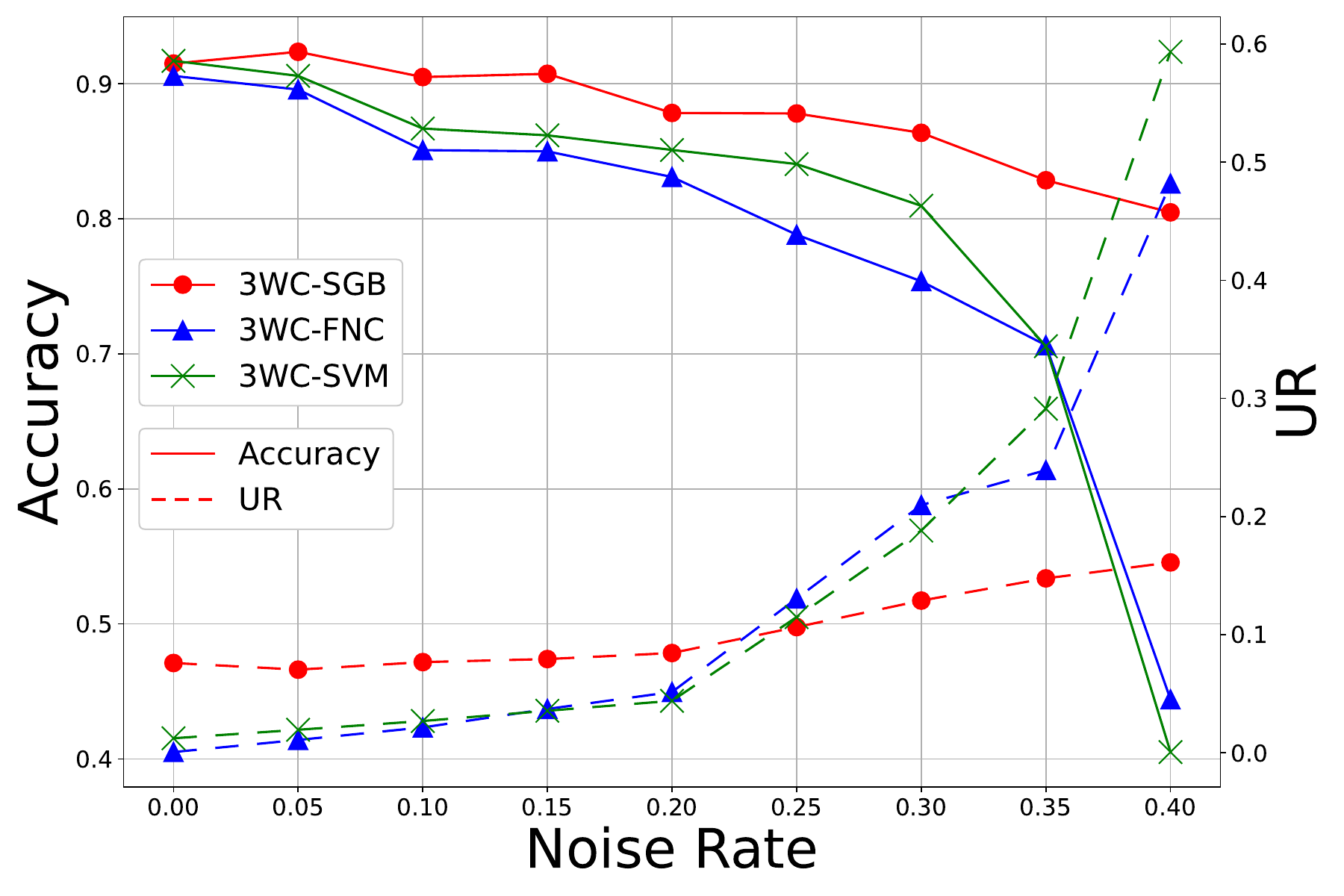}
            \caption{Satimage}
            \label{fig:5:9}
        \end{subfigure}
        \hfill
        \begin{subfigure}[b]{0.25\textwidth}
            \centering
            \includegraphics[width=\textwidth]{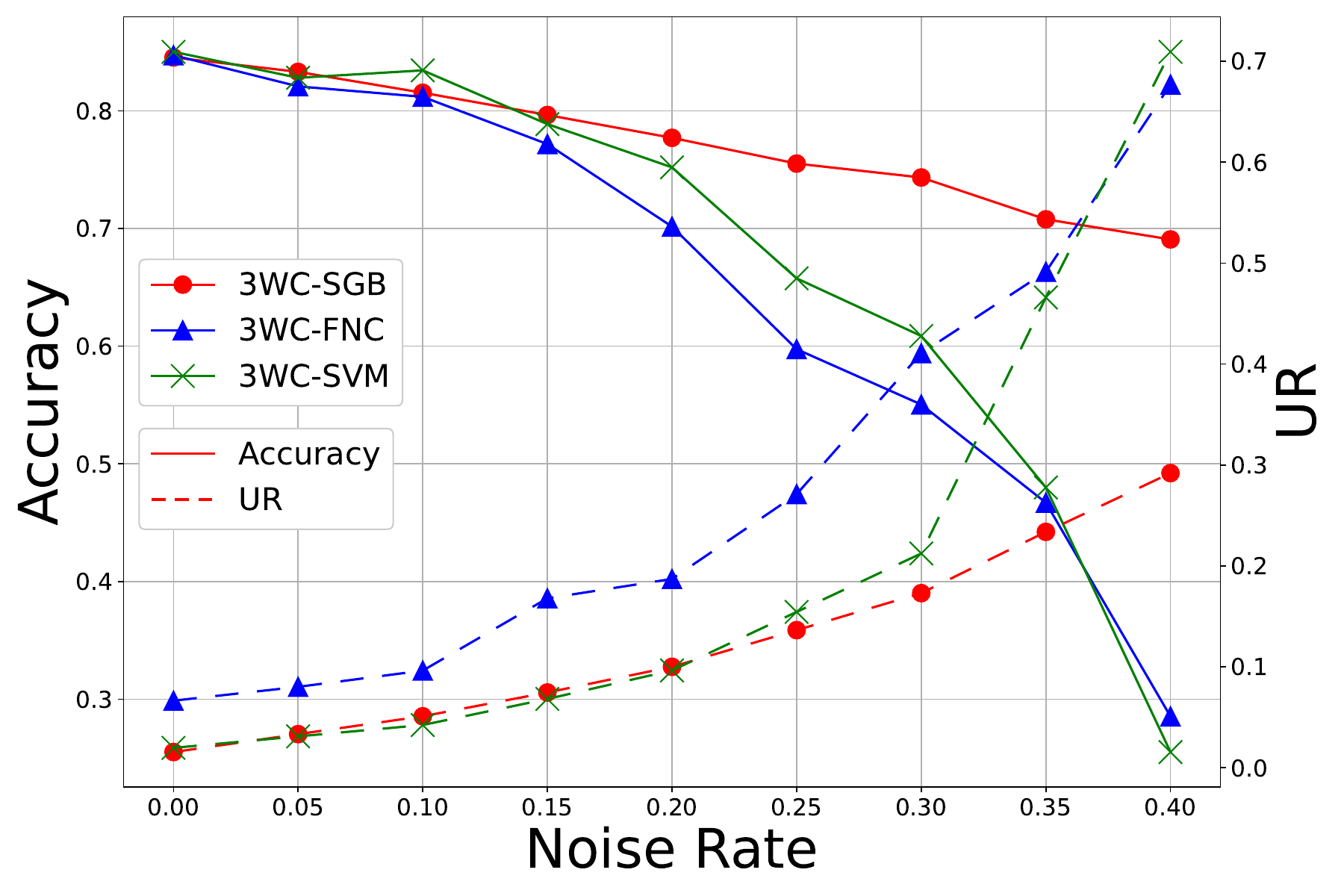}
            \caption{mushroom}
            \label{fig:5:10}
        \end{subfigure}
        \hfill
        \begin{subfigure}[b]{0.25\textwidth}
            \centering
            \includegraphics[width=\textwidth]{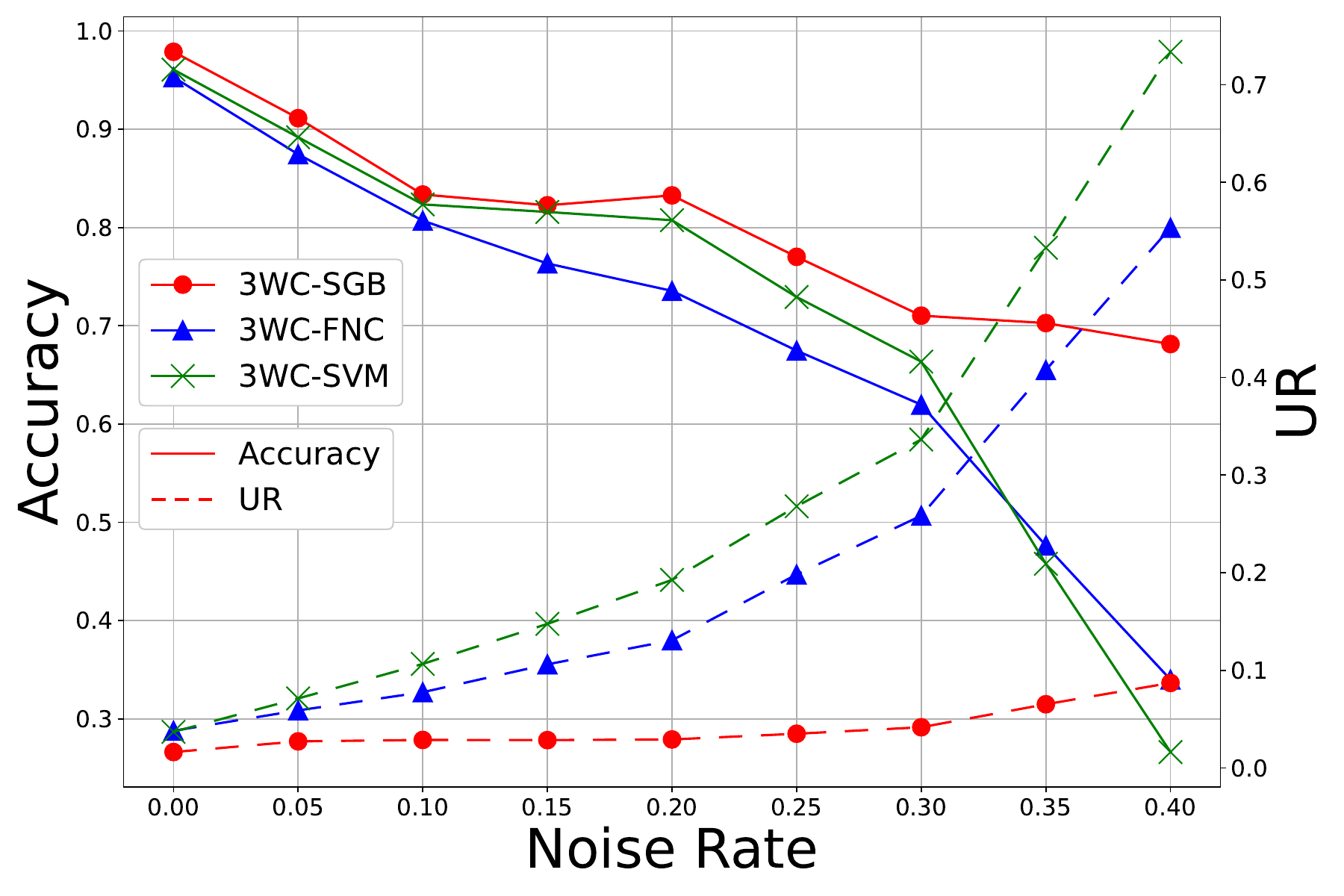}
            \caption{Elect}
            \label{fig:5:11}
        \end{subfigure}
        \hfill
        \begin{subfigure}[b]{0.25\textwidth}
            \centering
            \includegraphics[width=\textwidth]{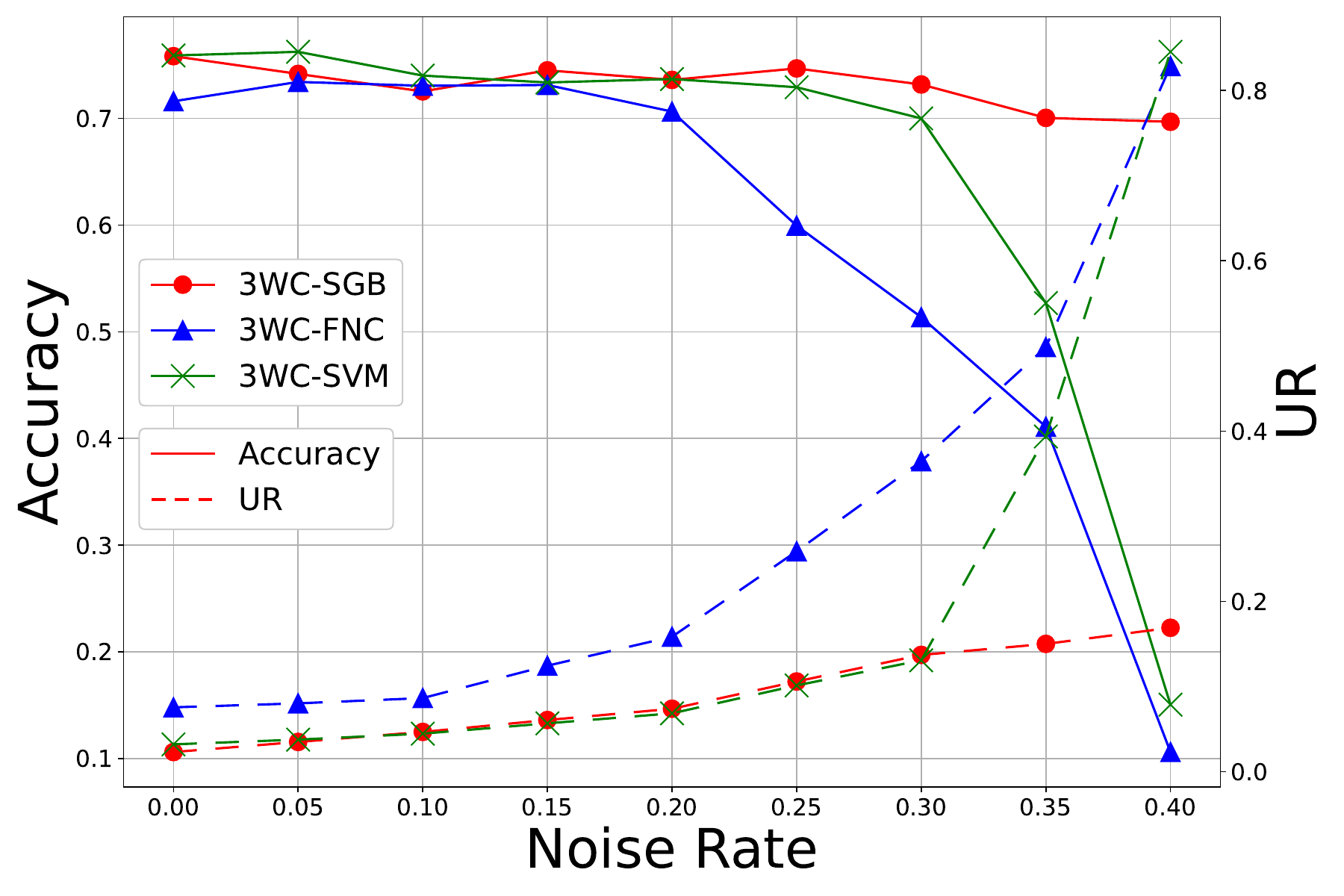}
            \caption{Online Shoppers Intention}
            \label{fig:5:12}
        \end{subfigure}
    }
    \captionsetup{justification=centering}
    \caption{The comparison with three-way classifiers under different noise rates}
    \label{fig:5}
\end{figure*}

\begin{table*}[htbp]
    \caption{The statistical analysis of GB-based classifiers under different noise rates}
    \label{tab:2}
    \centering
    \resizebox{\textwidth}{!}{
        \begin{tabular}{cc*{16}{>{\centering\arraybackslash}p{1.1cm}}}
        \toprule
        \multirow{3}{*}[-2pt]{\centering No.} & \multirow{3}{*}[-2pt]{\centering Metrics} & 
        \multicolumn{4}{c}{Noise Rate: 0.1}& \multicolumn{4}{c}{Noise Rate: 0.2} &
        \multicolumn{4}{c}{Noise Rate: 0.3}& \multicolumn{4}{c}{Noise Rate: 0.4} \\
        \cmidrule(lr){3-6} \cmidrule(lr){7-10} \cmidrule(lr){11-14} \cmidrule(lr){15-18}
        && 3WC-SGB & \multirow{2}{*}{GBKNN} & \multirow{2}{*}{GBG++} & ACC-GBKNN
        & 3WC-SGB & \multirow{2}{*}{GBKNN} & \multirow{2}{*}{GBG++} & ACC-GBKNN
        & 3WC-SGB & \multirow{2}{*}{GBKNN} & \multirow{2}{*}{GBG++} & ACC-GBKNN
        & 3WC-SGB & \multirow{2}{*}{GBKNN} & \multirow{2}{*}{GBG++} & ACC-GBKNN \\
        \midrule
        \multirow{5}{*}{1}
        & $Accuracy$ &
        0.7658 &0.7381 &0.7288 &0.6840 &0.7175 &0.6731 &0.6615 &0.6466 &0.6744 &0.6324 &0.6168 &0.6058 &0.6617 &0.5802 &0.6193 &0.5724 \\
        & $Precision$ &
        0.4000 &0.5324 &0.3663 &0.3238 &0.3500 &0.5750 &0.2918 &0.3089 &0.3933 &0.2631 &0.2794 &0.2951 &0.3385 &0.2544 &0.3263 &0.2386 \\
        & $Recall$ &
        0.3733 &0.4845 &0.4100 &0.5500 &0.3463 &0.4775 &0.3833 &0.6033 &0.2333 &0.2394 &0.5133 &0.6533 &0.2767 &0.3419 &0.6467 &0.5100 \\
        & $\mathit{F1}\ Sorce$ &
        0.3862 &0.5073 &0.3869 &0.4076 &0.3481 &0.5217 &0.3313 &0.4086 &0.2929 &0.2507 &0.3619 &0.4065 &0.3045 &0.2917 &0.4337 &0.3251 \\
        & $Cost$ &
        2.8772 &2.9369 &3.1880 &3.5174 &3.1652 &2.6999 &3.8895 &3.8464 &3.4792 &3.2998 &4.2348 &4.2248 &3.7066 &3.7341 &3.8889 &4.6769 \\
        \cmidrule(r){1-6} \cmidrule(lr){7-10} \cmidrule(lr){11-14} \cmidrule(l){15-18}
        \multirow{5}{*}{2}
        & $Accuracy$ &
        0.9675 &0.9304 &0.9780 &0.9037 &0.9338 &0.8445 &0.9269 &0.8028 &0.8771 &0.7390 &0.8353 &0.6820 &0.7159 &0.5928 &0.6751 &0.6045 \\
        & $Precision$ &
        0.9927 &0.9575 &0.9823 &0.9229 &0.9723 &0.9120 &0.9513 &0.8688 &0.9582 &0.8328 &0.8902 &0.7732 &0.8336 &0.7185 &0.7820 &0.7198 \\
        & $Recall$ &
        0.9837 &0.9333 &0.9839 &0.9295 &0.9384 &0.8396 &0.9351 &0.8180 &0.8756 &0.7441 &0.8504 &0.7188 &0.7354 &0.6036 &0.6939 &0.6305 \\
        & $\mathit{F1}\ Sorce$ &
        0.9882 &0.9452 &0.9831 &0.9262 &0.9551 &0.8743 &0.9431 &0.8426 &0.9151 &0.7860 &0.8698 &0.7450 &0.7814 &0.6560 &0.7353 &0.6722 \\
        & $Cost$ &
        0.2068 &0.8677 &0.2620 &1.1439 &0.5553 &1.9675 &0.8980 &2.4407 &1.1806 &3.2691 &2.0322 &3.9036 &3.0494 &5.0928 &4.0381 &4.9064 \\
        \cmidrule(r){1-6} \cmidrule(lr){7-10} \cmidrule(lr){11-14} \cmidrule(l){15-18}
        \multirow{5}{*}{3}
        & $Accuracy$ &
        0.7660 &0.7088 &0.7422 &0.6994 &0.7076 &0.6461 &0.6347 &0.6158 &0.6629 &0.6472 &0.6044 &0.6042 &0.6170 &0.5438 &0.5544 &0.5387 \\
        & $Precision$ &
        0.6640 &0.7177 &0.8263 &0.7875 &0.6596 &0.6910 &0.7422 &0.7113 &0.6648 &0.7080 &0.7365 &0.7143 &0.6617 &0.6818 &0.6908 &0.6607 \\
        & $Recall$ &
        0.9905 &0.9137 &0.7683 &0.7444 &0.9873 &0.8275 &0.6841 &0.6947 &0.9776 &0.7875 &0.6262 &0.6581 &0.9126 &0.5735 &0.5736 &0.6024 \\
        & $\mathit{F1}\ Sorce$ &
        0.7950 &0.8040 &0.7962 &0.7653 &0.7908 &0.7531 &0.7119 &0.7029 &0.7915 &0.7457 &0.6769 &0.6851 &0.7672 &0.6230 &0.6268 &0.6302 \\
        & $Cost$ &
        3.0529 &3.1378 &3.1835 &3.6737 &3.4509 &3.9896 &4.4788 &4.6395 &3.4293 &4.0835 &4.9332 &4.8514 &3.5467 &5.6764 &5.5706 &5.6522 \\
        \cmidrule(r){1-6} \cmidrule(lr){7-10} \cmidrule(lr){11-14} \cmidrule(l){15-18}
        \multirow{5}{*}{4}
        & $Accuracy$ &
        0.9646 &0.9834 &0.9658 &0.8633 &0.9456 &0.9395 &0.8760 &0.7423 &0.8623 &0.8604 &0.7636 &0.6543 &0.6191 &0.6162 &0.6181 &0.4745 \\
        & $Precision$ &
        0.6900 &0.4688 &0.2833 &0.1066 &0.3500 &0.1705 &0.0632 &0.0275 &0.4750 &0.3968 &0.3152 &0.2455 &0.5787 &0.2578 &0.2211 &0.1135 \\
        & $Recall$ &
        0.8500 &0.9375 &0.5000 &0.8000 &0.8000 &0.7946 &0.5500 &0.6000 &0.4000 &0.9375 &0.5000 &0.6000 &0.4500 &0.6161 &0.4500 &0.3500 \\
        & $\mathit{F1}\ Sorce$ &
        0.7617 &0.6250 &0.3617 &0.1882 &0.4870 &0.2808 &0.1133 &0.0526 &0.4343 &0.5576 &0.3866 &0.3484 &0.5063 &0.3635 &0.2965 &0.1714 \\
        & $Cost$ &
        1.1103 &0.1699 &1.3768 &1.3788 &0.6974 &0.6172 &1.2674 &2.6048 &0.8159 &1.4004 &2.3947 &3.4803 &2.0259 &3.8613 &3.8461 &5.2899 \\
        \cmidrule(r){1-6} \cmidrule(lr){7-10} \cmidrule(lr){11-14} \cmidrule(l){15-18}
        \multirow{5}{*}{5}
        & $Accuracy$ &
        0.9373 &0.9439 &0.9767 &0.8885 &0.9490 &0.8549 &0.9111 &0.7879 &0.8979 &0.7543 &0.8091 &0.6852 &0.7333 &0.6188 &0.6669 &0.5889 \\
        & $Precision$ &
        0.9964 &0.9444 &0.9745 &0.8537 &0.9958 &0.8396 &0.8821 &0.7287 &0.9808 &0.7145 &0.7756 &0.6246 &0.8440 &0.5654 &0.6082 &0.5322 \\
        & $Recall$ &
        0.9361 &0.9295 &0.9738 &0.9082 &0.9295 &0.8361 &0.9246 &0.8361 &0.8738 &0.7557 &0.8115 &0.7426 &0.7230 &0.6066 &0.7066 &0.6295 \\
        & $\mathit{F1}\ Sorce$ &
        0.9653 &0.9369 &0.9741 &0.8801 &0.9615 &0.8378 &0.9028 &0.7787 &0.9242 &0.7346 &0.7931 &0.6785 &0.7788 &0.5853 &0.6537 &0.5768 \\
        & $Cost$ &
        0.2113 &0.6869 &0.2801 &1.2786 &0.2201 &1.7423 &1.0234 &2.4123 &0.5017 &2.8918 &2.2446 &3.6061 &1.8440 &4.5123 &3.8528 &4.7699 \\
        \cmidrule(r){1-6} \cmidrule(lr){7-10} \cmidrule(lr){11-14} \cmidrule(l){15-18}
        \multirow{5}{*}{6}
        & $Accuracy$ &
        0.9593 &0.9342 &0.9434 &0.8494 &0.9224 &0.9088 &0.8635 &0.7423 &0.8556 &0.8223 &0.7555 &0.6128 &0.7957 &0.7823 &0.6453 &0.5457 \\
        & $Precision$ &
        0.9353 &0.9543 &0.7933 &0.5205 &0.9216 &0.8853 &0.5532 &0.3599 &0.8816 &0.4500 &0.3706 &0.2533 &0.6078 &0.3935 &0.2487 &0.2007 \\
        & $Recall$ &
        0.8420 &0.6264 &0.8764 &0.8486 &0.8839 &0.5797 &0.8017 &0.7661 &0.6655 &0.5220 &0.7557 &0.7277 &0.5408 &0.3929 &0.6074 &0.6156 \\
        & $\mathit{F1}\ Sorce$ &
        0.8862 &0.7563 &0.8327 &0.6452 &0.9023 &0.7006 &0.6547 &0.4897 &0.7585 &0.4833 &0.4974 &0.3758 &0.5723 &0.3932 &0.3529 &0.3027 \\
        & $Cost$ &
        0.5558 &0.8973 &0.6452 &1.6023 &0.7039 &1.0808 &1.4917 &2.7259 &0.7656 &1.9824 &2.6015 &4.0458 &1.6250 &2.5654 &3.7980 &4.7891 \\
        \cmidrule(r){1-6} \cmidrule(lr){7-10} \cmidrule(lr){11-14} \cmidrule(l){15-18}
        \multirow{5}{*}{7}
        & $Accuracy$ &
        0.9520 &0.9307 &0.9580 &0.8887 &0.8958 &0.8619 &0.8933 &0.7485 &0.8299 &0.7143 &0.7814 &0.6455 &0.7013 &0.6113 &0.6485 &0.5563 \\
        & $Precision$ &
        0.9933 &0.8613 &0.7987 &0.5686 &0.8597 &0.3056 &0.6012 &0.3423 &0.9192 &0.3949 &0.3755 &0.2496 &0.5879 &0.2359 &0.2398 &0.1888 \\
        & $Recall$ &
        0.9102 &0.6091 &0.9515 &0.9515 &0.8594 &0.2333 &0.8758 &0.8091 &0.4212 &0.5394 &0.7818 &0.7333 &0.4727 &0.5061 &0.6576 &0.6364 \\
        & $\mathit{F1}\ Sorce$ &
        0.9500 &0.7136 &0.8684 &0.7118 &0.8596 &0.2646 &0.7130 &0.4811 &0.5777 &0.4560 &0.5073 &0.3725 &0.5241 &0.3218 &0.3514 &0.2912 \\
        & $Cost$ &
        0.6693 &0.9160 &0.4476 &1.1403 &0.5723 &1.8190 &1.0883 &2.6242 &0.6944 &2.1203 &2.3108 &3.6978 &1.6078 &3.1697 &3.7108 &4.6450 \\
        \cmidrule(r){1-6} \cmidrule(lr){7-10} \cmidrule(lr){11-14} \cmidrule(l){15-18}
        \multirow{5}{*}{8}
        & $Accuracy$ &
        0.9517 &0.9658 &0.9530 &0.8711 &0.9299 &0.9015 &0.8820 &0.7469 &0.8189 &0.7587 &0.7754 &0.6494 &0.7663 &0.7105 &0.6418 &0.5621 \\
        & $Precision$ &
        0.9203 &0.8296 &0.7319 &0.4473 &0.9305 &0.6384 &0.4728 &0.2697 &0.9011 &0.4372 &0.2817 &0.1964 &0.7676 &0.2126 &0.1788 &0.1540 \\
        & $Recall$ &
        0.8498 &0.8578 &0.8963 &0.8757 &0.8073 &0.8238 &0.8296 &0.7925 &0.7824 &0.7540 &0.7037 &0.7376 &0.7423 &0.6296 &0.6532 &0.6857 \\
        & $\mathit{F1}\ Sorce$ &
        0.8837 &0.8435 &0.8058 &0.5922 &0.8645 &0.7194 &0.6023 &0.4024 &0.8376 &0.5535 &0.4023 &0.3102 &0.7547 &0.3179 &0.2808 &0.2515 \\
        & $Cost$ &
        0.7537 &0.4025 &0.5142 &1.3425 &0.6857 &0.7603 &1.2529 &2.6198 &0.7445 &1.5177 &2.3730 &3.6178 &1.0277 &3.0533 &3.7298 &4.5132 \\
        \cmidrule(r){1-6} \cmidrule(lr){7-10} \cmidrule(lr){11-14} \cmidrule(l){15-18}
        \multirow{5}{*}{9}
        & $Accuracy$ &
        0.9050 &0.9221 &0.9378 &0.9088 &0.8784 &0.8582 &0.8553 &0.8162 &0.8637 &0.7900 &0.7358 &0.6962 &0.8048 &0.7023 &0.6196 &0.5859 \\
        & $Precision$ &
        0.9890 &0.9252 &0.8289 &0.7447 &0.9848 &0.8023 &0.6539 &0.5768 &0.9839 &0.7519 &0.4661 &0.4252 &0.8861 &0.5777 &0.3380 &0.3239 \\
        & $Recall$ &
        0.8250 &0.9211 &0.9347 &0.9459 &0.6596 &0.8250 &0.8369 &0.8722 &0.7908 &0.8036 &0.7280 &0.7684 &0.5591 &0.6999 &0.6223 &0.6796 \\
        & $\mathit{F1}\ Sorce$ &
        0.8996 &0.9231 &0.8787 &0.8333 &0.7900 &0.8135 &0.7342 &0.6944 &0.8769 &0.7769 &0.5683 &0.5475 &0.6856 &0.6330 &0.4380 &0.4387 \\
        & $Cost$ &
        0.4917 &0.6544 &0.6838 &0.9639 &0.4768 &0.4895 &1.6022 &1.9602 &0.4610 &1.2873 &2.9011 &3.2589 &0.8635 &2.2626 &4.1642 &4.4467 \\
        \cmidrule(r){1-6} \cmidrule(lr){7-10} \cmidrule(lr){11-14} \cmidrule(l){15-18}
        \multirow{5}{*}{10}
        & $Accuracy$ &
        0.8153 &0.7900 &0.8503 &0.8886 &0.7770 &0.7480 &0.7729 &0.7816 &0.7433 &0.6949 &0.7275 &0.6945 &0.6907 &0.6657 &0.6670 &0.5960 \\
        & $Precision$ &
        0.9956 &0.8606 &0.9478 &0.8711 &1.0000 &0.9009 &0.8699 &0.7716 &0.9901 &0.8521 &0.7810 &0.6920 &0.9590 &0.7163 &0.6929 &0.6026 \\
        & $Recall$ &
        0.6206 &0.7619 &0.8622 &0.9256 &0.6601 &0.5803 &0.7962 &0.8258 &0.6869 &0.4974 &0.7697 &0.7367 &0.6218 &0.5741 &0.6490 &0.6459 \\
        & $\mathit{F1}\ Sorce$ &
        0.7646 &0.8082 &0.9030 &0.8975 &0.7952 &0.7059 &0.8314 &0.7978 &0.8111 &0.6281 &0.7753 &0.7137 &0.7545 &0.6374 &0.6702 &0.6235 \\
        & $Cost$ &
        0.7508 &2.5933 &0.5756 &1.2680 &0.7411 &3.3892 &1.4867 &2.5445 &0.8081 &4.0928 &2.8023 &3.6006 &1.0077 &4.2255 &4.0571 &4.7734 \\
        \cmidrule(r){1-6} \cmidrule(lr){7-10} \cmidrule(lr){11-14} \cmidrule(l){15-18}
        \multirow{5}{*}{11}
        & $Accuracy$ &
        0.8336 &0.8229 &0.8133 &0.7838 &0.8327 &0.8295 &0.7441 &0.7027 &0.7103 &0.6805 &0.6665 &0.6246 &0.6814 &0.6670 &0.5803 &0.5470 \\
        & $Precision$ &
        0.7436 &0.8432 &0.7215 &0.6699 &0.7440 &0.7736 &0.6261 &0.5683 &0.7381 &0.6359 &0.5327 &0.4869 &0.6742 &0.5353 &0.4368 &0.4183 \\
        & $Recall$ &
        0.7276 &0.6378 &0.7906 &0.7945 &0.7931 &0.7641 &0.7293 &0.7450 &0.4052 &0.7262 &0.6481 &0.6812 &0.3453 &0.5818 &0.5492 &0.6436 \\
        & $\mathit{F1}\ Sorce$ &
        0.7355 &0.7263 &0.7545 &0.7269 &0.7678 &0.7688 &0.6738 &0.6448 &0.5232 &0.6781 &0.5848 &0.5679 &0.4567 &0.5576 &0.4866 &0.5071 \\
        & $Cost$ &
        2.0176 &2.2954 &2.1702 &2.4596 &2.1868 &2.0466 &2.9510 &3.3422 &2.5158 &2.8914 &3.8446 &4.2156 &3.8334 &3.9356 &4.8498 &5.0460 \\
        \cmidrule(r){1-6} \cmidrule(lr){7-10} \cmidrule(lr){11-14} \cmidrule(l){15-18}
        \multirow{5}{*}{12}
        & $Accuracy$ &
        0.7253 &0.7212 &0.7806 &0.7066 &0.7361 &0.7123 &0.7200 &0.6338 &0.7318 &0.6340 &0.6599 &0.5751 &0.6968 &0.6064 &0.5791 &0.5182 \\
        & $Precision$ &
        0.5111 &0.3981 &0.3060 &0.2363 &0.5426 &0.1935 &0.2304 &0.1935 &0.5940 &0.2067 &0.1998 &0.1739 &0.5267 &0.5612 &0.5712 &0.5667 \\
        & $Recall$ &
        0.5236 &0.3040 &0.3102 &0.3977 &0.4188 &0.2531 &0.3411 &0.4333 &0.4273 &0.3632 &0.3956 &0.4659 &0.4209 &0.3669 &0.4450 &0.5288 \\
        & $\mathit{F1}\ Sorce$ &
        0.5173 &0.3448 &0.3081 &0.2964 &0.4728 &0.2194 &0.2751 &0.2675 &0.4971 &0.2635 &0.2655 &0.2533 &0.4679 &0.4437 &0.5002 &0.5471 \\
        & $Cost$ &
        2.3199 &3.2191 &2.6208 &3.3071 &1.2934 &3.3390 &3.2075 &4.0125 &1.3135 &4.0543 &3.7755 &4.5796 &1.4225 &4.3278 &4.5528 &5.1100 \\
        \midrule
        \multirow{5}{*}{$Average$}
        & $Accuracy$ &
        0.8786 &0.8659 &\textbf{0.8857} &0.8280 &
        \textbf{0.8522} &0.8149 &0.8118 &0.7306 &
        \textbf{0.7940} &0.7273 &0.7276 &0.6441 &
        \textbf{0.7070} &0.6414 &0.6263 &0.5575 \\
        & $Precision$ &
        \textbf{0.8193} &0.7744 &0.7134 &0.5877 &
        \textbf{0.7759} &0.6406 &0.5782 &0.4773 &
        \textbf{0.7900} &0.5537 &0.5004 &0.4275 &
        \textbf{0.6888} &0.4759 &0.4445 &0.3933 \\
        & $Recall$ &
        0.7860 &0.7431 &0.7715 &\textbf{0.8060} &
        \textbf{0.7570} &0.6529 &0.7240 &0.7330 &
        0.6283 &0.6392 &0.6737 &\textbf{0.6853} &
        0.5667 &0.5411 &\textbf{0.6045} &0.5965 \\
        & $\mathit{F1}\ Sorce$ &
        \textbf{0.7944} &0.7445 &0.7378 &0.6559 &
        \textbf{0.7496} &0.6217 &0.6239 &0.5469 &
        \textbf{0.6867} &0.5761 &0.5574 &0.5004 &
        \textbf{0.6128} &0.4853 &0.4855 &0.4448 \\
        & $Cost$ &
        \textbf{1.2514} &1.5648 &1.3290 &1.9230 &
        \textbf{1.2291} &1.9951 &2.0531 &2.9811 &
        \textbf{1.3925} &2.7409 &3.0374 &3.9235 &
        \textbf{2.1300} &3.8681 &4.1716 &4.8849 \\
        \cmidrule(r){1-6} \cmidrule(lr){7-10} \cmidrule(lr){11-14} \cmidrule(l){15-18}
        \multirow{3}{*}{$Statistics$}
        & $win/loss$
        &123/57&41/19&34/26&48/12
        &153/27&48/12&53/7&52/8
        &152/28&51/9&51/9&50/10
        &149/31&52/8&48/12&49/11
        \\
        & $rank$
        &2.3833 &2.4833 &\textbf{2.2750} &2.8583 
        &\textbf{1.9500} &2.6000 &2.5333 &2.9167 
        &\textbf{2.0333} &2.5500 &2.4500 &2.9667 
        &\textbf{2.1250} &2.7167 &2.3583 &2.8000 \\
        & $p\text{-}values$
        &-&1.43E-03&2.93E-02&5.35E-07 &-&8.73E-06&1.38E-08&4.63E-09
        &-&3.37E-07&6.48E-07&7.10E-08 &-&4.51E-08&6.06E-07&3.23E-08 \\
        \bottomrule
    \end{tabular}
    }
\end{table*}

\begin{figure*}[htbp]
    \centering
    \resizebox{\textwidth}{!}{
        \begin{subfigure}[b]{0.25\textwidth}
            \centering
            \includegraphics[width=\textwidth]{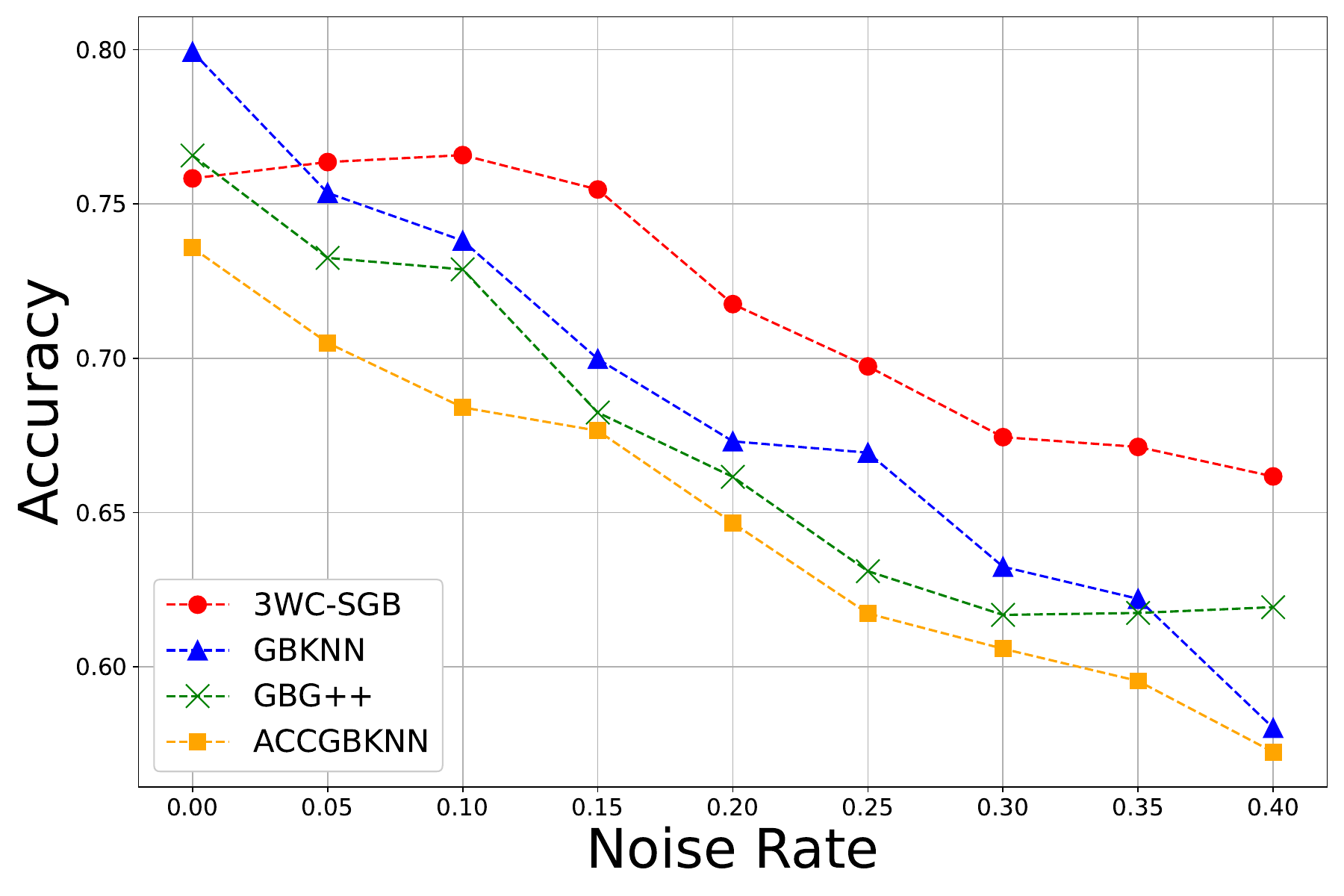}
            \caption{SPECTF}
            \label{fig:7:1}
        \end{subfigure}
        \hfill
        \begin{subfigure}[b]{0.25\textwidth}
            \centering
            \includegraphics[width=\textwidth]{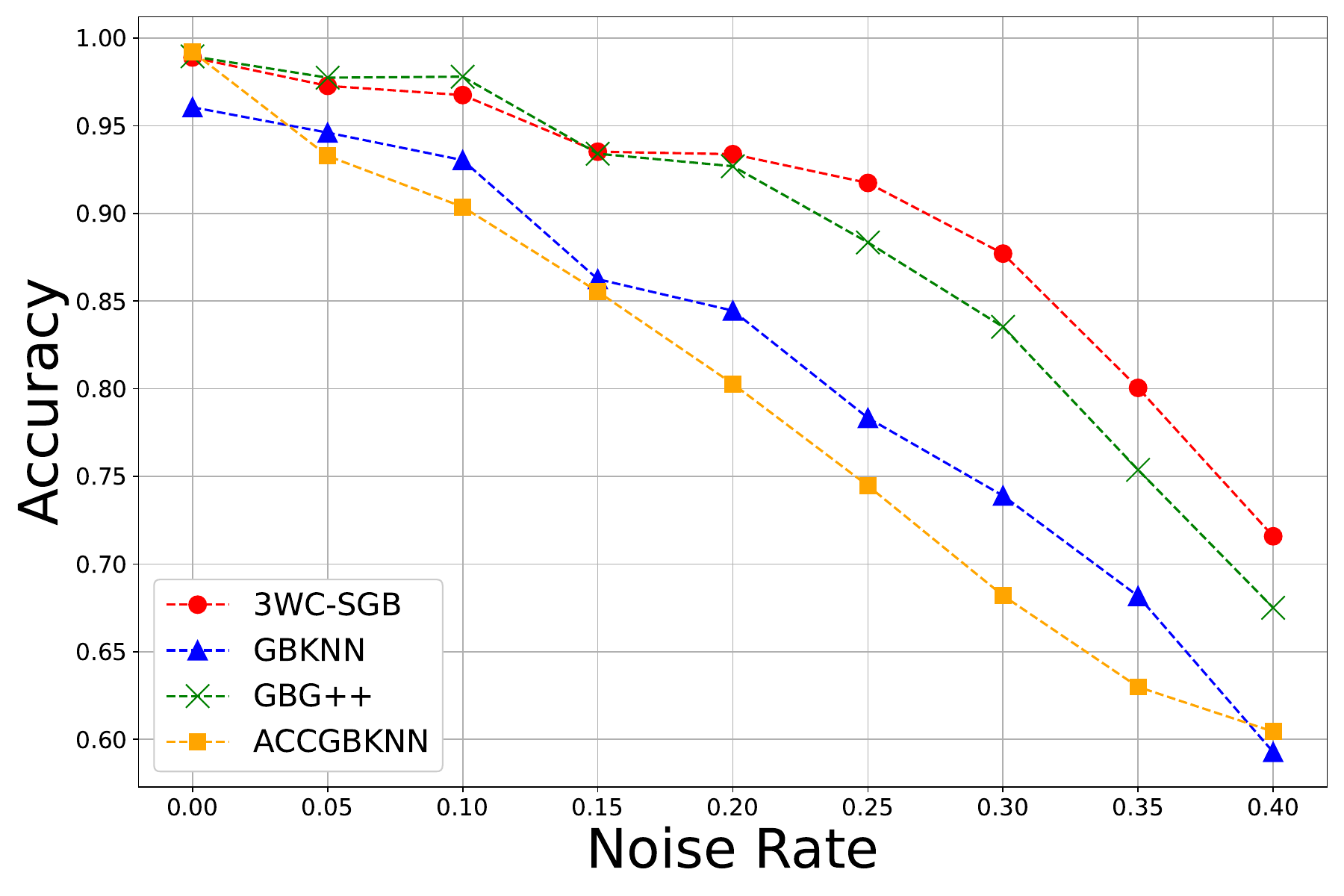}
            \caption{Fourclass}
            \label{fig:7:2}
        \end{subfigure}
        \hfill
        \begin{subfigure}[b]{0.25\textwidth}
            \centering
            \includegraphics[width=\textwidth]{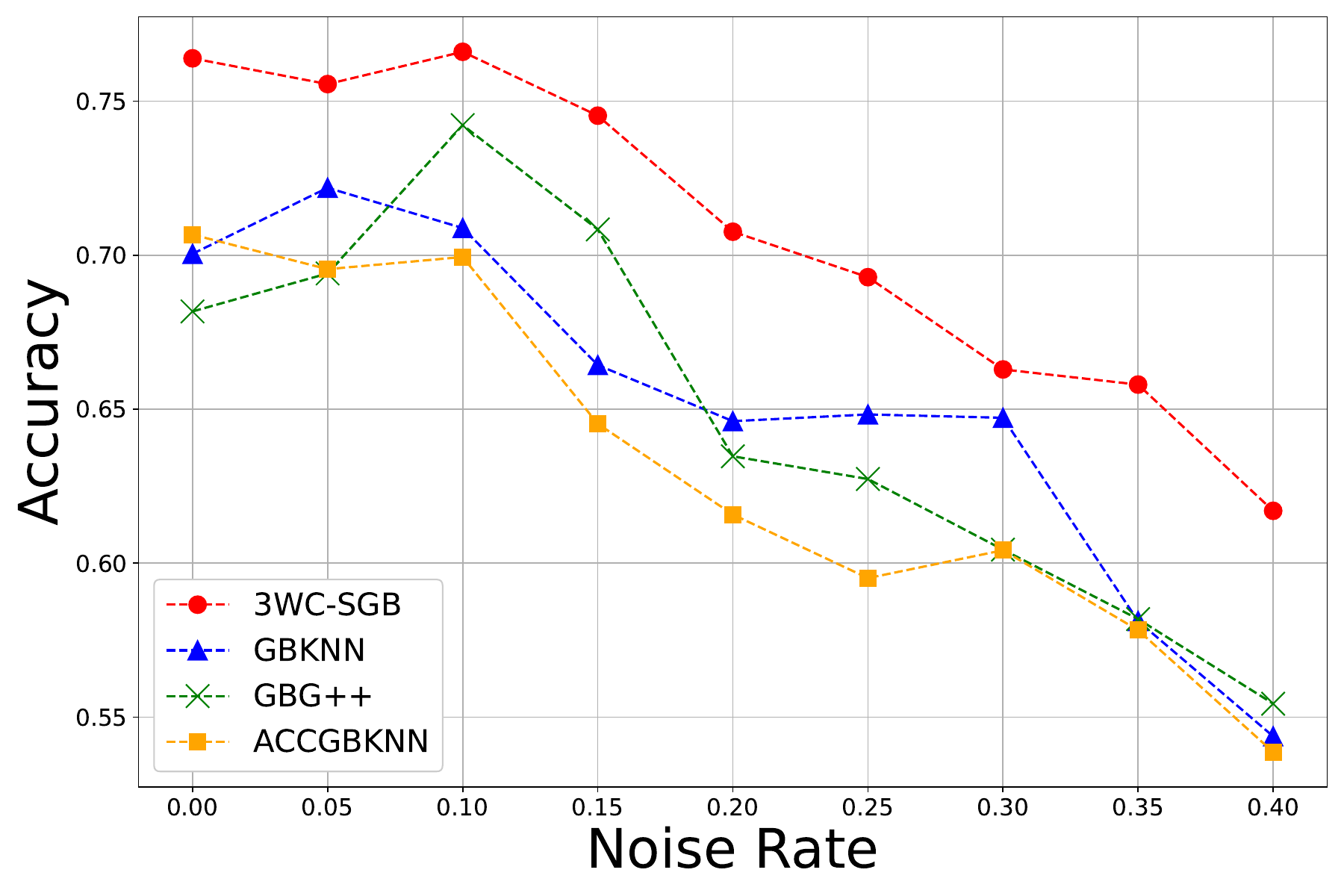}
            \caption{Endgame}
            \label{fig:7:3}
        \end{subfigure}
        \hfill
        \begin{subfigure}[b]{0.25\textwidth}
            \centering
            \includegraphics[width=\textwidth]{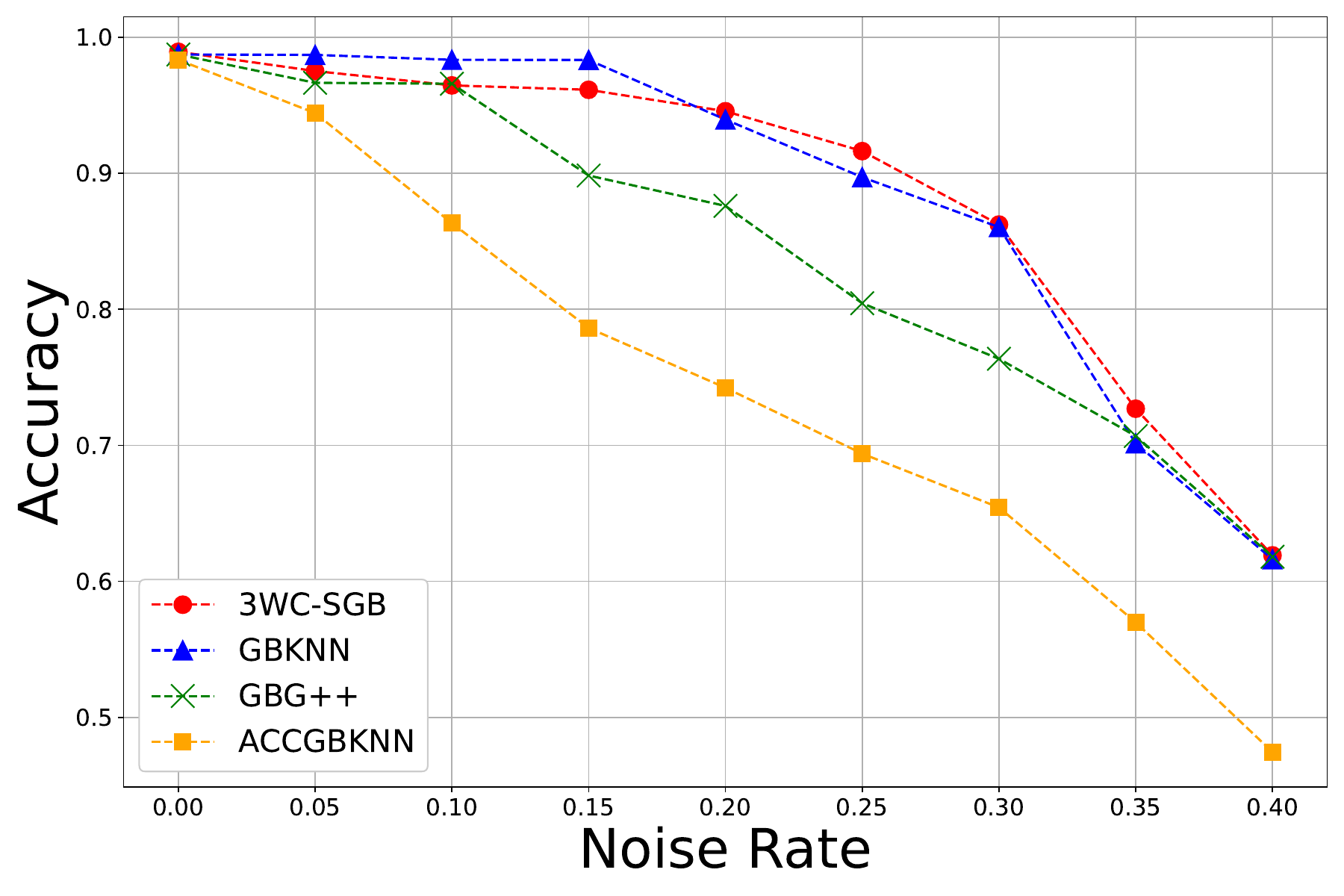}
            \caption{Cloud}
            \label{fig:7:4}
        \end{subfigure}
    }

    \medskip

    \resizebox{\textwidth}{!}{
        \begin{subfigure}[b]{0.25\textwidth}
            \centering
            \includegraphics[width=\textwidth]{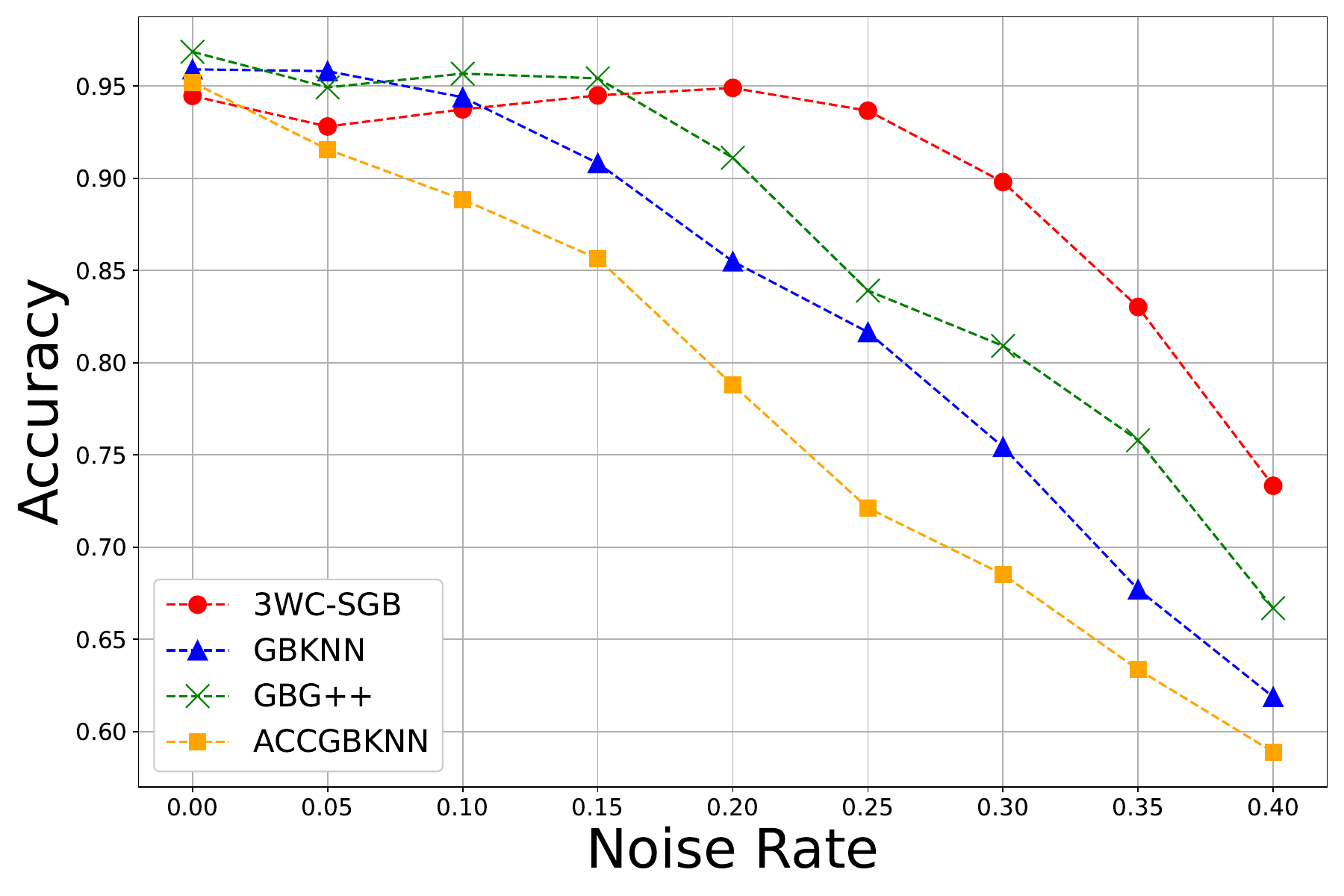}
            \caption{Banknote Authentication}
            \label{fig:7:5}
        \end{subfigure}
        \hfill
        \begin{subfigure}[b]{0.25\textwidth}
            \centering
            \includegraphics[width=\textwidth]{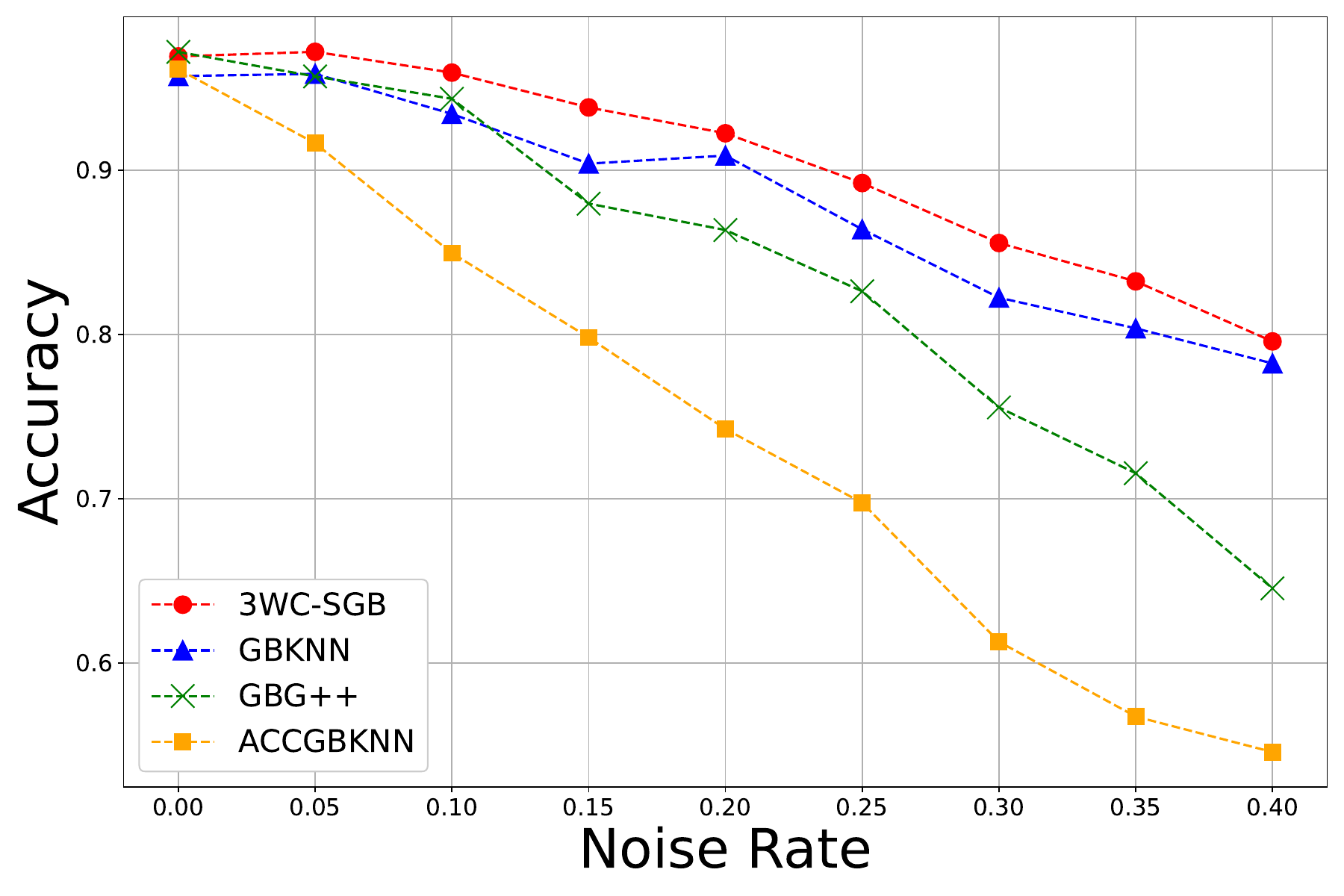}
            \caption{NHANES Age Prediction}
            \label{fig:7:6}
        \end{subfigure}
        \hfill
        \begin{subfigure}[b]{0.25\textwidth}
            \centering
            \includegraphics[width=\textwidth]{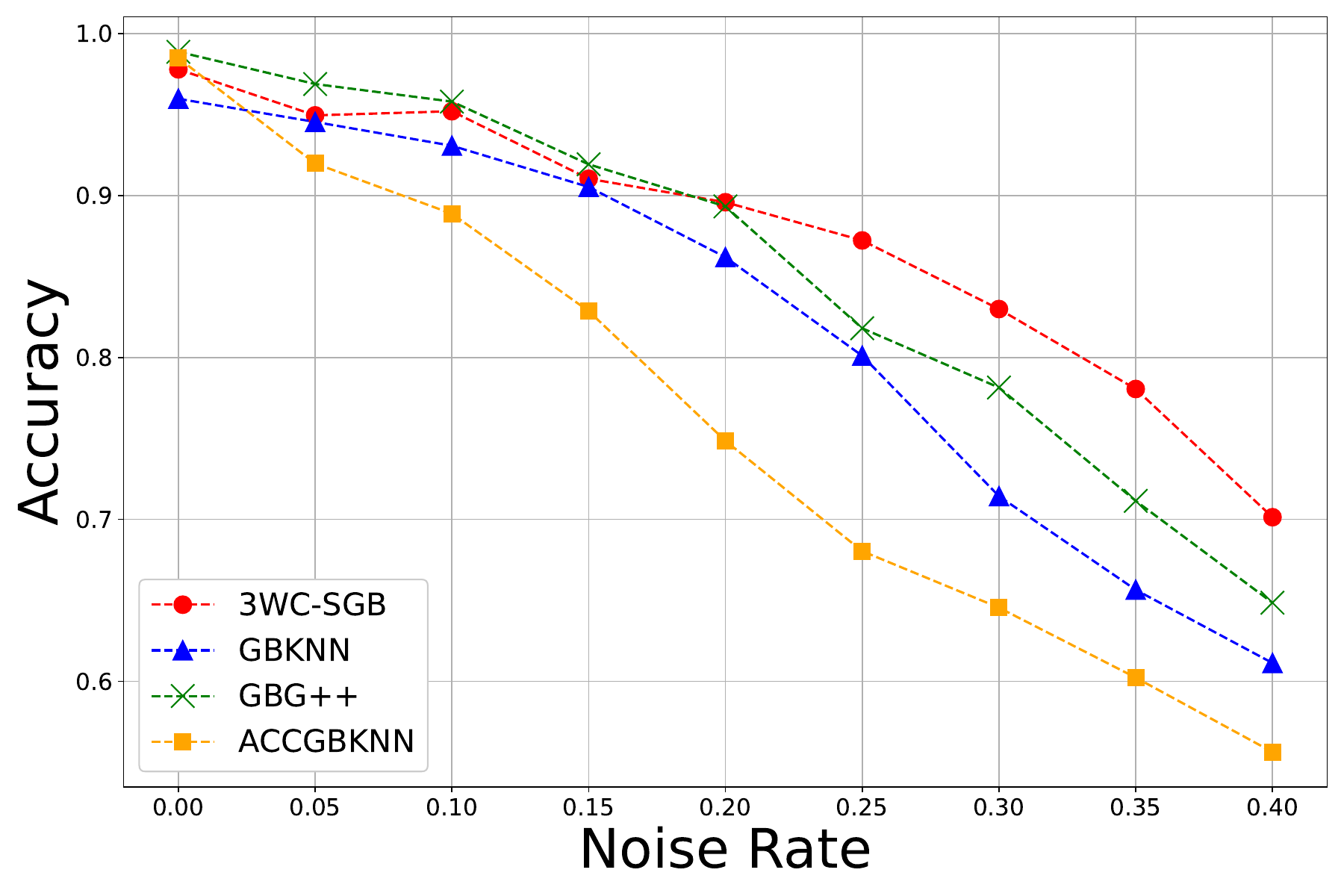}
            \caption{Segment}
            \label{fig:7:7}
        \end{subfigure}
        \hfill
        \begin{subfigure}[b]{0.25\textwidth}
            \centering
            \includegraphics[width=\textwidth]{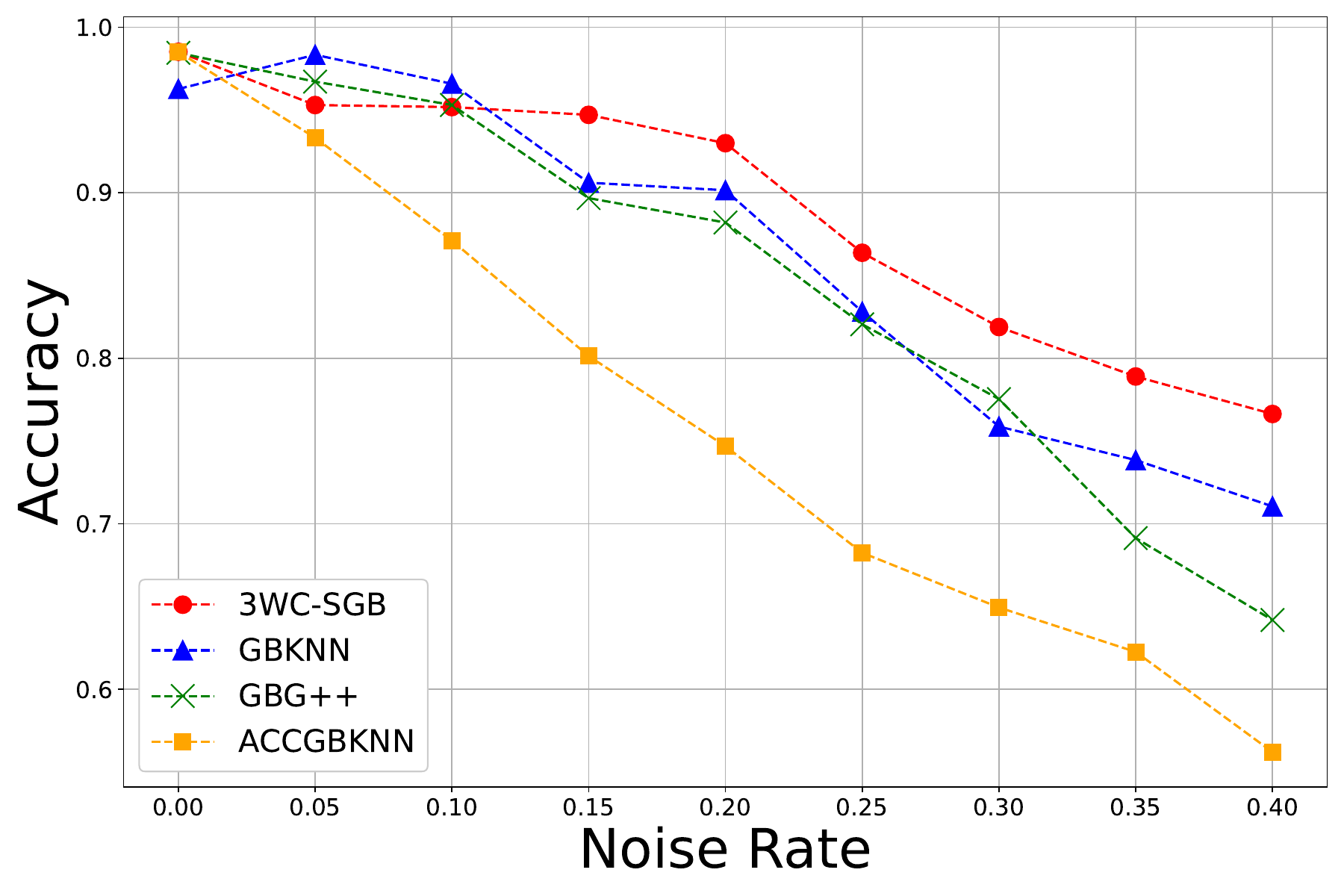}
            \caption{Shill Bidding}
            \label{fig:7:8}
        \end{subfigure}
    }

    \medskip

    \resizebox{\textwidth}{!}{
        \begin{subfigure}[b]{0.25\textwidth}
            \centering
            \includegraphics[width=\textwidth]{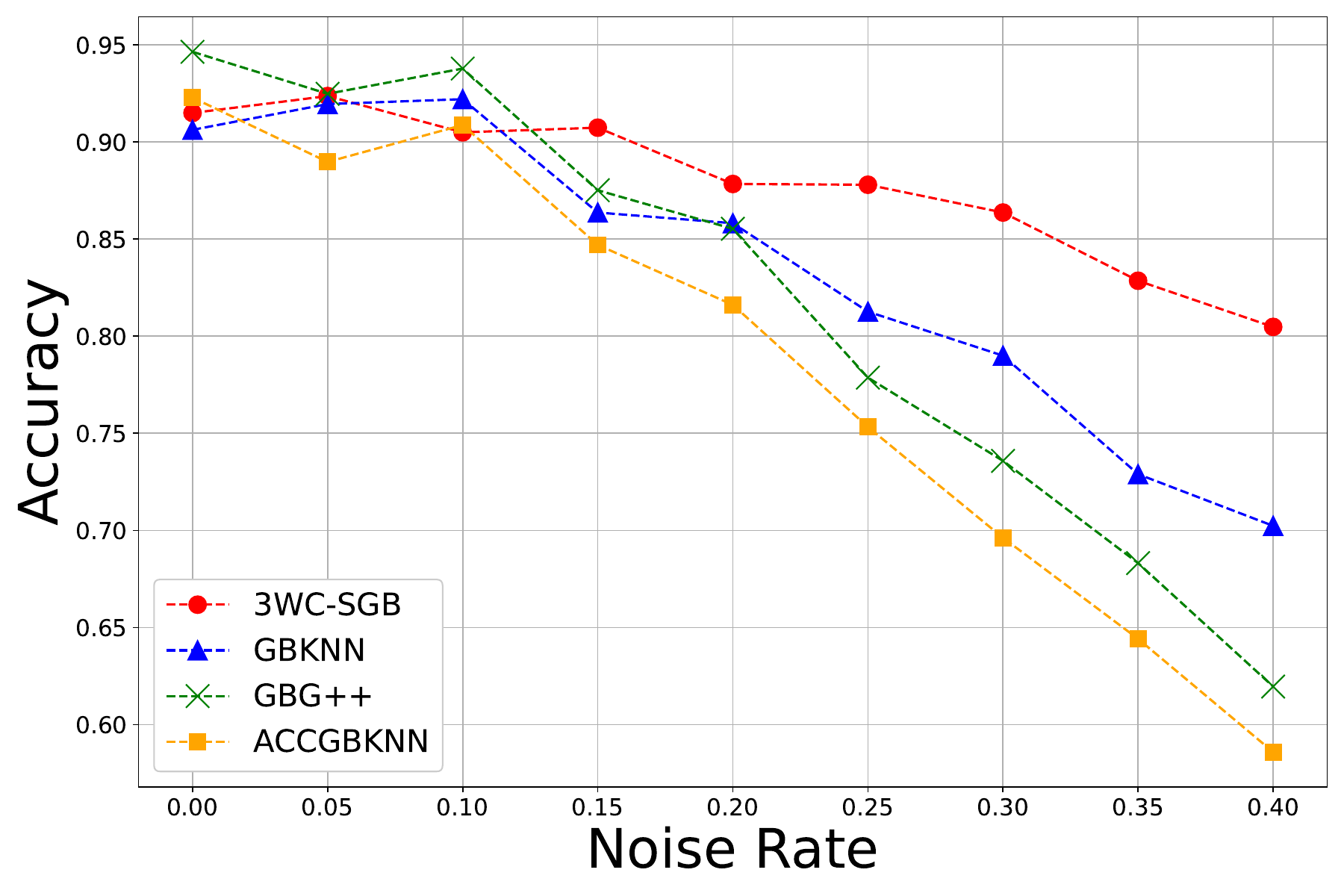}
            \caption{Satimage}
            \label{fig:7:9}
        \end{subfigure}
        \hfill
        \begin{subfigure}[b]{0.25\textwidth}
            \centering
            \includegraphics[width=\textwidth]{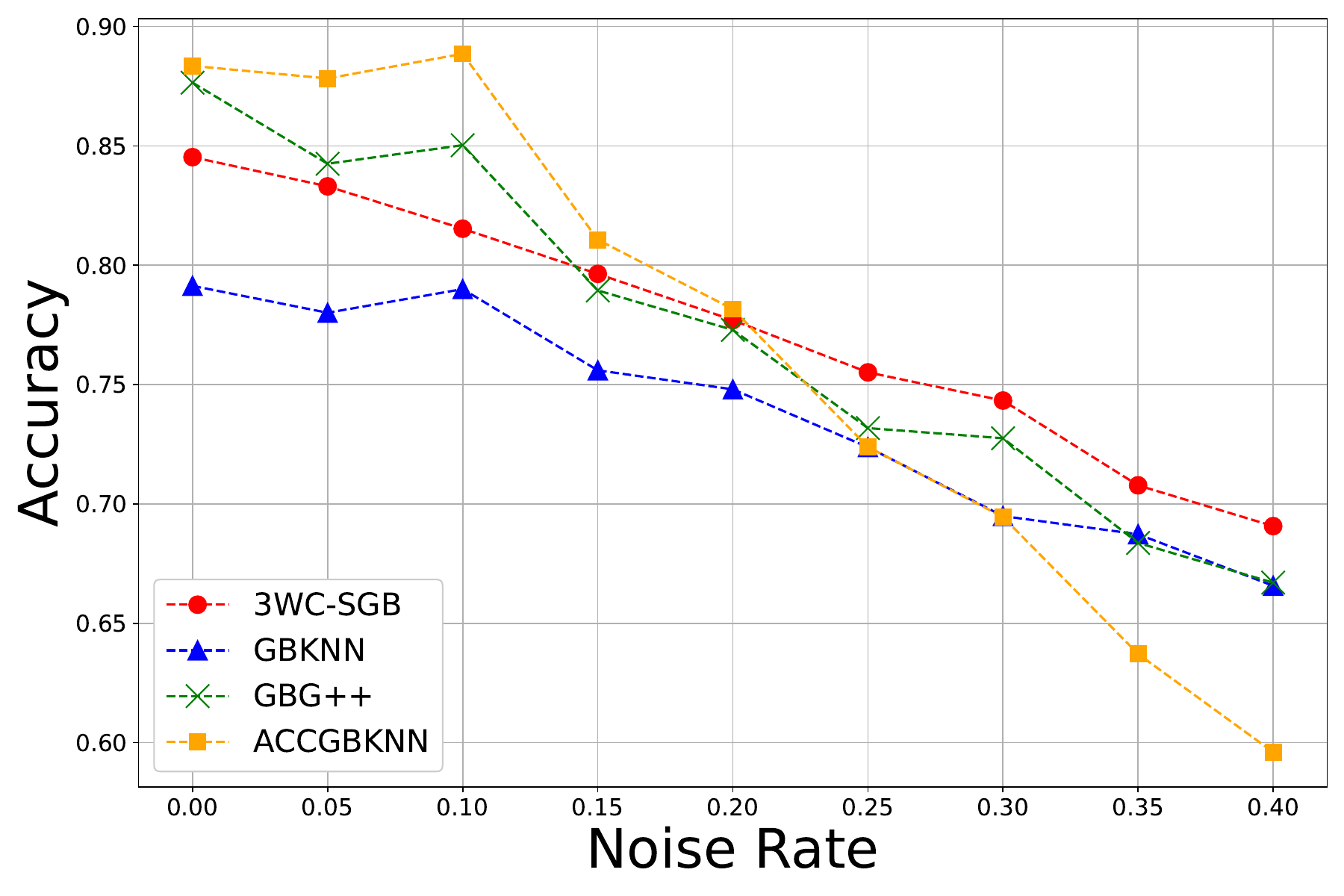}
            \caption{mushroom}
            \label{fig:7:10}
        \end{subfigure}
        \hfill
        \begin{subfigure}[b]{0.25\textwidth}
            \centering
            \includegraphics[width=\textwidth]{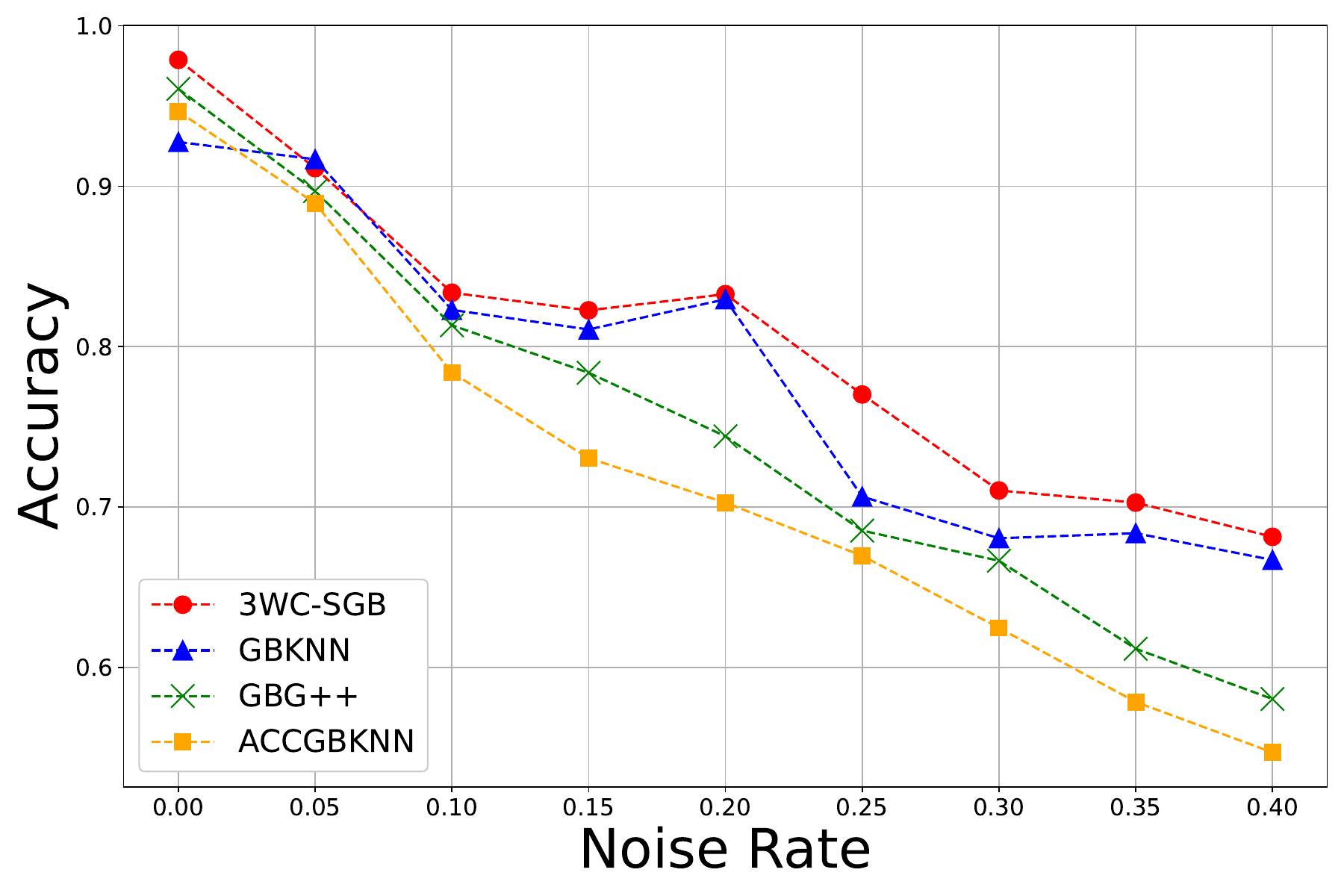}
            \caption{Elect}
            \label{fig:7:11}
        \end{subfigure}
        \hfill
        \begin{subfigure}[b]{0.25\textwidth}
            \centering
            \includegraphics[width=\textwidth]{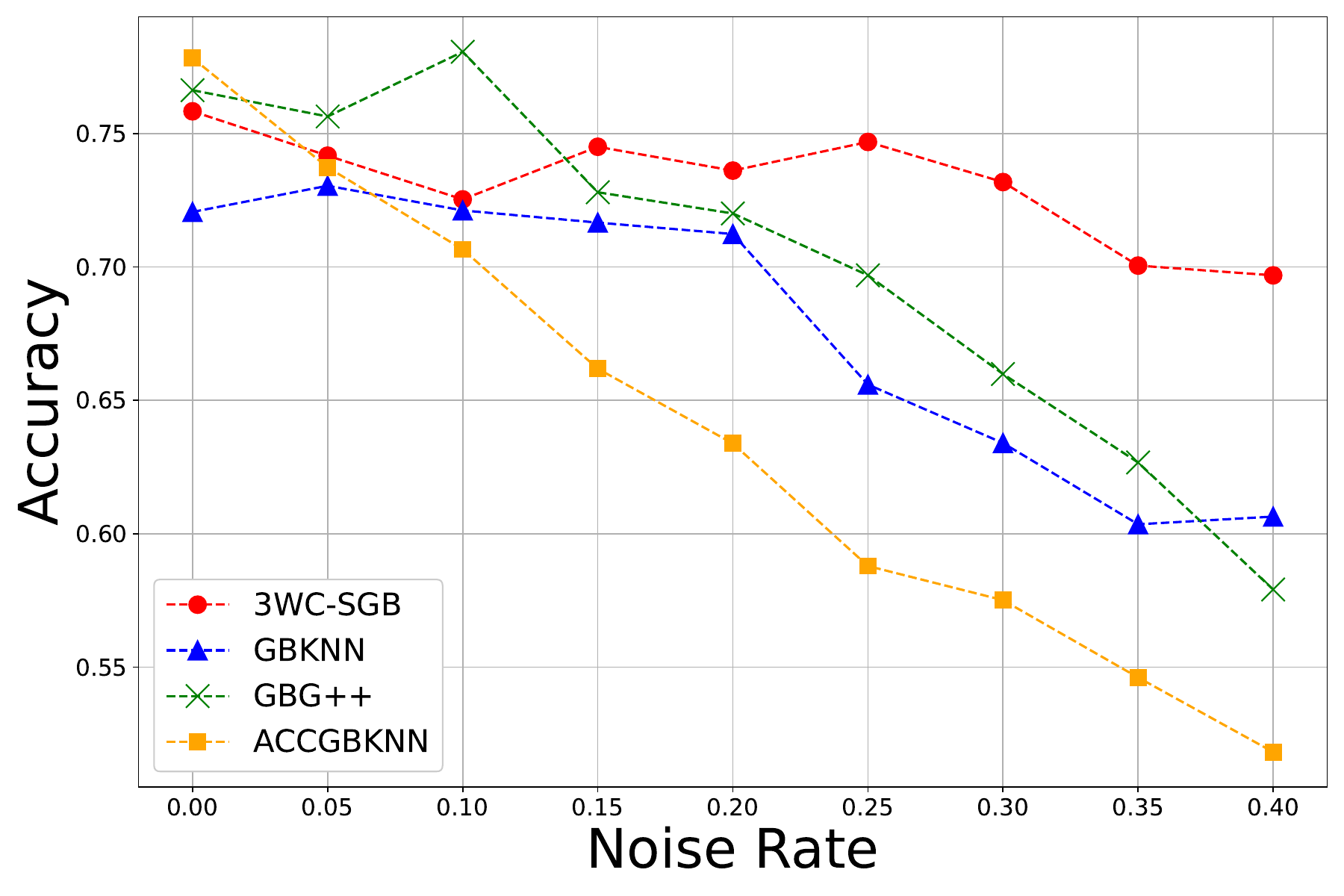}
            \caption{Online Shoppers Intention}
            \label{fig:7:12}
        \end{subfigure}
    }
    \captionsetup{justification=centering}
    \caption{The Comparison with GB-based Classifiers under Different Noise Rates}
    \label{fig:7}
\end{figure*}

\begin{figure*}[htbp]
    \centering
    \resizebox{0.9\textwidth}{!}{
        \begin{subfigure}[b]{0.25\textwidth}
            \centering
            \includegraphics[width=\textwidth]{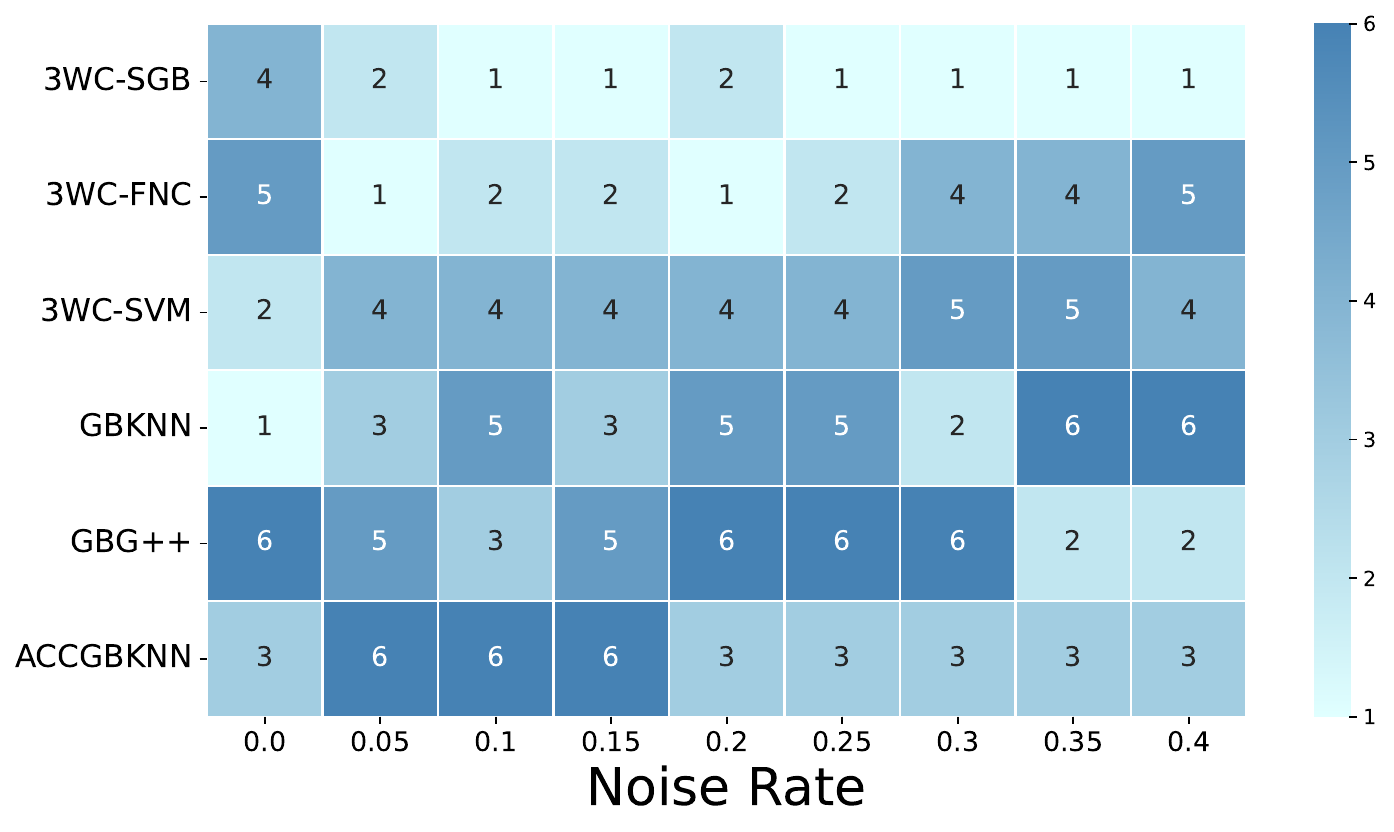}
            \caption{SPECTF}
            \label{fig:8:1}
        \end{subfigure}
        \hfill
        \begin{subfigure}[b]{0.25\textwidth}
            \centering
            \includegraphics[width=\textwidth]{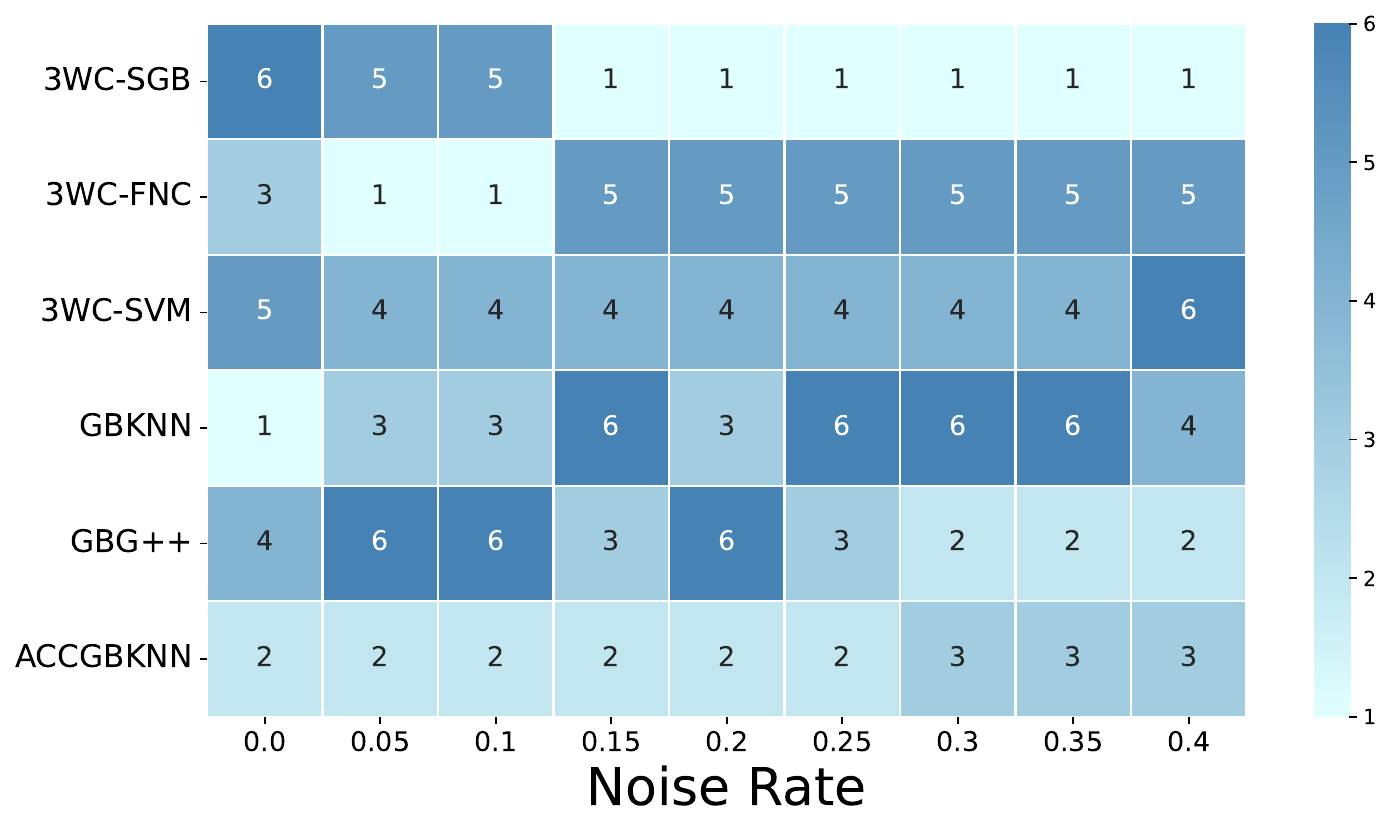}
            \caption{Fourclass}
            \label{fig:8:2}
        \end{subfigure}
        \hfill
        \begin{subfigure}[b]{0.25\textwidth}
            \centering
            \includegraphics[width=\textwidth]{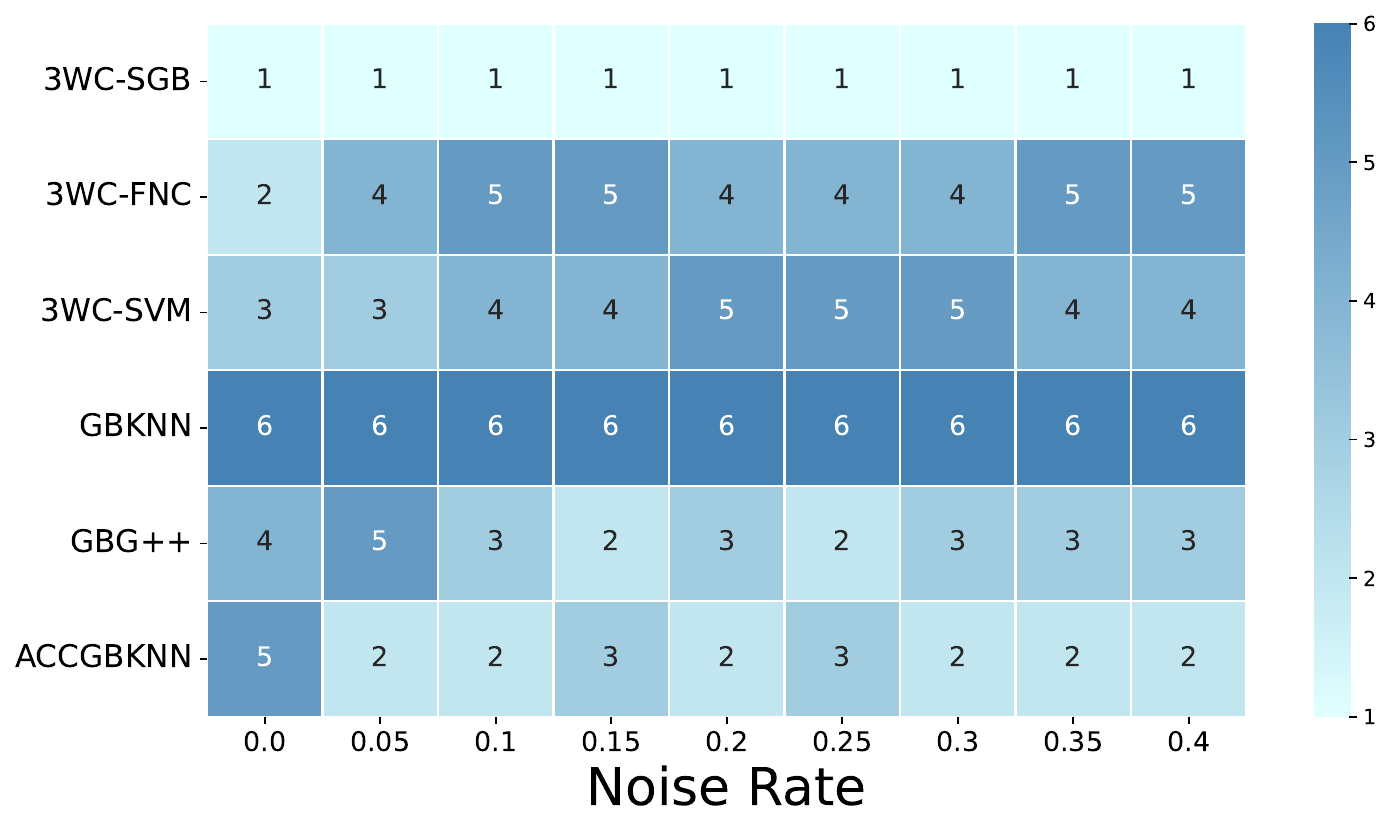}
            \caption{Endgame}
            \label{fig:8:3}
        \end{subfigure}
        \hfill
        \begin{subfigure}[b]{0.25\textwidth}
            \centering
            \includegraphics[width=\textwidth]{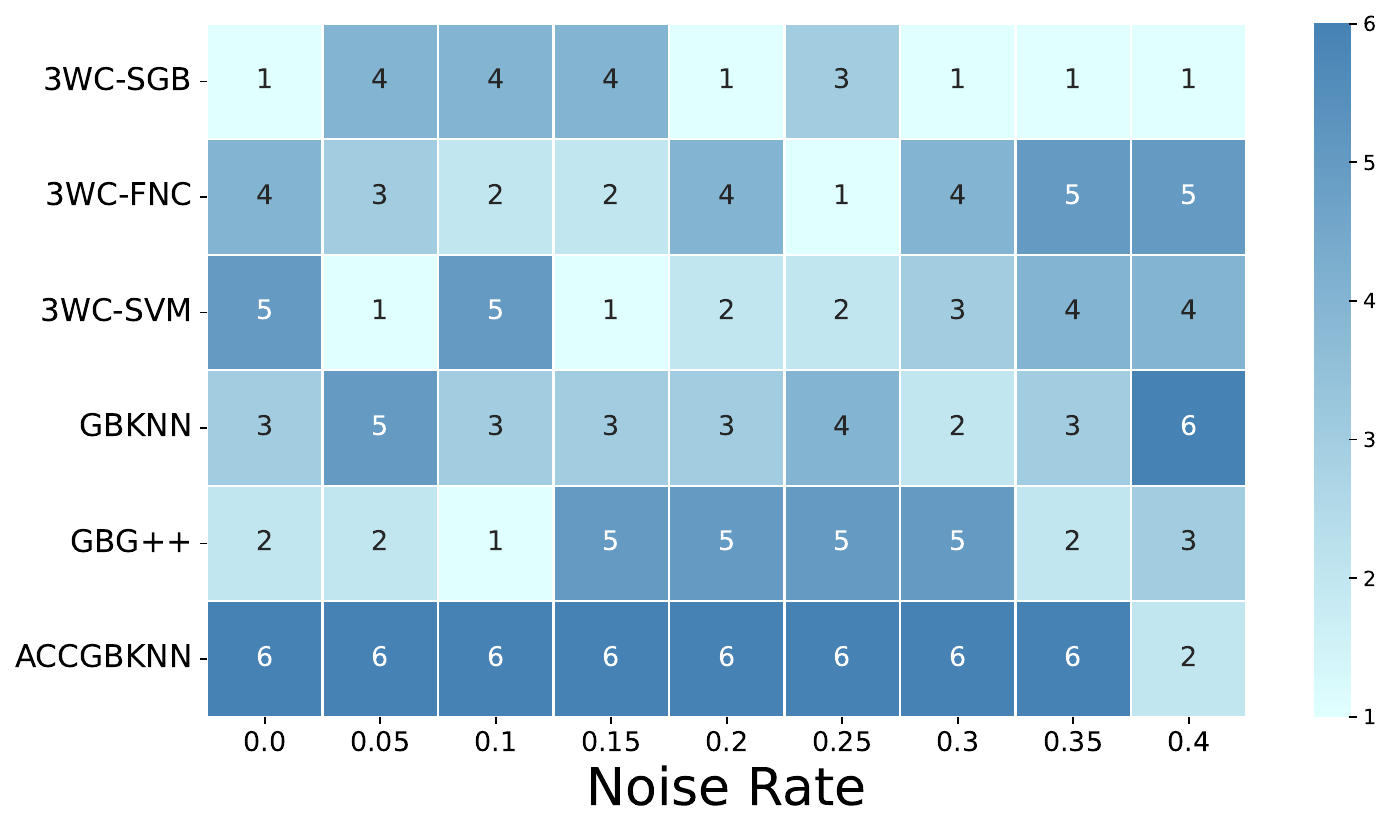}
            \caption{Cloud}
            \label{fig:8:4}
        \end{subfigure}
    }

    \medskip

    \resizebox{0.9\textwidth}{!}{
        \begin{subfigure}[b]{0.25\textwidth}
            \centering
            \includegraphics[width=\textwidth]{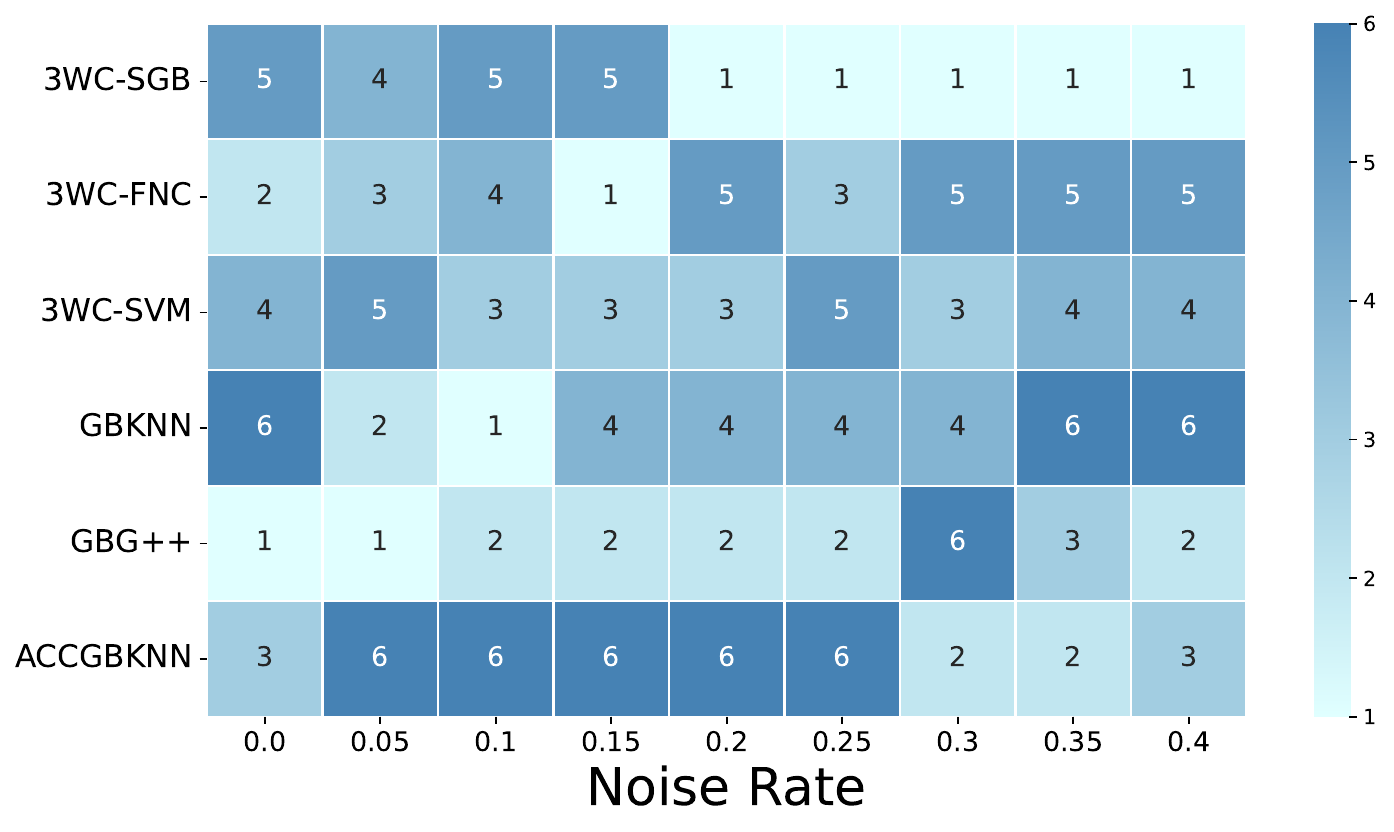}
            \caption{Banknote Authentication}
            \label{fig:8:5}
        \end{subfigure}
        \hfill
        \begin{subfigure}[b]{0.25\textwidth}
            \centering
            \includegraphics[width=\textwidth]{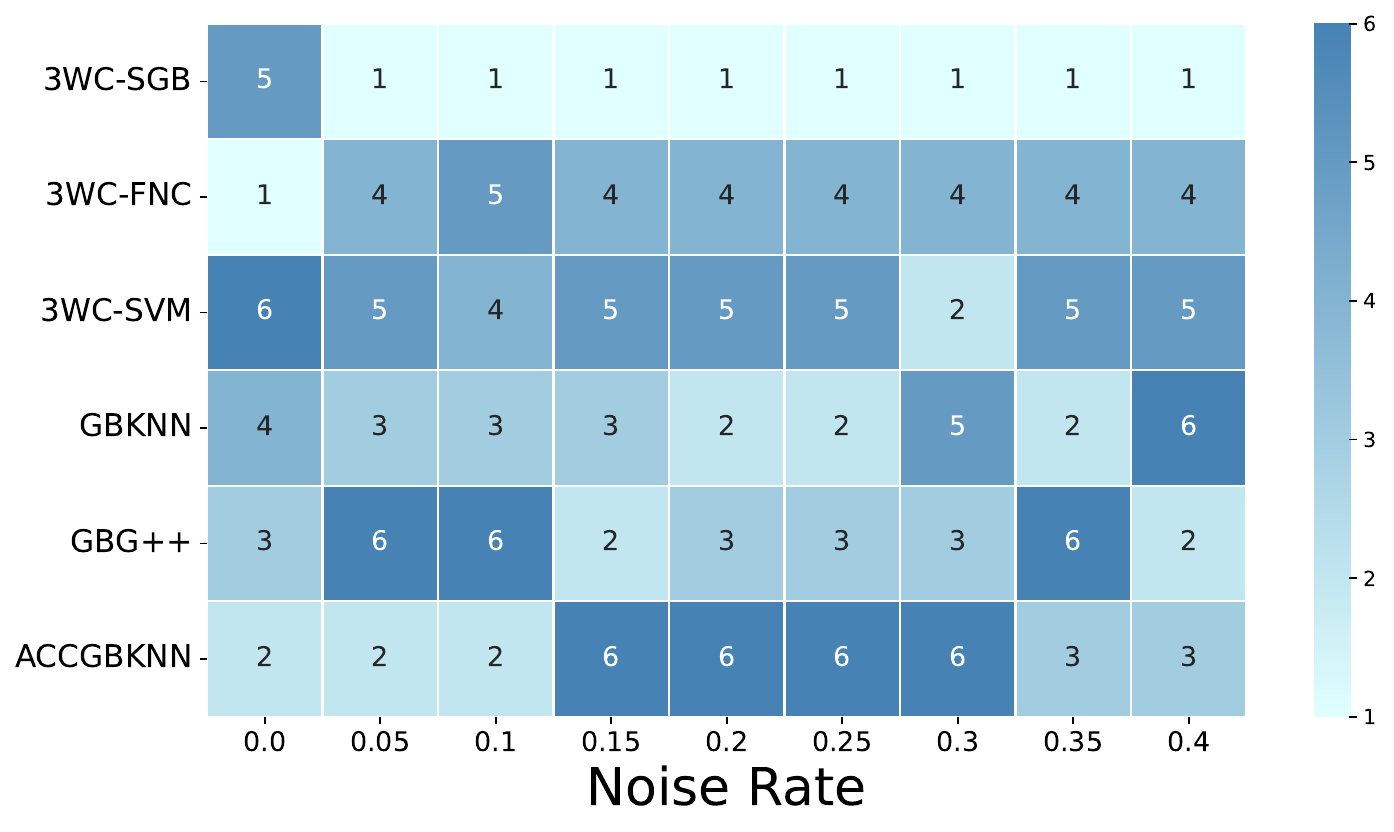}
            \caption{NHANES Age Prediction}
            \label{fig:8:6}
        \end{subfigure}
        \hfill
        \begin{subfigure}[b]{0.25\textwidth}
            \centering
            \includegraphics[width=\textwidth]{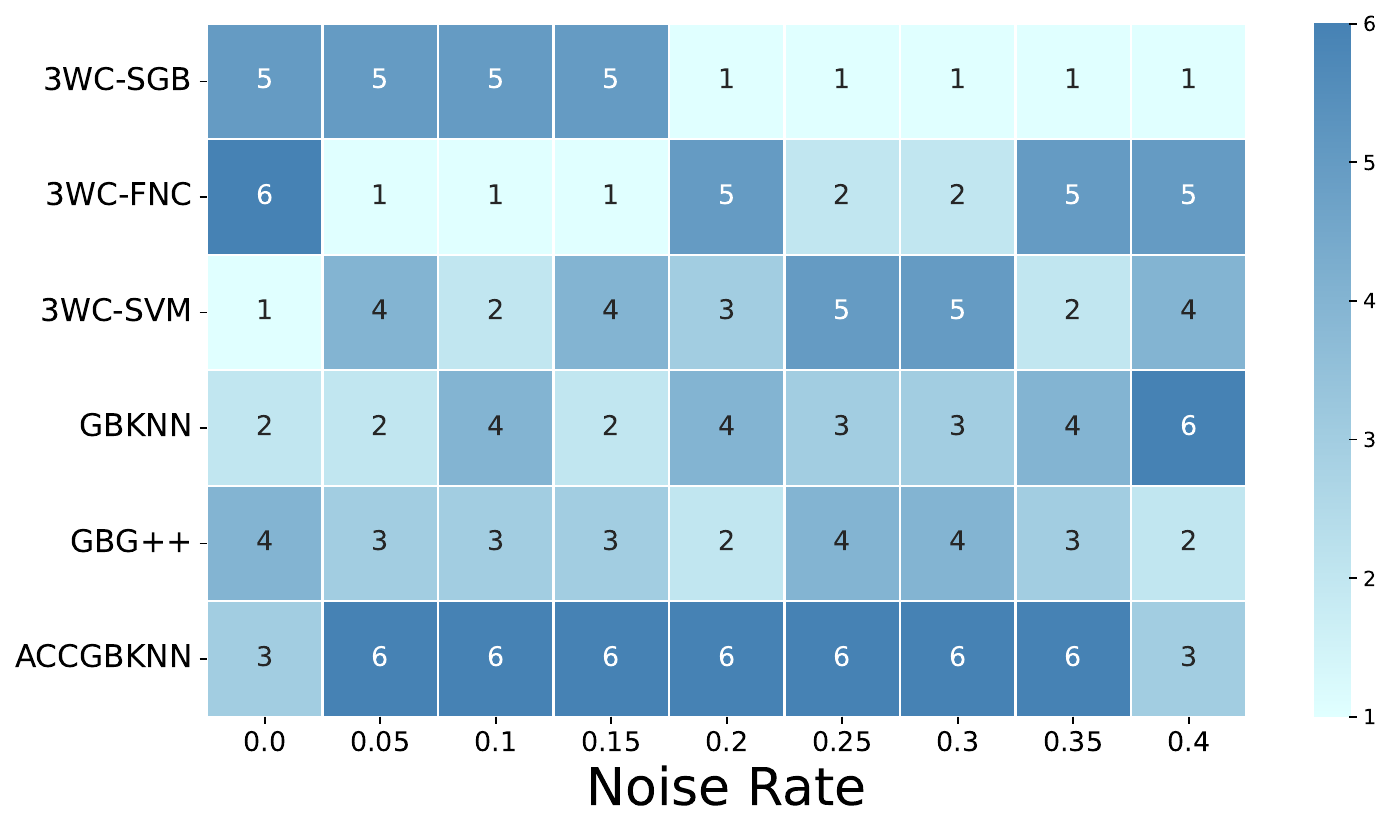}
            \caption{Segment}
            \label{fig:8:7}
        \end{subfigure}
        \hfill
        \begin{subfigure}[b]{0.25\textwidth}
            \centering
            \includegraphics[width=\textwidth]{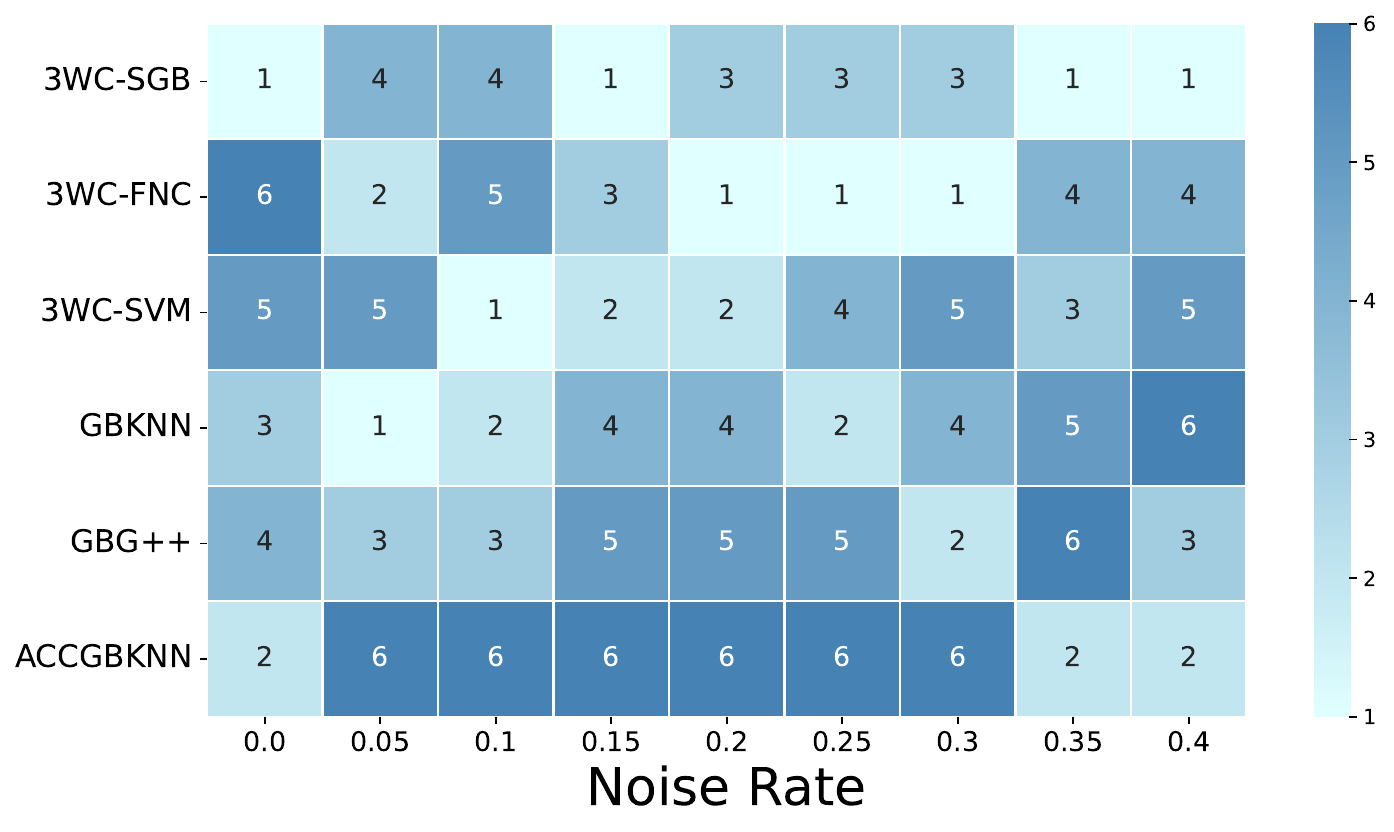}
            \caption{Shill Bidding}
            \label{fig:8:8}
        \end{subfigure}
    }

    \medskip

    \resizebox{0.9\textwidth}{!}{
        \begin{subfigure}[b]{0.25\textwidth}
            \centering
            \includegraphics[width=\textwidth]{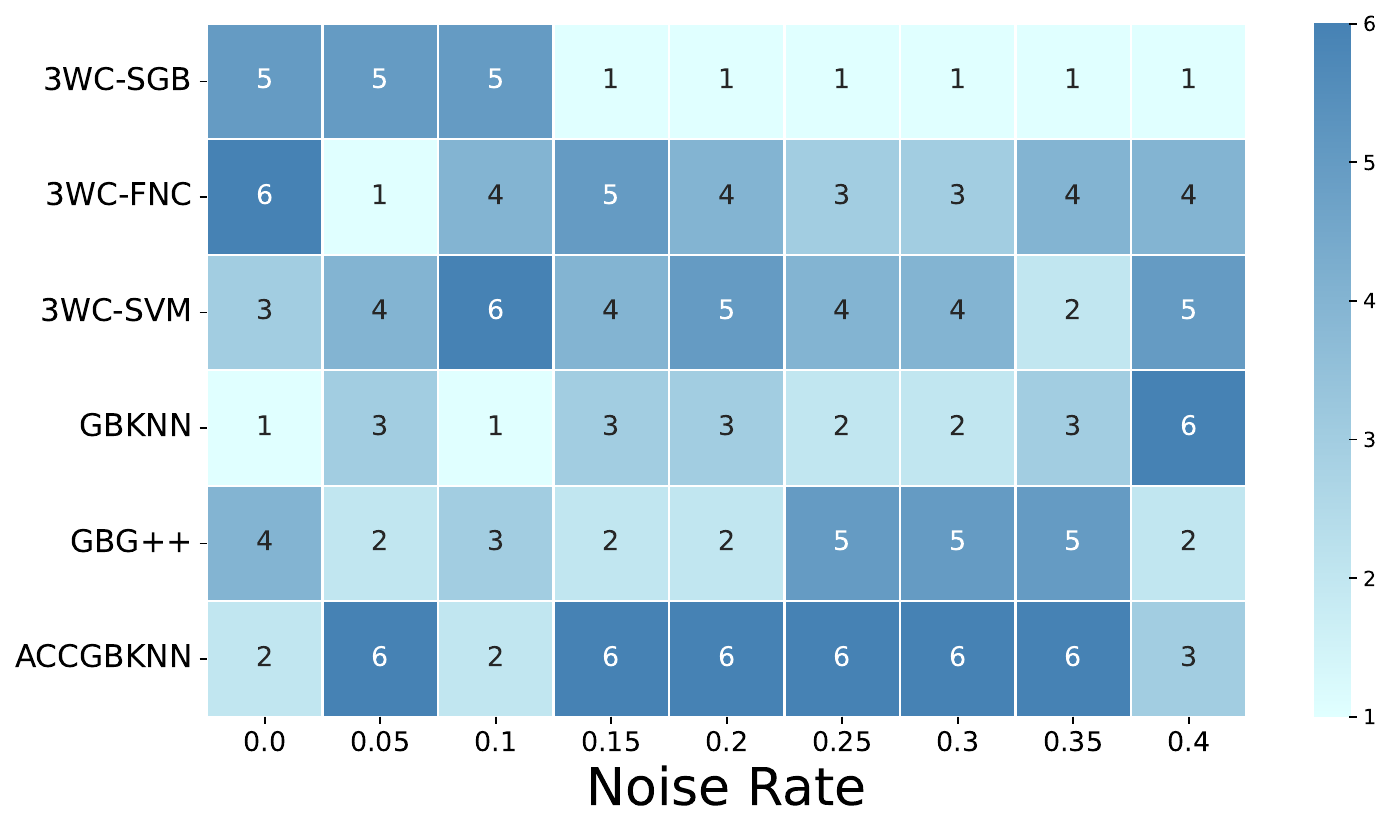}
            \caption{Satimage}
            \label{fig:8:9}
        \end{subfigure}
        \hfill
        \begin{subfigure}[b]{0.25\textwidth}
            \centering
            \includegraphics[width=\textwidth]{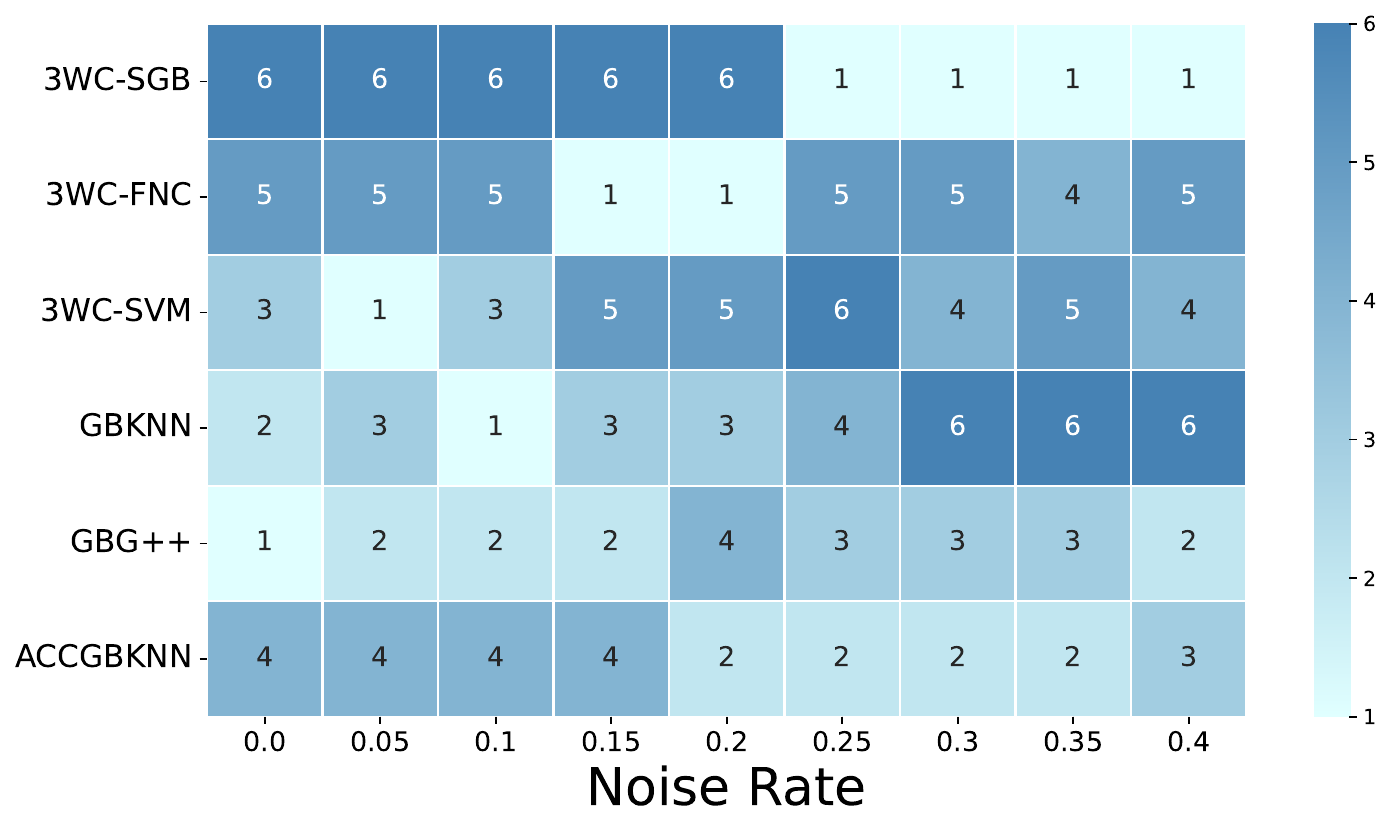}
            \caption{mushroom}
            \label{fig:hmgb:10}
        \end{subfigure}
        \hfill
        \begin{subfigure}[b]{0.25\textwidth}
            \centering
            \includegraphics[width=\textwidth]{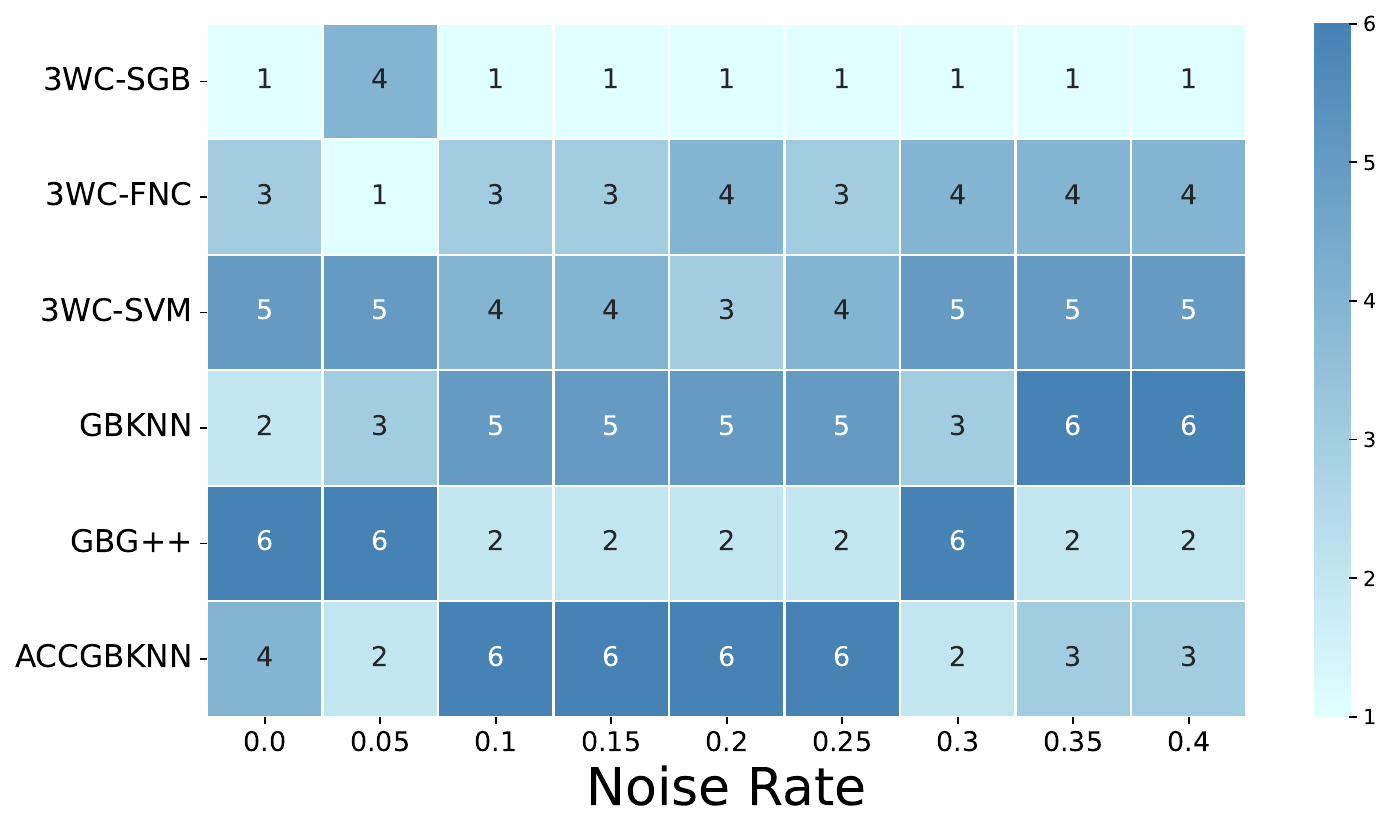}
            \caption{Elect}
            \label{fig:8:11}
        \end{subfigure}
        \hfill
        \begin{subfigure}[b]{0.25\textwidth}
            \centering
            \includegraphics[width=\textwidth]{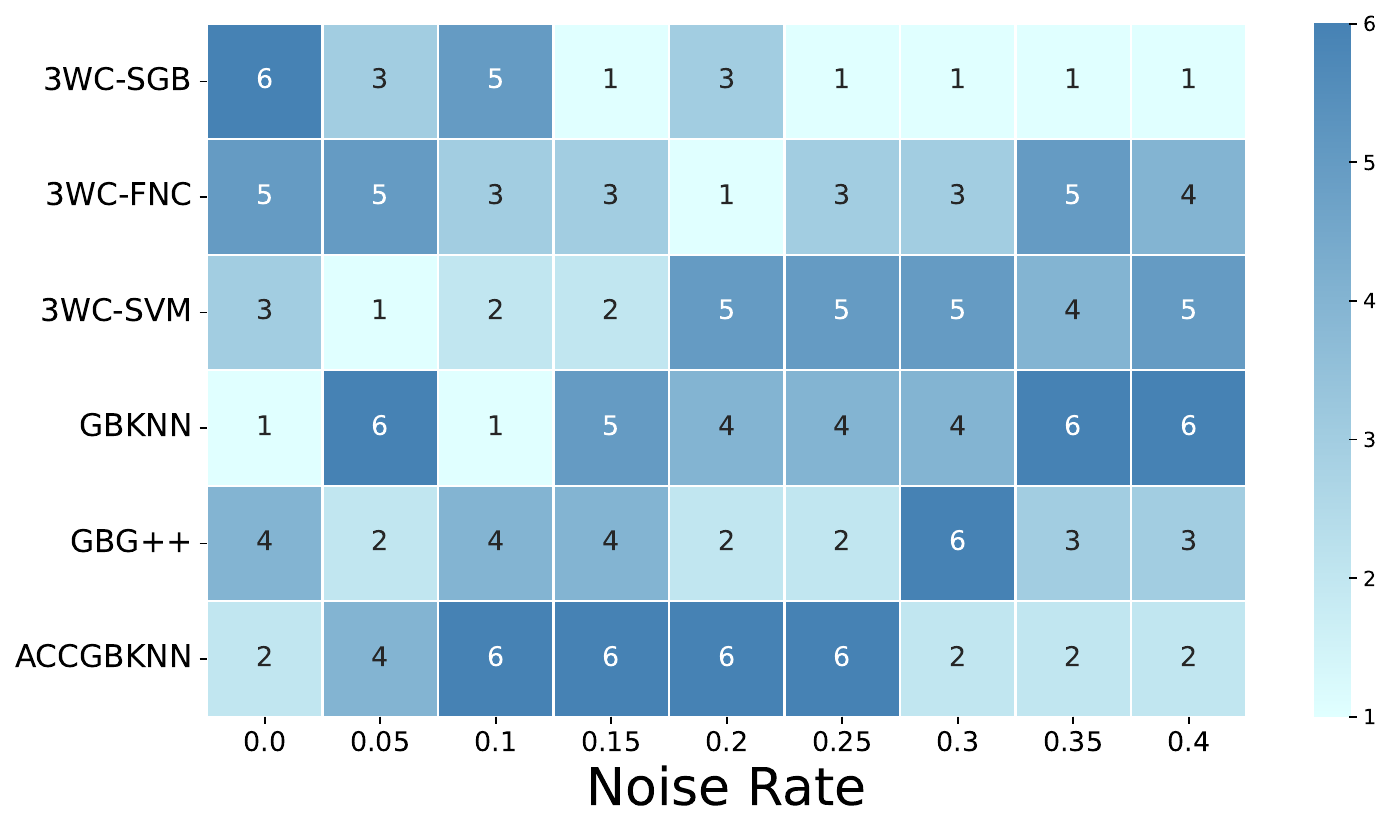}
            \caption{Online Shoppers Intention}
            \label{fig:8:12}
        \end{subfigure}
    }
    \captionsetup{justification=centering}
    \caption{The Ranking Order of $Accuracy$ under Different Noise Rates}
    \label{fig:8}
\end{figure*}

\subsection{The Comparison with Three-way Classifiers}\label{sec:5:sub1}
Firstly, we conducted experiments on various datasets with different noise rates with two other three-way classifiers. The $Accuracy$ and $UR$ curves are depicted in Fig. \ref{fig:5}. Due to the significant differences in high noise rates, detailed analysis are provided in Section S4 of supplementary file. Along the horizontal axis, different noise rates are represented, while $Accuracy$ and $UR$ are depicted by solid lines and dashed lines, respectively. $UR$ serves as a measure of the extent that the classifier classifies samples to uncertain case. There lacks of a clear-cut comparative method for this metric, but the high value of $UR$ indicates that the classifier categorizes too many samples as uncertain case, resulting in reduced $Accuracy$. Conversely, the low value of $UR$ indicates that which will make the classifier attempt classification on high-risk data, deviating from the intention of three-way classification and indistinguishable from binary classifier. From Fig. \ref{fig:5}, 3WC-FNC and 3WC-SVM have a substantial increase in $UR$ and a corresponding significant decrease in $Accuracy$ when the noise rate exceeds a certain threshold. In the presence of high-noise interference, 3WC-SGB appropriately increases the proportion of uncertain samples, avoiding the high cost of misclassification while maintaining a relatively high level of $Accuracy$. In conclusion, 3WC-SGB solved the issue of over-classifying samples to uncertain cases under high noise rates and improved robustness of classifier.

\subsection{The Comparison with GB-based Classifiers}\label{sec:5:sub2}
To investigate the advantages of 3WC-SGB in the context of GB-based classifiers, we should first understand the characteristics of GB-based classifiers, that is robustness and efficiency. For robustness comparison, Fig. \ref{fig:7} shows the variation of $Accuracy$ with different noise rates. The robustness of GB stems from the use of balls to cover data, naturally possessing certain noise resistance capabilities. In specific cases, a slight amount of noise may even lead to more reasonable GB generation, as depicted in Fig. \ref{fig:7:3}. 3WC-SGB mainly enhances the tolerance of GB to noise through by justifiable granularity. Compared to other classifiers, 3WC-SGB exhibits a more gradual decrease in accuracy curves as noise increases and has greater robustnes. At higher noise rates, it can better maintain the stability of the generated model without significant fluctuations, as illustrated in Fig. \ref{fig:7:5}. 

The detailed statistical analysis can be found in Table \ref{tab:2}. It is evident that 3WC-SGB maintains a significant lead within a certain level of noise. The primary weakness of 3WC-SGB lies in the $Recall$, which is attributed to the three-way classification. However, considering the $\mathit{F1}\ Sorce$ comprehensively, the loss in $Recall$ is acceptable. Regarding $P\text{-}values$, 3WC-SGB exhibits significant differences compared to other classifiers, and this difference increases with the addition of noise. In order to better understand the performance of the algorithm, the ranking order of $Accuracy$ under different noise rates compared with three-way classifiers and GB-based classifiers is presented in Fig. \ref{fig:8}. Obviously, 3WC-SGB maintains a lead on most datasets under low noise rates and consistently outperforms all datasets under high noise rates. In summary, 3WC-SGB demonstrates possesses stronger robustness, enabling better handling of noise.

\begin{table}[htbp]
    \captionsetup{justification=centering}
    \caption{The comparison of prediction time ($10^{\text{-}2}\text{s}$) \\under different noise rates}
    \label{tab:3}
    \centering
    \resizebox{0.82\linewidth}{!}{
        \begin{tabular}{cc*{4}{>{\centering\arraybackslash}p{1.1cm}}}
        \toprule
        \multirow{2}{*}{No.} & \multirow{2}{*}{Noise Rate} & 3WC-SGB & \multirow{2}{*}{GBKNN} & \multirow{2}{*}{GBG++} & ACC-GBKNN\\
        \midrule
        \multirow{3}{*}{1}
        & 0.0 & 0.080 & 0.212 & 0.070 & 0.390 \\
        & 0.2 & 0.080 & 0.877 & 0.100 & 0.440 \\
        & 0.4 & 0.070 & 0.877 & 0.090 & 0.520 \\
        \cmidrule(lr){1-6}
        \multirow{3}{*}{2}
        & 0.0 & 0.100 & 1.200 & 0.309 & 0.498 \\
        & 0.2 & 0.169 & 4.745 & 0.488 & 1.874 \\
        & 0.4 & 0.179 & 6.010 & 0.349 & 2.402 \\
        \cmidrule(lr){1-6}
        \multirow{3}{*}{3}
        & 0.0 & 0.349 & 5.053 & 0.528 & 2.332 \\       
        & 0.2 & 0.319 & 6.603 & 0.504 & 2.855 \\
        & 0.4 & 0.359 & 7.027 & 0.468 & 3.533 \\
        \cmidrule(lr){1-6}
        \multirow{3}{*}{4}
        & 0.0 & 0.010 & 0.718 & 0.140 & 0.578 \\
        & 0.2 & 0.319 & 7.156 & 0.825 & 3.508 \\
        & 0.4 & 0.199 & 7.933 & 0.638 & 3.020 \\
        \cmidrule(lr){1-6}
        \multirow{3}{*}{5}
        & 0.0 & 0.130 & 3.887 & 0.608 & 0.777 \\
        & 0.2 & 0.249 & 39.010 & 0.808 & 4.615 \\
        & 0.4 & 0.279 & 50.681 & 0.787 & 5.442 \\
        \cmidrule(lr){1-6}
        \multirow{3}{*}{6}
        & 0.0 & 0.229 & 4.664 & 0.857 & 2.063 \\
        & 0.2 & 0.628 & 30.373 & 2.212 & 10.645 \\
        & 0.4 & 0.538 & 39.588 & 2.162 & 10.924 \\
        \cmidrule(lr){1-6}
        \multirow{3}{*}{7}
        & 0.0 & 0.229 & 2.450 & 1.635 & 2.043 \\
        & 0.2 & 0.638 & 34.262 & 2.247 & 10.419 \\
        & 0.4 & 0.508 & 43.259 & 2.064 & 12.229 \\
        \cmidrule(lr){1-6}
        \multirow{3}{*}{8}
        & 0.0 & 1.206 & 14.138 & 3.537 & 9.409 \\
        & 0.2 & 2.033 & 235.358 & 15.269 & 43.196 \\
        & 0.4 & 1.744 & 331.595 & 14.212 & 51.797 \\
        \cmidrule(lr){1-6}
        \multirow{3}{*}{9}
        & 0.0 & 0.967 & 13.705 & 5.178 & 9.890 \\
        & 0.2 & 2.133 & 233.564 & 22.066 & 50.850 \\
        & 0.4 & 1.854 & 290.588 & 23.711 & 95.102 \\
        \cmidrule(lr){1-6}
        \multirow{3}{*}{10}
        & 0.0 & 0.737 & 45.049 & 2.716 & 12.643 \\
        & 0.2 & 2.143 & 339.754 & 21.956 & 70.282 \\
        & 0.4 & 2.312 & 437.833 & 22.335 & 90.076 \\
        \cmidrule(lr){1-6}
        \multirow{3}{*}{11}
        & 0.0 & 3.977 & 318.948 & 36.327 & 61.313 \\
        & 0.2 & 4.156 & 638.084 & 47.560 & 106.293 \\
        & 0.4 & 4.026 & 787.947 & 49.285 & 120.007 \\
        \cmidrule(lr){1-6}
        \multirow{3}{*}{12}
        & 0.0 & 11.631 & 513.518 & 59.361 & 102.268 \\
        & 0.2 & 5.741 & 1032.884 & 68.890 & 167.490 \\
        & 0.4 & 3.922 & 1103.761 & 73.386 & 207.505 \\
        \bottomrule
        \end{tabular}
    }
\end{table}

The efficiency of GB-based classifiers stems from their ability to replace traditional information granules with GBs for data processing. This significantly enhances prediction efficiency by reduced the computational load since GBs serve as inputs. We compared with GBKNN, GBG++, and ACCGBKNN in terms of prediction time and the results are in Table \ref{tab:3}. It can be observed that based on the mapping approach of shadowed GB, the time required for comparing samples with each GB is significantly reduced during sample classification. This is because only samples mapped to the important area require further processing, while those located in the core area samples can directly proceed to voting, greatly reducing prediction time. This performance becomes more pronounced as noise rates increase because GB-based classifiers tend to produce more GBs when facing noise, and these noise-induced GBs are often particularly small. However, dealing with noise GB-based classifiers still compute and sort these balls, while 3WC-SGB directly maps them to unessential area. Therefore, 3WC-SGB will exhibit a greatly improved on efficiency. 

\begin{table*}[htbp]
    \caption{The statistical analysis of traditional classifiers under different noise rates}
    \label{tab:4}
    \centering
    \resizebox{\textwidth}{!}{
        \begin{tabular}{cc*{16}{>{\centering\arraybackslash}p{1.1cm}}}
        \toprule
        \multirow{3}{*}[-2pt]{\centering No.} & \multirow{3}{*}[-2pt]{\centering Metrics} & 
        \multicolumn{4}{c}{Noise Rate: 0.1}& \multicolumn{4}{c}{Noise Rate: 0.2} &
        \multicolumn{4}{c}{Noise Rate: 0.3}& \multicolumn{4}{c}{Noise Rate: 0.4} \\
        \cmidrule(lr){3-6} \cmidrule(lr){7-10} \cmidrule(lr){11-14} \cmidrule(lr){15-18}
        && 3WC-SGB & Naive Bayes & Decision Tree & \multirow{2}{*}{SVM}
        & 3WC-SGB & Naive Bayes & Decision Tree & \multirow{2}{*}{SVM}
        & 3WC-SGB & Naive Bayes & Decision Tree & \multirow{2}{*}{SVM}
        & 3WC-SGB & Naive Bayes & Decision Tree & \multirow{2}{*}{SVM} \\
        \midrule
        \multirow{5}{*}{1}
        & $Accuracy$ &
        0.7658 &0.5353 &0.7027 &0.7956 &0.7175 &0.5167 &0.6500 &0.7364 &0.6744 &0.4647 &0.6165 &0.6433 &0.6617 &0.4202 &0.5607 &0.6030 \\
        & $Precision$ &
        0.4000 &0.3105 &0.2730 &0.0143 &0.3500 &0.2712 &0.3253 &0.0143 &0.3933 &0.2781 &0.2844 &0.1125 &0.3385 &0.2314 &0.2425 &0.2095 \\
        & $Recall$ &
        0.3733 &0.9800 &0.3233 &0.0200 &0.3463 &0.8800 &0.4833 &0.0200 &0.2333 &0.9800 &0.5200 &0.2533 &0.2767 &0.8800 &0.4633 &0.3367 \\
        & $\mathit{F1}\ Sorce$ &
        0.3862 &0.4716 &0.2961 &0.0167 &0.3481 &0.4146 &0.3889 &0.0167 &0.2929 &0.4333 &0.3677 &0.1558 &0.3045 &0.3664 &0.3184 &0.2583 \\
        & $Cost$ &
        2.8772 &4.6615 &3.5228 &2.8618 &3.1652 &4.9222 &3.9302 &3.0387 &3.4792 &5.3681 &4.2222 &3.3402 &3.7066 &5.9014 &4.8245 &3.5051 \\
        \cmidrule(r){1-6} \cmidrule(lr){7-10} \cmidrule(lr){11-14} \cmidrule(l){15-18}
        \multirow{5}{*}{2}
        & $Accuracy$ &
        0.9675 &0.7577 &0.8921 &0.7658 &0.9338 &0.7507 &0.7911 &0.7448 &0.8771 &0.7519 &0.6914 &0.7228 &0.7159 &0.7014 &0.5929 &0.7052 \\
        & $Precision$ &
        0.9927 &0.7937 &0.9421 &0.7896 &0.9723 &0.7867 &0.8690 &0.7913 &0.9582 &0.7889 &0.8050 &0.7918 &0.8336 &0.8008 &0.7246 &0.7686 \\
        & $Recall$ &
        0.9837 &0.8434 &0.8883 &0.8686 &0.9384 &0.8416 &0.8000 &0.8199 &0.8756 &0.8398 &0.6864 &0.7930 &0.7354 &0.8243 &0.5946 &0.8424 \\
        & $\mathit{F1}\ Sorce$ &
        0.9882 &0.8178 &0.9144 &0.8272 &0.9551 &0.8132 &0.8331 &0.8054 &0.9151 &0.8135 &0.7410 &0.7924 &0.7814 &0.8123 &0.6532 &0.8038 \\
        & $Cost$ &
        0.2068 &2.8266 &1.3671 &2.6802 &0.5553 &2.9015 &2.6045 &3.0154 &1.1806 &2.8939 &3.8930 &3.3053 &3.0494 &3.2406 &5.1153 &3.3566 \\
        \cmidrule(r){1-6} \cmidrule(lr){7-10} \cmidrule(lr){11-14} \cmidrule(l){15-18}
        \multirow{5}{*}{3}
        & $Accuracy$ &
        0.7660 &0.6733 &0.7298 &0.6640 &0.7076 &0.6608 &0.6649 &0.6661 &0.6629 &0.6452 &0.5813 &0.6482 &0.6170 &0.6286 &0.5451 &0.6170 \\
        & $Precision$ &
        0.6640 &0.7051 &0.8541 &0.6730 &0.6596 &0.7011 &0.7847 &0.6822 &0.6648 &0.6903 &0.7275 &0.6623 &0.6617 &0.7034 &0.6879 &0.6834 \\
        & $Recall$ &
        0.9905 &0.8710 &0.7112 &0.9539 &0.9873 &0.8452 &0.6744 &0.9347 &0.9776 &0.8314 &0.5864 &0.9603 &0.9126 &0.7405 &0.5531 &0.8000 \\
        & $\mathit{F1}\ Sorce$ &
        0.7950 &0.7793 &0.7761 &0.7892 &0.7908 &0.7665 &0.7254 &0.7887 &0.7915 &0.7543 &0.6494 &0.7840 &0.7672 &0.7215 &0.6132 &0.7371 \\
        & $Cost$ &
        3.0529 &3.6046 &3.4568 &3.4813 &3.4509 &3.7958 &4.2023 &3.5096 &3.4293 &3.9897 &5.2691 &3.6218 &3.5467 &4.3931 &5.7176 &4.3547 \\
        \cmidrule(r){1-6} \cmidrule(lr){7-10} \cmidrule(lr){11-14} \cmidrule(l){15-18}
        \multirow{5}{*}{4}
        & $Accuracy$ &
        0.9646 &0.9347 &0.8730 &0.9854 &0.9456 &0.8848 &0.7334 &0.9354 &0.8623 &0.8281 &0.7052 &0.7854 &0.6191 &0.5911 &0.5920 &0.5854 \\
        & $Precision$ &
        0.6900 &0.4491 &0.0691 &0.0338 &0.3500 &0.3565 &0.0338 &0.0315 &0.4750 &0.3807 &0.3415 &0.3807 &0.5787 &0.2682 &0.2293 &0.3676 \\
        & $Recall$ &
        0.8500 &1.0000 &0.5000 &0.5000 &0.8000 &1.0000 &0.6000 &0.5500 &0.4000 &0.8500 &0.6500 &0.6000 &0.4500 &0.7000 &0.6500 &0.3500 \\
        & $\mathit{F1}\ Sorce$ &
        0.7617 &0.6198 &0.1214 &0.0633 &0.4870 &0.5257 &0.0640 &0.0596 &0.4343 &0.5259 &0.4478 &0.4658 &0.5063 &0.3878 &0.3390 &0.3586 \\
        & $Cost$ &
        1.1103 &1.6529 &1.2975 &1.2048 &0.6974 &1.1516 &2.6933 &0.8048 &0.8159 &0.9273 &2.9641 &1.2048 &2.0259 &2.3007 &4.0999 &2.2048 \\
        \cmidrule(r){1-6} \cmidrule(lr){7-10} \cmidrule(lr){11-14} \cmidrule(l){15-18}
        \multirow{5}{*}{5}
        & $Accuracy$ &
        0.9373 &0.8389 &0.8754 &0.9589 &0.9490 &0.8323 &0.7675 &0.9274 &0.8979 &0.8199 &0.7099 &0.8723 &0.7333 &0.7185 &0.5868 &0.7628 \\
        & $Precision$ &
        0.9964 &0.8532 &0.8584 &0.9551 &0.9958 &0.8588 &0.7254 &0.9019 &0.9808 &0.8463 &0.6689 &0.8430 &0.8440 &0.7354 &0.5332 &0.7374 \\
        & $Recall$ &
        0.9361 &0.7705 &0.8639 &1.0000 &0.9295 &0.7459 &0.7705 &0.9519 &0.8738 &0.7295 &0.7033 &0.8984 &0.7230 &0.7426 &0.5787 &0.7836 \\
        & $\mathit{F1}\ Sorce$ &
        0.9653 &0.8098 &0.8612 &0.9770 &0.9615 &0.7984 &0.7473 &0.9263 &0.9242 &0.7836 &0.6857 &0.8698 &0.7788 &0.7390 &0.5550 &0.7598 \\
        & $Cost$ &
        0.2113 &2.0194 &1.4885 &0.2115 &0.2201 &2.1287 &2.7333 &0.4260 &0.5017 &2.2820 &3.4281 &1.2799 &1.8440 &2.2723 &4.8816 &1.9010 \\
        \cmidrule(r){1-6} \cmidrule(lr){7-10} \cmidrule(lr){11-14} \cmidrule(l){15-18}
        \multirow{5}{*}{6}
        & $Accuracy$ &
        0.9593 &0.9276 &0.8788 &0.9816 &0.9224 &0.9047 &0.8033 &0.8402 &0.8556 &0.7947 &0.6774 &0.7485 &0.7957 &0.7319 &0.5760 &0.7389 \\
        & $Precision$ &
        0.9353 &0.7489 &0.5789 &0.9456 &0.9216 &0.6968 &0.4394 &0.8515 &0.8816 &0.6541 &0.2913 &0.6792 &0.6078 &0.5283 &0.2026 &0.5232 \\
        & $Recall$ &
        0.8420 &0.8351 &0.9038 &0.9394 &0.8839 &0.7339 &0.8047 &0.6514 &0.6655 &0.8350 &0.7088 &0.6100 &0.5408 &0.6307 &0.5685 &0.6278 \\
        & $\mathit{F1}\ Sorce$ &
        0.8862 &0.7897 &0.7058 &0.9425 &0.9023 &0.7149 &0.5684 &0.7381 &0.7585 &0.7335 &0.4129 &0.6428 &0.5723 &0.5750 &0.2988 &0.5707 \\
        & $Cost$ &
        0.5558 &0.8296 &1.2731 &0.2230 &0.7039 &1.1229 &2.0915 &2.2370 &0.7656 &1.1586 &3.4125 &2.0905 &1.6250 &1.9177 &4.5162 &2.2484 \\
        \cmidrule(r){1-6} \cmidrule(lr){7-10} \cmidrule(lr){11-14} \cmidrule(l){15-18}
        \multirow{5}{*}{7}
        & $Accuracy$ &
        0.9520 &0.5710 &0.8926 &0.9118 &0.8958 &0.5048 &0.8065 &0.8303 &0.8299 &0.6697 &0.6788 &0.7571 &0.7013 &0.5554 &0.6048 &0.5571 \\
        & $Precision$ &
        0.9933 &0.2525 &0.5843 &0.9676 &0.8597 &0.2335 &0.4031 &0.7549 &0.9192 &0.3432 &0.2469 &0.3124 &0.5879 &0.2524 &0.1951 &0.1894 \\
        & $Recall$ &
        0.9102 &0.9909 &0.8848 &0.9758 &0.8594 &0.9939 &0.7273 &0.5273 &0.4212 &0.9394 &0.6121 &0.5126 &0.4727 &0.7758 &0.5636 &0.5682 \\
        & $\mathit{F1}\ Sorce$ &
        0.9500 &0.4025 &0.7039 &0.9717 &0.8596 &0.3781 &0.5187 &0.6209 &0.5777 &0.5028 &0.3519 &0.3882 &0.5241 &0.3809 &0.2899 &0.2841 \\
        & $Cost$ &
        0.6693 &4.2952 &1.1394 &0.0961 &0.5723 &4.9558 &2.0909 &0.9671 &0.6944 &3.3377 &3.4338 &2.0000 &1.6078 &4.5740 &4.2017 &2.0000 \\
        \cmidrule(r){1-6} \cmidrule(lr){7-10} \cmidrule(lr){11-14} \cmidrule(l){15-18}
        \multirow{5}{*}{8}
        & $Accuracy$ &
        0.9517 &0.9410 &0.8731 &0.8703 &0.9299 &0.9356 &0.7804 &0.8701 &0.8189 &0.8375 &0.6883 &0.7682 &0.7663 &0.6355 &0.5846 &0.6673 \\
        & $Precision$ &
        0.9203 &0.6453 &0.4527 &0.8392 &0.9305 &0.6262 &0.3020 &0.8797 &0.9011 &0.6321 &0.2104 &0.7736 &0.7676 &0.6259 &0.1437 &0.6622 \\
        & $Recall$ &
        0.8498 &1.0000 &0.8652 &0.8932 &0.8073 &1.0000 &0.8045 &0.8354 &0.7824 &1.0000 &0.6950 &0.7266 &0.7423 &0.8985 &0.5840 &0.6281 \\
        & $\mathit{F1}\ Sorce$ &
        0.8837 &0.7844 &0.5944 &0.8654 &0.8645 &0.7701 &0.4391 &0.8570 &0.8376 &0.7746 &0.3230 &0.7493 &0.7547 &0.7378 &0.2307 &0.6447 \\
        & $Cost$ &
        0.7537 &0.5901 &1.3264 &0.3430 &0.6857 &0.6439 &2.2794 &0.3693 &0.7445 &0.6249 &3.2470 &0.3920 &1.0277 &0.6461 &4.3322 &0.4009 \\
        \cmidrule(r){1-6} \cmidrule(lr){7-10} \cmidrule(lr){11-14} \cmidrule(l){15-18}
        \multirow{5}{*}{9}
        & $Accuracy$ &
        0.9050 &0.8735 &0.8570 &0.8860 &0.8784 &0.7720 &0.7658 &0.8449 &0.8637 &0.7483 &0.6735 &0.7831 &0.8048 &0.7092 &0.5832 &0.7804 \\
        & $Precision$ &
        0.9890 &0.6996 &0.6566 &0.9042 &0.9848 &0.5400 &0.5063 &0.8805 &0.9839 &0.5166 &0.3931 &0.7865 &0.8861 &0.4748 &0.3024 &0.7897 \\
        & $Recall$ &
        0.8250 &0.9081 &0.8454 &0.9660 &0.6596 &0.9277 &0.7587 &0.8056 &0.7908 &0.9349 &0.6752 &0.8412 &0.5591 &0.9354 &0.5740 &0.7269 \\
        & $\mathit{F1}\ Sorce$ &
        0.8996 &0.7903 &0.7392 &0.9341 &0.7900 &0.6826 &0.6073 &0.8414 &0.8769 &0.6655 &0.4969 &0.8130 &0.6856 &0.6299 &0.3961 &0.7570 \\
        & $Cost$ &
        0.4917 &1.3528 &1.5771 &0.5723 &0.4768 &2.3491 &2.5720 &0.7931 &0.4610 &2.5787 &3.5746 &1.2254 &0.8635 &2.9692 &4.5737 &1.2655 \\
        \cmidrule(r){1-6} \cmidrule(lr){7-10} \cmidrule(lr){11-14} \cmidrule(l){15-18}
        \multirow{5}{*}{10}
        & $Accuracy$ &
        0.8153 &0.7743 &0.8337 &0.7938 &0.7770 &0.7790 &0.7416 &0.7672 &0.7433 &0.7262 &0.6352 &0.6706 &0.6907 &0.5969 &0.5789 &0.5820 \\
        & $Precision$ &
        0.9956 &0.8550 &0.8439 &0.8977 &1.0000 &0.8549 &0.7502 &0.7964 &0.9901 &0.8541 &0.6491 &0.6897 &0.9590 &0.5582 &0.5949 &0.5754 \\
        & $Recall$ &
        0.6206 &0.7312 &0.8470 &0.8337 &0.6601 &0.7397 &0.7609 &0.7924 &0.6869 &0.7045 &0.6607 &0.7062 &0.6218 &0.6625 &0.5891 &0.5435 \\
        & $\mathit{F1}\ Sorce$ &
        0.7646 &0.7882 &0.8454 &0.8645 &0.7952 &0.7932 &0.7555 &0.7944 &0.8111 &0.7721 &0.6548 &0.6979 &0.7545 &0.6059 &0.5920 &0.5590 \\
        & $Cost$ &
        0.7508 &2.8138 &1.9799 &1.1996 &0.7411 &2.7493 &3.0789 &1.5513 &0.8081 &2.9501 &4.3514 &2.4882 &1.0077 &2.5234 &5.0622 &4.2975 \\
        \cmidrule(r){1-6} \cmidrule(lr){7-10} \cmidrule(lr){11-14} \cmidrule(l){15-18}
        \multirow{5}{*}{11}
        & $Accuracy$ &
        0.8336 &0.8452 &0.8834 &0.8871 &0.8327 &0.7433 &0.7810 &0.7776 &0.7103 &0.5383 &0.6765 &0.5612 &0.6814 &0.5289 &0.5845 &0.5393 \\
        & $Precision$ &
        0.7436 &0.8815 &0.8151 &0.8712 &0.7440 &0.8813 &0.6696 &0.7482 &0.7381 &0.5754 &0.5421 &0.5075 &0.6742 &0.5648 &0.4440 &0.5676 \\
        & $Recall$ &
        0.7276 &0.7807 &0.8773 &0.8939 &0.7931 &0.6749 &0.7807 &0.7925 &0.4052 &0.6677 &0.6831 &0.6942 &0.3453 &0.5533 &0.5856 &0.5829 \\
        & $\mathit{F1}\ Sorce$ &
        0.7355 &0.8280 &0.8451 &0.8824 &0.7678 &0.7644 &0.7209 &0.7697 &0.5232 &0.6181 &0.6045 &0.5864 &0.4567 &0.5590 &0.5051 &0.5751 \\
        & $Cost$ &
        2.0176 &2.5760 &1.3436 &2.1378 &2.1868 &2.6034 &2.5076 &2.2348 &2.5158 &0.6638 &3.6938 &4.3964 &3.8334 &5.7786 &4.7550 &4.6318 \\
        \cmidrule(r){1-6} \cmidrule(lr){7-10} \cmidrule(lr){11-14} \cmidrule(l){15-18}
        \multirow{5}{*}{12}
        & $Accuracy$ &
        0.7253 &0.7362 &0.7683 &0.7809 &0.7361 &0.7213 &0.6929 &0.7444 &0.7318 &0.7279 &0.6384 &0.7095 &0.6968 &0.6388 &0.5612 &0.6597 \\
        & $Precision$ &
        0.5111 &0.4996 &0.3423 &0.7676 &0.5426 &0.5179 &0.2448 &0.7832 &0.5940 &0.5020 &0.2189 &0.8136 &0.5267 &0.5072 &0.5760 &0.7213 \\
        & $Recall$ &
        0.5236 &0.5430 &0.5189 &0.3281 &0.4188 &0.5058 &0.4789 &0.2626 &0.4273 &0.5026 &0.5168 &0.2018 &0.4209 &0.5063 &0.5000 &0.2174 \\
        & $\mathit{F1}\ Sorce$ &
        0.5173 &0.5204 &0.4125 &0.4597 &0.4728 &0.5117 &0.3240 &0.3933 &0.4971 &0.5023 &0.3075 &0.3234 &0.4679 &0.5067 &0.5353 &0.3341 \\
        & $Cost$ &
        2.3199 &1.9212 &2.6149 &1.6073 &1.2934 &1.8931 &3.3938 &1.7127 &1.3135 &1.9291 &3.9147 &1.7990 &1.4225 &1.9171 &4.6980 &1.9494 \\
        \midrule
        \multirow{5}{*}{$Average$}
        & $Accuracy$ &
        \textbf{0.8786} &0.7841 &0.8383 &0.8567 &
        \textbf{0.8522} &0.7505 &0.7482 &0.8071 &
        \textbf{0.7940} &0.7127 &0.6644 &0.7225 &
        \textbf{0.7070} &0.6214 &0.5792 &0.6498 \\
        & $Precision$ &
        \textbf{0.8193} &0.6412 &0.6059 &0.7216 &
        \textbf{0.7759} &0.6104 &0.5045 &0.6763 &
        \textbf{0.7900} &0.5885 &0.4483 &0.6128 &
        \textbf{0.6888} &0.5209 &0.4063 &0.5663 \\
        & $Recall$ &
        0.7860 &\textbf{0.8545} &0.7524 &0.7644 &
        0.7570 &\textbf{0.8241} &0.7037 &0.6620 &
        0.6283 &\textbf{0.8179} &0.6415 &0.6498 &
        0.5667 &\textbf{0.7375} &0.5670 &0.5840 \\
        & $\mathit{F1}\ Sorce$ &
        \textbf{0.7944} &0.7002 &0.6513 &0.7161 &
        \textbf{0.7496} &0.6611 &0.5577 &0.6343 &
        \textbf{0.6867} &0.6566 &0.5036 &0.6057 &
        \textbf{0.6128} &0.5852 &0.4439 &0.5535 \\
        & $Cost$ &
        \textbf{1.2514} &2.4286 &1.8656 &1.3849 &
        \textbf{1.2291} &2.6014 &2.8481 &1.7217 &
        \textbf{1.3925} &2.3920 &3.7837 &2.2620 &
        \textbf{2.1300} &3.2029 &4.7315 &2.6763 \\
        \cmidrule(r){1-6} \cmidrule(lr){7-10} \cmidrule(lr){11-14} \cmidrule(l){15-18}
        \multirow{3}{*}{$Statistics$}
        & $win/loss$
        &121/59&42/18&47/31&32/28
        &145/35&44/16&54/6&47/13
        &141/39&43/17&50/10&48/12
        &134/46&41/19&48/12&45/15
        \\
        & $rank$
        &\textbf{2.3167} &2.5500 &2.7750 &2.3583 
        &\textbf{2.0833} &2.4333 &2.9167 &2.5667 
        &\textbf{2.1167} &2.2750 &2.9333 &2.6750 
        &\textbf{2.2750} &2.2833 &2.8333 &2.6083 
        \\
        & $p\text{-}values$
        &-&9.85E-05&2.46E-06&2.69E-01 &-&2.27E-05&1.98E-09&3.40E-06
        &-&1.85E-03&5.32E-08&6.40E-06 &-&1.43E-03&2.30E-08&2.51E-04
        \\
        \bottomrule
    \end{tabular}
    }
\end{table*}

\begin{figure}[htbp]
    \centering
    \resizebox{\linewidth}{!}{
        \begin{subfigure}[b]{0.5\linewidth}
            \centering
            \includegraphics[width=\textwidth]{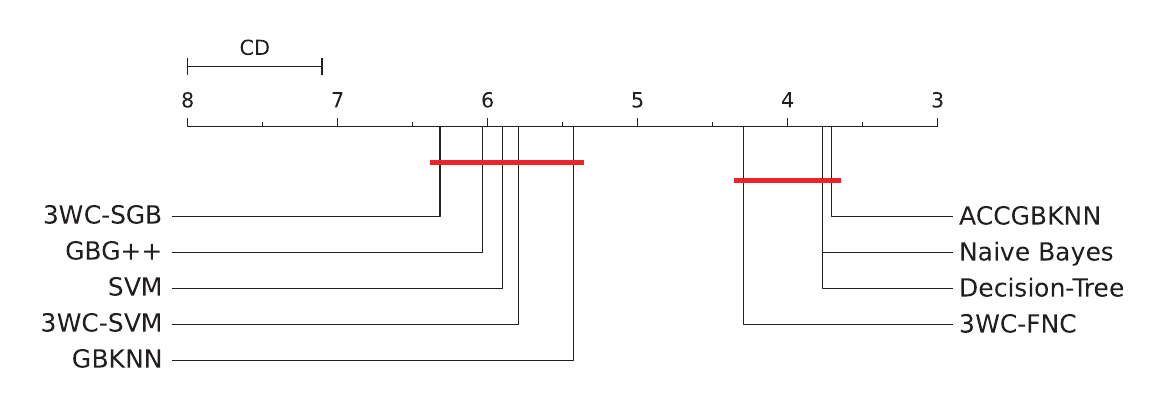}
            \caption{Noise rate: 0.1}
            \label{fig:9:1}
        \end{subfigure}
        \hfill
        \begin{subfigure}[b]{0.5\linewidth}
            \centering
            \includegraphics[width=\linewidth]{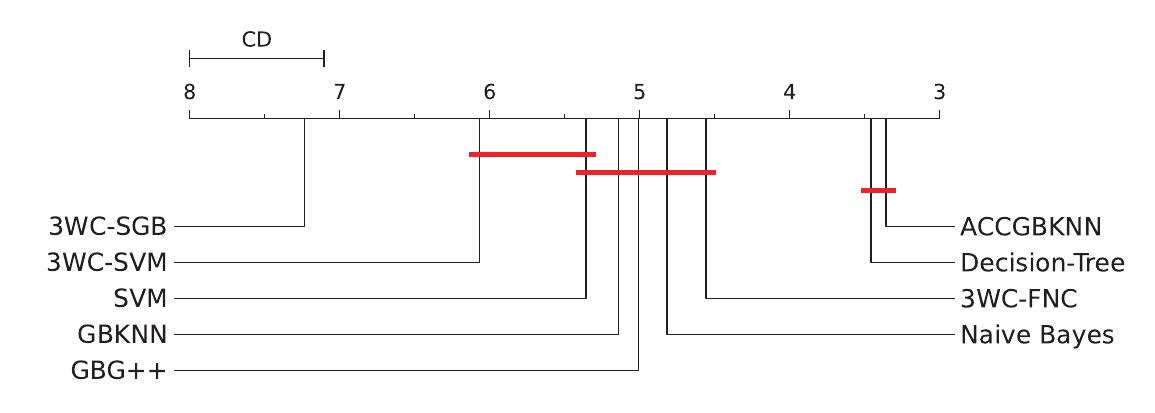}
            \caption{Noise rate: 0.2}
            \label{fig:9:2}
        \end{subfigure}
    }

    \medskip

    \resizebox{\linewidth}{!}{
        \begin{subfigure}[b]{0.5\linewidth}
            \centering
            \includegraphics[width=\linewidth]{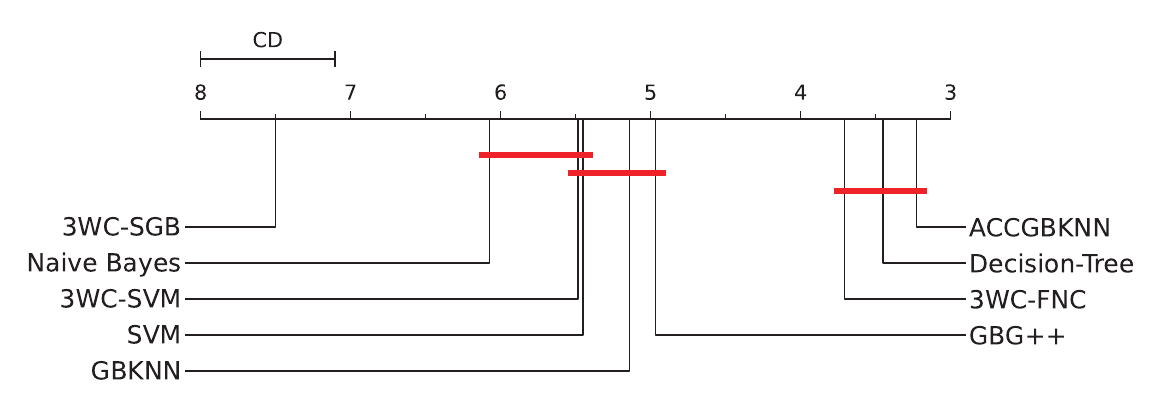}
            \caption{Noise rate: 0.3}
            \label{fig:9:3}
        \end{subfigure}
        \hfill
        \begin{subfigure}[b]{0.5\linewidth}
            \centering
            \includegraphics[width=\linewidth]{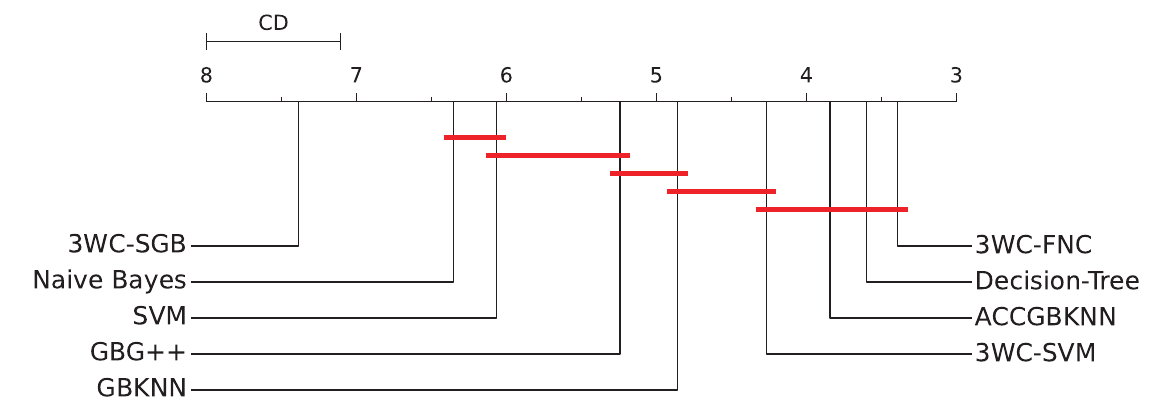}
            \caption{Noise rate: 0.4}
            \label{fig:9:4}
        \end{subfigure}
    }
    \caption{Nemenyi test under different noise rates}
    \label{fig:9}
\end{figure}

\subsection{The Overall Comparison with Other Classifiers}\label{sec:5:sub3}
We conducted experiments with all compared classifiers on various datasets with different noise rates. Here, only the results with traditional classifiers is listed in Table \ref{tab:4}, as the comparison with 3WD classifiers and GB-based classifiers has already been mentioned in Section \ref{sec:5:sub1} and Section \ref{sec:5:sub2}. From Table \ref{tab:4}, although 3WC-SGB experiences some loss in $Recall$, the analysis of the $\mathit{F1}\ Sorce$ indicates that this loss is acceptable. 3WC-SGB consistently maintains a significant lead over traditional algorithms across various noise rates. It closely resembles SVM only in low noise rate, but significant differences emerge as the noise rate increases. It is evident that 3WC-SGB holds a considerable advantage in robustness.

Next, we conducted Nemenyi post-hoc tests on all compared algorithms. This test aim to visually demonstrate the significant differences between different algorithms. The results are presented in Fig. \ref{fig:9}. "CD" stands for Critical Difference, representing the critical range. It is obvious that the results obtained from the Friedman test are consistent with those from the Wilcoxon test. At low noise rate, there is no significant difference between 3WC-SGB and some other algorithms. As noise increases, there is a noticeable rise in the algorithm's ranking, leading to significant differences between 3WC-SGB and other algorithms. This indicates that the robustness of 3WC-SGB indeed presents a significant advantage over other algorithms.

The comprehensive analysis underscores the robustness of 3WC-SGB across various noise rates. Its consistent outperformance in $Precision$, $\mathit{F1}\ Sorce$, and $Cost$, particularly in high noise environments. Therefore, 3WC-SGB matches the effectiveness of top-tier classifiers on most datasets. Moreover 3WC-SGB significantly enhances classification effectiveness in scenarios characterized by high levels of noise.

\subsection{The study of hyperparameter}\label{sec:5:sub4}
In order to achieve higher performance of 3WC-SGB, we studied the value of the parameter $\theta$. The detailed research can be found in Section S4 of supplementary file.

\section{Conclusion}
\label{sec:6}
In this paper, we improved GBC by addressing the limitations in GB generation, and then proposed the 3WC-SGB. Experimental results on 12 datasets demonstrate that the shadowed GBs generated based on justifiable granularity have excellent compatibility with 3WD and significantly improve the performance of the classifier, particularly its robustness. However, our work still has some limitations, for example, searching for justifiable granularity GBs still has an additional time overhead in sorting that cannot be ignored. The focus of our future work will be on the following three aspects:
\begin{itemize}
    \item[-] Design an incremental mechanism into 3WC-SGB to deal with  streaming data, enabling the model can dynamic and efficient classifies in real-time scenarios.
    \item[-] Searched for a novel GB split method to ensure that each split consistently increases the distinct semantics of the new GB, leading to a effectively reduce in GB generation time.
    \item[-] Introducing shadowed GBs into other fields to comprehensive explore its strengths and weaknesses with normal GB.
\end{itemize}

\section*{Acknowledgment}
This work was supported by the National Science Foundation of China (Grant number 62276038, Grant number 62066049, Grant number 62221005), the Guizhou Provincial Department of Education Colleges and Universities Science and Technology Innovation Team (QJJ[2023]084), Excellent Young Scientific and Technological Talents Foundation of Guizhou Province (QKH-platform talent (2021) 5627), Science and Technology Project of Zunyi(ZSKRPT[2023] 3).

\ifCLASSOPTIONcaptionsoff
  \newpage
\fi

\bibliographystyle{IEEEtran}
\bibliography{ref}

\end{document}